\documentclass[preprint]{article}

\usepackage{neurips_2026}

\usepackage{amsmath,amssymb,amsfonts}
\usepackage{graphicx}
\usepackage{booktabs}
\usepackage{hyperref}
\usepackage{xcolor}
\usepackage{algorithm}
\usepackage{algorithmic}
\usepackage{microtype}
\usepackage{subcaption}
\usepackage{textcomp}
\usepackage{float}

\newcommand{\etal}{\textit{et al.}}
\newcommand{\stride}{\textsc{Stride}}
\newcommand{\target}{\textsc{Target}}
\newcommand{\random}{\textsc{Random}}
\newcommand{\fixedrandom}{\textsc{Fixed-Random}}

\title{The Order Is The Message}

\author{
  Jordan LeDoux\\
  Independent Researcher\\
  \texttt{jordan.ledoux@gmail.com}
}

\begin{document}

\maketitle

\begin{abstract}

Neural network training under independent and identically distributed (IID) shuffling treats example ordering as a nuisance variable. We show that it is an information channel. Counterfactual gradient decomposition reveals that the ordering-dependent component accounts for approximately 85\% of each epoch's cumulative gradient norm across all four ordering strategies tested, including IID shuffling. The channel is always active; IID training ensures only that its contributions are incoherent, not that they are absent.

The channel operates through Hessian-gradient entanglement between consecutive training steps: each parameter update displaces the model in a curved loss landscape, modifying subsequent gradients in a direction determined by the ordering. Under IID shuffling or sampling, these ordering effects cancel over many steps; under consistent ordering, they constructively interfere, coherently driving feature acquisition.

In a controlled experiment on modular arithmetic ($p = 9973$), varying only example ordering while holding all else constant, two fixed-ordering strategies achieve 99.5\% test accuracy by epochs 487 and 659 respectively from a training set comprising 0.3\% of the input space, well below established sample complexity lower bounds for this task under IID ordering. The IID baseline achieves 0.30\% after 5{,}000 epochs from identical data. An adversarially structured ordering suppresses learning entirely. The generalizing model reliably constructs a Fourier representation whose fundamental frequency is the Fourier dual of the ordering structure, encoding information present in no individual training example, with the same fundamental emerging across all seeds tested regardless of initialization or training set composition. We discuss implications for training efficiency, the reinterpretation of grokking, and the safety risks of a channel that evades all content-level auditing.

\end{abstract}

\section{Introduction}
\label{sec:intro}

\subsection{Reading Guide}

This paper presents results across multiple levels of analysis. The following guide helps readers navigate to the sections most relevant to their interests, though the paper will make most sense if read in its entirety in the order presented.

\paragraph{For headline results.} Read Section~\ref{sec:models-and-training} to understand the different models that were trained, Section~\ref{sec:generalization} to understand how each model performed, and Section~\ref{sec:spectral} to understand the evidence of the model learning information that was only present in the order of the examples and not the examples themselves.

\paragraph{For reproduction.} Read Section~\ref{sec:methodology} to understand the experimental setup and how to access the code and data, and Section~\ref{sec:generalization} to understand the nature of the results you will be looking for.

\paragraph{For theoretical explanation.} Read Section~\ref{sec:theory} and \ref{sec:theory-experiment} to understand the theoretical basis proposed and how the results confirm the predictions made from a theoretical basis.

\paragraph{For safety and alignment discussion.} Read Section~\ref{sec:theory} to understand the underlying basis of the claims, Section~\ref{sec:spectral} to understand the evidence of ordering channel learning, and Section~\ref{sec:safety} for the safety and alignment concerns this work raises.

\subsection{Motivation}

Large language models are trained on text corpora containing orders of magnitude more words than a human encounters in a lifetime. A frontier model may train on trillions of tokens, roughly three orders of magnitude more linguistic data than a typical child encounters in their first decade of life~\cite{warstadt2022what}. However, humans routinely surpass these models in certain competencies after experiencing only a fraction of the data.

Both humans and neural networks process natural language. Neural networks are demonstrably superior at statistical pattern extraction across large corpora: they can absorb and retain statistical regularities from billions of examples in ways that human memory cannot. The efficiency gap therefore cannot be attributed to data quantity, nor straightforwardly to pattern recognition capability. Something provides a learning signal to humans that models, as currently trained, are not receiving.

Multi-modality of human learning is also unsatisfactory as a complete explanation, as many of the tasks for which language models clearly underperform are strictly linguistic. While the multi-modal experience of humans no-doubt contributes to superior performance in some areas, it cannot on its own explain the entire discrepancy.

The primary candidate is temporal structure. Human learning is profoundly non-IID. A child does not encounter sentences randomly sampled from the distribution of all language. A child hears a sentence about an object, then sees the object, then hears another sentence about it. Related concepts are encountered in temporal proximity. New ideas are presented in the context of previously established ideas. The sequential relationships between consecutive experiences carry information that no individual experience contains in isolation.

Neural network training, by contrast, is designed to destroy this structure. The IID assumption that training examples should be shuffled or sampled into random order was imported from classical statistics, where it serves as a mathematical convenience enabling tractable proofs of convergence and generalization bounds. It was adopted as standard practice in machine learning because it provides reliable, predictable training behavior: under IID sampling, the gradient at each step is an unbiased estimator of the true gradient of the content, and the mathematical machinery of stochastic optimization applies cleanly.

However, the IID assumption was never empirically demonstrated to be \emph{optimal} for learning. It was demonstrated to be \emph{sufficient}, and in a field where the alternative was poorly understood and difficult to control, sufficiency was enough to establish it as universal practice.

We revisit this assumption. Recent work provides empirical grounds for doing so: Lu~\etal~\cite{lu2022grab} demonstrated that deliberately optimizing the ordering of training examples, without changing the examples themselves, produces provably and empirically faster convergence on tasks including CIFAR-10, WikiText, and GLUE benchmarks. These results establish that ordering is not merely a nuisance variable, however both the mechanism through which ordering enters the gradient and the nature of the information it carries have remained uncharacterized. We attempt to explain both of these.

\subsection{Core Hypothesis}

We propose that the sequential ordering of training examples constitutes an information channel that is distinct from the content signal within individual examples. The mechanism is straightforward: training is not a single operation but a sequence of gradient updates, and each update modifies the loss landscape on which subsequent updates operate. When consecutive examples produce gradients that are aligned with respect to a particular feature, the cumulative effect is a coherent directional signal toward that feature's representation. When consecutive examples produce gradients that are randomly oriented, these directional effects cancel over time, leaving only the weaker signal contained in the statistical properties of individual examples.

Under IID shuffling, the ordering signal is randomized. Consecutive batches bear no systematic relationship to each other, so the inter-example gradient correlations point in random directions from step to step. Over sufficient training, these random directional signals destructively interfere and average toward zero. What remains is the content signal: the information that can be extracted from the aggregate statistical properties of individual examples, independent of their ordering.

Under structured ordering, the ordering signal can be made coherent. If examples are sequenced such that consecutive batches produce gradients that are consistently aligned with respect to a target feature, the cumulative effect is a powerful, directed learning signal toward that feature, a signal that operates on top of the content signal from the examples themselves.

Critically, this channel is not introduced by our method. It is always present in every training run. IID shuffling does not eliminate it; it ensures that the channel carries noise rather than signal. Every model ever trained under IID conditions has been shaped by accidental ordering effects arising from the particular random shuffle used in that run. These effects are uncontrolled, unmeasured, and have been invisible to the field because IID guarantees their long-run cancellation. Our contribution is to identify this channel, demonstrate its strength, and show that it can be deliberately controlled.

\subsection{Contributions}

This work makes the following contributions:

\textbf{Empirical demonstration of the ordering channel.} We show that data ordering alone, with all other variables held constant including the exact set of training tokens, produces fundamentally different learned representations, training dynamics, and generalization outcomes. In our most controlled experiment, two fixed-ordering strategies achieve 99.5\% test accuracy on a task where the IID baseline achieves 0.30\% from identical data and compute.

\textbf{Mechanistic characterization.} We provide a detailed mechanistic account of how the ordering signal operates, supported by comprehensive gradient decomposition, spectral analysis, and structural metrics collected at every epoch. In the modular arithmetic setting, we trace the model's internal construction of a Fourier solution harmonic by harmonic, with each new frequency predictable in advance from the theory.

\textbf{Four-way directional control.} We demonstrate four ordering strategies applied to the same data: one that deliberately promotes a target feature (\stride{}), one whose accidental structure is sufficient to drive generalization (\fixedrandom{}), one whose strong but anti-coherent signal suppresses learning entirely (\target{}), and one that is neutral (IID baseline, \random{}). This four-way design isolates the ordering effect from all possible confounds we could account for and establishes that the channel provides controllable influence over feature acquisition, operating on a continuum determined by the interaction between the ordering signal's accumulated strength and its alignment with the task's representational requirements.

\textbf{Counterfactual gradient decomposition.} We introduce a measurement methodology that directly quantifies the ordering-dependent and ordering-independent components of the gradient at each training step, revealing that the ordering-dependent component accounts for the majority of gradient norm under all four ordering strategies.

\textbf{Identification of a covert training channel with safety implications.} The ordering channel is invisible to all existing data auditing methods: every individual example is legitimate, the aggregate data distribution is unchanged, and within-batch gradient statistics are indistinguishable between structured and random ordering. We discuss the implications for AI safety oversight, particularly in settings where one model influences the training of another.

\textbf{Full reproducibility.} All code, training configurations, random seeds, collected metrics, and final model weights are published for immediate reproduction. The headline results can be verified on a single GPU in under 36 hours; full reproduction with all instrumentation requires approximately 60 hours.

\subsection{Terminology}

To avoid ambiguity, we adopt the following terminology throughout this paper.

\textbf{Ordering channel.} The information pathway through which the sequential arrangement of training examples influences learned representations. The channel is a property of gradient-based training itself, not of any particular ordering method. It is always active; the question is whether it can carry productive, coherent signal, or if it can only act as regularizing noise.

\textbf{Ordering strategy.} A method for determining the sequence in which training examples are presented. We distinguish this from curriculum learning, which selects \emph{which} examples the model sees; our ordering strategies determine only the \emph{sequence} in which a fixed set of examples is presented. All strategies see the same examples the same number of times. The four strategies used in our experiment (\stride{}, \fixedrandom{}, \random{}, and \target{}) are described in Section~\ref{sec:experiment}.

\textbf{Content component} and \textbf{ordering component.} The two components of the gradient identified by our counterfactual decomposition (Section~\ref{sec:counterfactual}). The content component is the portion of the observed gradient attributable to the data independent of presentation order. The ordering component is the residual: the portion that arises from the sequential structure of the epoch rather than from the statistical properties of individual examples. The ordering component includes \emph{all} ordering-dependent gradient information, productive or otherwise, and its magnitude characterizes the total gradient energy flowing through the ordering channel at a given point in training.

\textbf{Content signal} and \textbf{ordering signal.} The productive subsets of the content and ordering components: the portions that result in lasting parameter changes which move the model toward competence at the task. The ordering signal is necessarily smaller than the ordering component, because the component includes ordering-dependent noise in addition to learnable information. Throughout this paper, we distinguish between the \emph{component} (which is directly measurable via counterfactual decomposition) and the \emph{signal} (which is inferred from its effects on learning outcomes such as spectral concentration and generalization).

\textbf{Entanglement term.}\footnote{The analogy is apt: the term arises from the
inseparable interaction between consecutive examples, not from either
example alone. We note that Cosson \etal~\cite{cosson2022gradient} recently used this same phrase to describe spatial, overlapping gradient signals from different loss terms that cannot be disentangled. However, we retain our usage here, as we believe it is the most mechanically precise description of the temporal intertwining of parameter trajectories in a curved loss landscape.} The Hessian-gradient interaction $\eta \, H_B(\theta) \cdot \nabla L_A(\theta)$ that arises when training on batch $B$ after batch $A$ has modified the parameters (described in Section~\ref{sec:entanglement}). This is the mathematical object through which data ordering enters the gradient. We use ``entanglement'' in its ordinary English sense of being intertwined: the term arises from the inseparable interaction between consecutive examples, not from either example alone.

\textbf{Constructive and destructive interference.} When entanglement terms across consecutive training steps point in a consistent direction, they sum linearly (\emph{constructive interference}) and the cumulative ordering signal grows proportionally with the number of steps. When they point in random directions, they sum as a random walk (\emph{destructive interference}) and the cumulative signal grows only as the square root of the number of steps. IID shuffling produces destructive interference; structured ordering can produce constructive interference along chosen directions.

\textbf{Critical learning period.} The training window during which a particular feature's representation is forming and the ordering signal has maximal influence over its structure (Section~\ref{sec:theory}). This is a per-feature property, not a global property of training: different features may have critical learning periods at different points during optimization.

\section{Related Work}
\label{sec:related}

\textbf{Curriculum learning.} Bengio~\etal~\cite{bengio2009curriculum} demonstrated that ordering training examples from easy to difficult can improve convergence speed and generalization. Subsequent work has explored various difficulty metrics and scheduling strategies~\cite{kumar2010self,graves2017automated}. Our work differs fundamentally: curriculum learning varies \emph{which} examples the model sees and when, while our ordering strategies present \emph{the same examples} in different sequences. The information channel we identify is carried by inter-example relationships, not by example selection. Wu~\etal~\cite{wu2021curricula} conducted thousands of ordering experiments and found that curricula have only marginal benefits on clean data trained to convergence, attributing most benefits to dynamic dataset size. Their experiments used IID-shuffled orderings within each curriculum stage; our results suggest that it is \emph{fixed} orderings, not curricula per se, that unlock the ordering channel, because IID shuffling within stages ensures the entanglement terms remain incoherent. Mohtashami~\etal~\cite{mohtashami2022characterizing} showed that deterministic orderings can outperform random reshuffling, consistent with our finding that ordering consistency is a key variable.

\textbf{Grokking and delayed generalization.} Power~\etal~\cite{power2022grokking} discovered that models trained on algorithmic tasks exhibit sudden generalization long after memorizing the training set. Nanda~\etal~\cite{nanda2023progress} provided mechanistic interpretability analysis showing that grokking in modular arithmetic involves the construction of Fourier features. Lee~\etal~\cite{lee2024grokfast} introduced Grokfast, which accelerates grokking by amplifying slow-varying gradient components. Our work provides a complementary perspective: structured ordering eliminates the grokking delay entirely by coherently driving the model toward the Fourier solution from the first epoch, bypassing the memorization phase that produces the delayed generalization phenomenon. We note that Mallinar~\etal~\cite{mallinar2025emergence} demonstrated grokking in non-neural kernel machines (Recursive Feature Machines), which may operate outside the Hessian-gradient entanglement mechanism as described here. Whether the ordering channel contributes to grokking dynamics in non-SGD-trained models remains an open question.

\textbf{Non-IID training.} The effects of non-IID data ordering have been studied primarily as a problem to be mitigated, particularly in federated learning~\cite{zhao2018federated,karimireddy2020scaffold}. Shumailov~\etal~\cite{shumailov2021manipulating} demonstrated that adversarial data ordering alone---without modifying any training examples---can degrade or manipulate model behavior, establishing the security relevance of ordering effects. Beneventano~\cite{beneventano2023trajectories} extended backward error analysis to SGD without replacement, showing that the Hessian-gradient interaction creates an ordering-specific implicit regularizer whose form depends on the permutation used. These works collectively demonstrate that non-IID ordering degrades convergence, reinforcing the field's commitment to IID shuffling.

\textbf{Ordering optimization.} Lu~\etal~\cite{lu2022grab} introduced GraB, which formulates example ordering as an online discrepancy minimization problem, using stale gradients from prior epochs to greedily construct permutations that provably converge faster than random reshuffling. GraB demonstrates practical gains on CIFAR-10, WikiText, and GLUE tasks, establishing that ordering optimization is beneficial beyond toy settings. Cooper~\etal~\cite{cooper2023coordinating} extended this approach to distributed training, coordinating orderings across workers for provably accelerated convergence. These works treat ordering as a convergence-rate phenomenon: GraB minimizes gradient sum discrepancy to reduce the variance of the stochastic gradient estimator. Our framework provides a complementary perspective: in the entanglement framework, GraB's gradient balancing shapes the coherence of consecutive entanglement terms, and the convergence gains it achieves are a specific consequence of making the ordering channel less noisy. Our contribution is to show that the channel can carry not only less noise but distinct, productive signal that the content channel does not provide, and to identify the mechanism through which this occurs.

\textbf{SGD ordering theory.} The Hessian-gradient interaction through which ordering enters the gradient has been studied in the optimization literature. Smith~\etal~\cite{smith2021origin} derived the same second-order cross-term via backward error analysis for SGD with random shuffling, showing it produces an ordering-dependent implicit regularizer. Barrett and Dherin~\cite{barrett2021implicit} identified the analogous term for full-batch gradient descent. G{\"u}rb{\"u}zbalaban~\etal~\cite{gurbuzbalaban2021why} built their convergence analysis for random reshuffling around controlling this Hessian-gradient term, and Recht and R{\'e}~\cite{recht2012beneath} formalized ordering effects through non-commuting matrix products. More recently, HaoChen and Sra~\cite{haochen2019random} and Ahn~\etal~\cite{ahn2020sgd} established tight convergence rates for SGD without replacement, and Mishchenko~\etal~\cite{mishchenko2020random} provided simplified analyses with improved bounds. Rajput~\etal~\cite{rajput2022permutation} showed that the convergence gap between optimal and random permutations ranges from exponential to nonexistent depending on the function class, establishing that the potential benefit of ordering optimization is fundamentally problem-dependent. These works analyze the ordering-dependent term in service of convergence guarantees. Our contribution is to measure its \emph{magnitude} empirically (showing it dominates per-step gradient norm), characterize its \emph{coherence} properties across ordering strategies, and demonstrate that it can be deliberately exploited as an information channel for feature acquisition.

\textbf{Data influence and selection.} Recent work on data attribution~\cite{koh2017understanding,ilyas2022datamodels} and data selection for pretraining~\cite{xie2023doremi,engstrom2024dsdm} has studied how individual examples affect model behavior. These approaches measure per-example influence on the trained model; our work identifies an orthogonal axis of influence that arises from the \emph{relationships between} consecutive examples rather than from any example individually.

\textbf{Fourier features in neural networks.} The emergence of Fourier features in networks trained on modular arithmetic is well-established~\cite{nanda2023progress,zhong2023clock,gromov2023grokking}. Gromov~\cite{gromov2023grokking} provided analytical solutions for the optimal Fourier representations. Our contribution is showing that structured ordering controls \emph{which} Fourier features emerge and at what rate, with the fundamental frequency determined by the number-theoretic properties of the ordering structure.

\section{Theoretical Framework}
\label{sec:theory}

This section presents a mechanistic account of how data ordering produces a learning signal. We begin from the well-known Taylor expansion of the gradient at consecutive training steps~\cite{smith2021origin,barrett2021implicit}, identify the specific term through which ordering exerts its influence, reinterpret it as an information channel with measurable coherence properties, and connect the theoretical predictions to the empirical measurements reported in Section~\ref{sec:experiment}.

\subsection{Non-Commutativity of SGD Updates}

It is well known that SGD updates do not commute in non-convex landscapes~\cite{bottou2018optimization}: training on batch $A$ then $B$ yields a different parameter state than $B$ then $A$, because each update shifts the point at which the next gradient is evaluated. Under convexity this is irrelevant to the final solution; under the non-convex losses of neural networks, it means the training trajectory, and thus the basin of convergence, depends on ordering. 

A substantial body of work has quantified how permutation order affects convergence rates~\cite{safran2020good,gurbuzbalaban2021why,haochen2019random,ahn2020sgd,mishchenko2020random}, and backward error analysis has shown that the Hessian-gradient cross-term arising from finite learning rates produces an ordering-dependent implicit regularizer~\cite{smith2021origin,barrett2021implicit}. Beneventano~\cite{beneventano2023trajectories} extended this analysis to SGD without replacement, showing that the resulting regularizer is \emph{permutation-specific}: different orderings of the same data produce different implicit regularizers, meaning the choice of permutation shapes the optimization landscape itself. Recht and R{\'e}~\cite{recht2012beneath} formalized the non-commutativity through products of non-commuting operators. These analyses focus on convergence speed to a fixed optimum and on the regularization properties of the cross-term. That the term can be practically exploited has been demonstrated by Lu~\etal~\cite{lu2022grab}, whose GraB algorithm optimizes data permutations by minimizing gradient discrepancy and achieves measurable convergence gains on standard benchmarks.

What has been missing is a characterization of this term as an \emph{information channel}: an empirical decomposition that measures how much of each gradient step is attributable to ordering, how the coherence of this component varies across ordering strategies, and whether the channel can be deliberately exploited to control which features a model acquires. We provide this next.

\subsection{The Ordering Signal: A Hessian-Gradient Interaction}
\label{sec:entanglement}

\subsubsection{Derivation}

Consider two consecutive training steps. At parameters $\theta$, we update on batch $A$:
$$\theta' = \theta - \eta \nabla L_A(\theta)$$

We then compute the gradient for batch $B$ at the new parameters $\theta'$. Expanding this gradient to first order around $\theta$:
$$\nabla L_B(\theta') \approx \nabla L_B(\theta) + H_B(\theta) \cdot (\theta' - \theta)$$

where $H_B(\theta) = \nabla^2 L_B(\theta)$ is the Hessian of the loss for batch $B$ evaluated at $\theta$. Substituting the update rule:
\begin{equation}
\nabla L_B(\theta') \approx \underbrace{\nabla L_B(\theta)}_{\text{content term}} - \underbrace{\eta \, H_B(\theta) \cdot \nabla L_A(\theta)}_{\text{entanglement term}}
\label{eq:decomposition}
\end{equation}

The gradient used for the second update thus has two components:

The \textbf{content term}, $\nabla L_B(\theta)$: the gradient that batch $B$ would produce if $A$ had not been trained on first. This is the standard gradient signal; it depends only on the content of $B$ and the current parameters.

The \textbf{entanglement term}, $-\eta \, H_B(\theta) \cdot \nabla L_A(\theta)$: an additional gradient contribution arising from the interaction between batch $A$'s gradient and batch $B$'s loss curvature. This term depends on the \emph{relationship} between $A$ and $B$, specifically on how the direction of $A$'s update aligns with the curvature structure of $B$'s loss.

This decomposition is standard; the same Hessian-gradient cross-term appears in analyses of implicit regularization in SGD~\cite{smith2021origin,barrett2021implicit} and in convergence proofs for random reshuffling~\cite{gurbuzbalaban2021why,haochen2019random}. What has not been appreciated is that this term constitutes an \emph{information channel}. The entanglement term does not exist in any individual example; it arises only from the sequential interaction between consecutive examples. It is the mathematical object through which data ordering exerts its influence on learning, and its coherence properties, (whether it constructively or destructively interferes across steps), determine whether the channel carries usable signal or self-canceling noise.

\subsubsection{Behavior Under Random Ordering}

Under IID shuffling, the gradient $\nabla L_A(\theta)$ is drawn from a distribution that is independent of $B$. The entanglement term $H_B(\theta) \cdot \nabla L_A(\theta)$ therefore projects a random direction through $B$'s Hessian. Over many steps, these projections point in random directions and their cumulative effect averages toward zero. This is the cancellation that underpins classical convergence guarantees for SGD with random reshuffling~\cite{gurbuzbalaban2021why,mishchenko2020random}.

The cancellation is statistical, not exact. In any finite training window, the random entanglement terms do not perfectly cancel; there is always some residual ordering effect. This residual is an uncontrolled source of variance between training runs that is typically attributed to ``initialization noise'' or ``training stochasticity.'' Our framework suggests that a portion of this variance is attributable to accidental ordering effects that persist through finite-sample noise in the cancellation.

In expectation, IID shuffling ensures that the entanglement terms cancel cumulatively, so the long-run parameter trajectory is driven by the content term alone. At each individual step, however, the entanglement term remains large: our measurements show it accounts for the majority of per-step gradient norm even under IID conditions (Section~\ref{sec:counterfactual}). This is the regime that classical convergence theory describes.

\subsubsection{Behavior Under Structured Ordering}

Under structured ordering, the relationship between consecutive batches is no longer random. If batches $A$ and $B$ are selected such that $\nabla L_A(\theta)$ aligns with a principal eigenvector of $H_B(\theta)$, the entanglement term is amplified: $B$'s curvature structure reinforces $A$'s gradient direction, producing a large, coherent contribution to the effective gradient.

More precisely, let $H_B = \sum_i \lambda_i v_i v_i^T$ be the eigendecomposition of the Hessian. Then:
\begin{equation}
H_B \cdot \nabla L_A = \sum_i \lambda_i (v_i^T \nabla L_A) \, v_i
\label{eq:hessian_decomp}
\end{equation}

The entanglement term is largest when $\nabla L_A$ is aligned with eigenvectors corresponding to large eigenvalues $\lambda_i$, that is, when $A$'s gradient points along directions of high curvature in $B$'s loss. Structured ordering exploits this by selecting consecutive examples whose gradients are mutually aligned and point along high-curvature directions of each other's loss surfaces.

Importantly, the entanglement term need not dominate the entire gradient to have a significant impact on learning. It is sufficient for it to operate on the subset of parameters that control the feature being acquired, for instance the embedding weights that encode a particular Fourier frequency. A coherent signal in this subspace can guide the corresponding feature into a particular basin of attraction, after which the remaining parameters adjust to accommodate the committed representation through ordinary gradient descent on the content signal.

\subsubsection{Signal Strength}
\label{sec:signal-strength}

The relative magnitude of the entanglement term versus the content term depends on three factors: the learning rate $\eta$ (which scales the displacement and thus the entanglement term linearly), the spectral norm of the Hessian (which determines how strongly curvature amplifies the displacement), and the alignment between consecutive gradients (which determines what fraction of the displacement is amplified).

For neural networks in typical training regimes, the Hessian has a small number of large eigenvalues and a large number of small ones, a well-documented spectral structure. When consecutive gradients are aligned with the large-eigenvalue directions, the entanglement term can be comparable to or larger than the content term. Our per-step Hessian measurements (Section~\ref{sec:adam-empirical}, Appendix~\ref{apdx:hessian}) reveal that the entanglement and content terms are in fact nearly identical in both magnitude and direction (cosine similarity $>0.999$), making the observed gradient a small residual of two large, nearly-canceling vectors. The ordering signal lives in this residual: small angular differences between entanglement and content, determined by which batch follows which, are amplified into large directional changes in the gradient the optimizer sees. Our epoch-level counterfactual decomposition (Section~\ref{sec:counterfactual}) confirms that this per-step structure accumulates: the ordering component accounts for the majority of each epoch's cumulative gradient norm.

\paragraph{Component vs. signal.} It is important to note that the ordering \emph{component} contains all ordering-dependent gradients, not only the productive ones. Even under highly structured ordering strategies, the component will contain noise in addition to learnable signal. The magnitude of the ordering component relative to the content component characterizes the channel's bandwidth, not the amount of productive learning that occurs through the ordering channel.

\subsubsection{Extension to Adaptive Optimizers}
\label{sec:adam-theory}

The derivation above uses vanilla SGD for clarity, but the mechanism is not specific to SGD. The entanglement term arises from the non-commutativity of parameter updates in curved loss landscapes, and any optimizer that produces a parameter displacement creates a Hessian-displacement interaction with the subsequent batch.

In the general case, let $\Delta\theta$ denote the actual parameter displacement produced by an optimizer step on batch $A$. The Taylor expansion of the subsequent gradient becomes:
\begin{equation}
\nabla L_B(\theta + \Delta\theta) \approx \nabla L_B(\theta) + H_B(\theta) \cdot \Delta\theta
\label{eq:general_decomposition}
\end{equation}

Under SGD, $\Delta\theta = -\eta \nabla L_A(\theta)$ and the entanglement term reduces to Equation~\ref{eq:decomposition}. Under Adam~\cite{kingma2015adam} or AdamW~\cite{loshchilov2019decoupled}, the displacement is:
\begin{equation}
\Delta\theta = -\eta \frac{\hat{m}_t}{\sqrt{\hat{v}_t} + \epsilon}
\label{eq:adam_update}
\end{equation}

where $\hat{m}_t$ and $\hat{v}_t$ are the bias-corrected first and second moment estimates. This transformation modulates the ordering channel through three pathways.

\paragraph{Amplified displacement.} Adam's per-parameter adaptive scaling can produce displacements whose norm substantially exceeds the raw gradient norm, particularly for parameters with small but consistent gradients. Because the entanglement term scales with $\|H_B \cdot \Delta\theta\|$, larger displacements produce proportionally larger entanglement, amplifying the ordering channel relative to the SGD baseline.

\paragraph{Momentum as temporal integration.} The first moment $\hat{m}_t$ is an exponential moving average of past gradients with decay rate $\beta_1$. Under structured ordering, if consecutive gradients share a consistent component in some subspace (the ordering signal), momentum accumulates this component across steps. Under IID ordering, the ordering components of consecutive gradients point in unrelated directions and partially cancel within the momentum buffer. Momentum therefore acts as a temporal integrator that preferentially preserves coherent signals, whether from content or from ordering, while attenuating incoherent ones. The displacement $\Delta\theta$ inherits this filtered signal, and the entanglement term on the next step reflects the accumulated, filtered history rather than only the most recent batch.

\paragraph{Selective parameter scaling.} The second moment $\hat{v}_t$ normalizes each parameter by its recent gradient variance. Parameters that receive consistent gradient signals (low variance) are assigned larger effective learning rates; parameters that receive noisy signals (high variance) are assigned smaller ones. Under structured ordering, the gradient signal is more consistent along parameters that participate in the feature being driven by the ordering, producing larger effective learning rates for precisely those parameters. This creates a feedback loop: the ordering signal is selectively amplified in the subspace it targets.

The interaction between these pathways can be complex. Momentum smoothing and adaptive scaling can either amplify or attenuate the ordering signal depending on the relationship between the ordering's temporal coherence and the optimizer's characteristic timescales ($1/(1-\beta_1)$ for momentum, $1/(1-\beta_2)$ for variance estimation). A full analytical treatment of these interactions is beyond the scope of this work. What matters for the present argument is that the Hessian-displacement mechanism persists under any optimizer that produces non-zero parameter updates, and that adaptive optimizers introduce additional structure into the displacement vector that can differentially modulate the ordering channel's strength across parameter subspaces. We verify this empirically in Section~\ref{sec:adam-empirical}.

\subsection{Constructive and Destructive Interference}

The wave-interference analogy is more than a metaphor. The entanglement terms from consecutive training steps are additive contributions to the effective gradient. When these contributions point in a consistent direction across many steps, they sum constructively: the cumulative ordering signal grows linearly with the number of steps. When they point in random directions, they sum like a random walk: the cumulative signal grows only as the square root of the number of steps, and the signal-to-noise ratio decreases with training duration.

This framing explains several empirical observations:

\paragraph{Why IID training eventually works mechanistically.} Even under random ordering, the content term is present at every step and points in a consistent direction (toward reducing the training loss). The ordering noise slows convergence and may misdirect the optimization into suboptimal basins, but it does not prevent convergence entirely: the content signal eventually dominates the random ordering noise given sufficient training.

\paragraph{Why structured ordering is dramatically more efficient.} Under structured ordering, both the content term and the entanglement term are aligned within some critical subspace of the feature. The model receives a coherent signal from two channels simultaneously rather than one.

\paragraph{Why anti-coherent ordering is destructive.} If the ordering produces entanglement terms that overwhelm the content signal but whose structure is anti-aligned with useful feature acquisition, the model is driven toward degenerate representations. The ordering signal does not merely add noise; it coherently organizes specific components of the model while preventing others from forming, because consecutive batches produce gradients that are structured but self-contradictory with respect to the task. This is distinct from the incoherent case (\random{}), where the ordering signal cancels over time. Under anti-coherent ordering, the signal does not cancel; it accumulates toward a basin that is organized but not useful. This is the failure mode we observe with the \target{} ordering strategy (Section~\ref{sec:experiment}), and it is the failure mode that historically led researchers to conclude that non-IID ordering is harmful.

\paragraph{Why the ordering signal is invisible to standard monitoring.} The entanglement term modifies the gradient for batch $B$ based on what happened during batch $A$. Within any single batch, the gradient statistics appear normal: the ordering signal manifests as temporal correlations \emph{across} batches, not as anomalies \emph{within} a batch. Standard training monitoring examines per-batch metrics but does not examine cross-step gradient correlations. The ordering channel operates in a dimension that existing monitoring tools do not measure.

Additionally, the ordering component of the gradient norm is large under \emph{all} ordering strategies (Section~\ref{sec:counterfactual}): the counterfactual decomposition shows that ordering accounts for the majority of each epoch's cumulative gradient magnitude even under IID shuffling. Because its cancellation toward zero is a long-horizon statistical effect, any short window of training will exhibit substantial ordering-induced gradient structure regardless of whether the ordering is deliberate, productive, or impactful to the overall training. Distinguishing a coherent signal embedded in the ordering component from the large but incoherent background present in every training run requires knowing which directions in parameter space to monitor, which in turn requires prior knowledge of the target behavior. Without this, the signal may be indistinguishable from the noise floor.

\subsection{Basin Selection and the Critical Learning Period}

In non-convex optimization, the loss landscape contains many basins of attraction, parameter regions that produce locally minimal loss. Different basins may achieve similar training loss but correspond to qualitatively different learned representations.

The ordering signal influences which basin the optimization trajectory enters. Because the entanglement term is most influential when the loss landscape is highly curved with respect to a forming feature, when the model's representation of that feature is still undifferentiated and small perturbations can determine its structure, the ordering during this period can determine the basin for that feature's representation.

We term this the \emph{critical learning period} for a feature. In our experiment, where the model learns a single representation, the critical learning period coincides with early training: the \stride{} model identifies the fundamental frequency $F = 101$ within the first 3 epochs, and by epoch 30 the spectral structure of the eventual solution is already visible. However, we expect the critical learning period to be a per-feature property in general, not a global property of training. A feature that begins forming midway through training would have its critical learning period at that point, not at initialization. The ordering signal's influence on that feature would be maximal during \emph{its} period of formation, regardless of when during overall training that occurs.

If the ordering signal's influence is concentrated during feature formation, this should be observable in the ordering fraction at the layers most involved in the emerging representation. As the model commits to a basin and the relevant parameters leave the region of high curvature, the same ordering should produce diminishing gradient displacement in those parameters, not because the external signal has weakened, but because the parameters have moved to a region where it no longer has leverage. The ordering fraction at specific layers should therefore decline as the corresponding feature crystallizes, with the rate of decline tracking the pace of feature formation. We observe this effect in Section~\ref{sec:counterfactual}.

This distinction has practical implications. If the critical learning period were global and early, one could in principle audit the ordering only at the start of training. If it is per-feature and distributed throughout training, the window of vulnerability to ordering-based influence is coextensive with training itself.

\subsection{Empirical Predictions}

The theoretical framework generates several empirically testable predictions:

\begin{enumerate}
\item The ordering component of the gradient should be measurable by comparing the gradient under structured ordering to the gradient under counterfactual random ordering of the same examples.
\item The ordering signal should produce measurable spectral concentration in the model's weight representations.
\item Ordering that is both strong and misaligned with the task's representational requirements should prevent convergence.
\item The ordering signal should be strongest early in feature formation, when the loss landscape is most curved.
\item The ordering and content components should be partially but not fully aligned, indicating the ordering channel carries partially independent information.
\item Under an adaptive optimizer, the displacement amplification and selective scaling should produce entanglement energies substantially exceeding the SGD prediction, with the amplification varying across parameter subspaces according to each subspace's gradient consistency.
\end{enumerate}

These predictions are confirmed by the experimental results presented in Section~\ref{sec:experiment}.

\section{Methodology}
\label{sec:methodology}

\subsection{Experimental Setup}

\subsubsection{Task and Motivation}

We study addition modulo a prime $p = 9973$. The task is to learn the function $f(a, b) = (a + b) \bmod p$ from a sparse subset of the input space. This setting is chosen for three reasons: (1) the problem has known analytical solutions in terms of Fourier features~\cite{nanda2023progress,gromov2023grokking}, enabling mechanistic verification of learned representations; (2) the problem is simple enough that all relevant metrics can be computed at every epoch within a tractable time budget; and (3) the grokking phenomenon in this setting is well-studied, providing a rich baseline of prior results against which our ordering effects can be measured.

Preliminary experiments at $p = 97$ showed rapid generalization across \stride{}, \fixedrandom{}, and \random{} strategies under high data density, consistent with the content signal being sufficient in that regime. The choice of $p = 9973$ at 0.3\% data density was designed to place the experiment below the theoretical critical data fraction for IID ordering, isolating the ordering channel's contribution in a regime where content alone is provably insufficient. The large prime additionally provides finer spectral resolution (9,973 possible frequencies versus 97), enabling detailed tracking of harmonic emergence and clear spectral differentiation between ordering strategies.

At $p = 97$ with 26.5\% data density, all three non-adversarial strategies generalize, with fixed orderings achieving approximately 20\% faster convergence than IID shuffling (single seed). The ordering channel's contribution is modest in this regime because the content signal alone is sufficient for generalization. This contrasts with the $p = 9973$ experiment at 0.3\% density, where the content signal is insufficient and the ordering channel determines whether generalization occurs at all. The ordering channel's practical importance thus scales inversely with content signal adequacy.

This aspect is discussed in greater depth in Section~\ref{sec:regime-dependence}.

\subsubsection{Models and Training}
\label{sec:models-and-training}

\paragraph{Architecture.} We use a two-layer pre-LayerNorm transformer with embedding dimension 256, 4 attention heads, and feedforward dimension 2048 (the PyTorch \texttt{nn.TransformerEncoderLayer} default). Each integer $a \in \{0, \ldots, p-1\}$ is embedded via a learned embedding table and summed with a learned positional embedding indicating operand position. The model processes pairs $(a, b)$, mean-pools over the sequence dimension, and produces a probability distribution over outputs $c \in \{0, \ldots, p-1\}$ via a linear decoder.

\paragraph{Training data.} We sample 300{,}000 input pairs $(a, b)$ uniformly at random as the training set, comprising approximately 0.3\% of the $9973^2 \approx 10^8$ possible pairs. A separate test set of 1{,}000{,}000 pairs is sampled from the remaining pairs. The same training and test sets are used across all ordering strategies.

\paragraph{Optimizer.} AdamW with weight decay $0.1$, and default momentum parameters ($\beta_1 = 0.9$, $\beta_2 = 0.999$). 

\paragraph{Scheduler.} Learning rate follows a cosine annealing schedule over 5{,}000 epochs with learning rate $1 \times 10^{-3}$, and minimum learning rate $5 \times 10^{-7}$.

\paragraph{Training regime.} Batch size is 256. All models are trained for 5{,}000 epochs, or until they score greater than or equal to 99.5\% on the test set.

\paragraph{Ordering strategies.} We compare four strategies:

\begin{itemize}
\item \stride: Examples are sorted by $(a \bmod s, \; a)$ where $s = \lfloor \sqrt{p} \rfloor = 99$. This creates a strided traversal of the input space that groups nearby elements within blocks of size 99. The ordering is computed once and held fixed across all epochs.

\item \fixedrandom: A single random permutation is generated at initialization (determined by the random seed) and held fixed across all epochs. This permutation will necessarily have \emph{some} cyclic structures that are useful for Fourier features, but those structures will be noisy and undesigned.

\item \random: Standard IID shuffling with a fresh random permutation each epoch.

\item \target: Examples are sorted directly by their output value $(a + b) \bmod p$. This creates maximum structure, with consecutive examples having consecutive outputs, producing an ordering signal that is both strong and self-contradictory, a combination that prevents stable learning.
\end{itemize}

\textbf{Shared conditions.} All four models share the same architecture, parameter initialization (via a shared random seed), training set, optimizer, hyperparameters, and compute budget. The only variable is the ordering of training examples within each epoch.

\subsection{Reproducibility and Data}

\subsubsection{Reproduction Steps}
\label{sec:reproduction-steps}

The code that was used to produce the results in this paper can be found at: 

\href{https://github.com/JordanRL/OrderedLearning/}{github.com/JordanRL/OrderedLearning}

Clone the repository and then checkout the tag 'paper-data-v3', which is the tag that was used to generate the exact data found in this paper.

The raw data collected, including weights, raw metrics, and checkpoints, can be found at:

\href{https://github.com/JordanRL/ExperimentOneData/}{github.com/JordanRL/ExperimentOneData}

This raw data includes an \texttt{experiment\_config.json} file for each strategy that contains the recorded environment information of the environment that generated the data. It was run on RunPod cloud computing resources, using an RTX 4090 pod. It ran in a template that was generated from the Dockerfile which is contained in the code repository.

The exact commands used, and the exact sequence they were used in, are explained in the \texttt{README.md} of the repository.

The code is written to generalize to other experiments, and is released under an MIT license. The code contains all of the metric collection processes, and also contains details on how another researcher could add their own additional metrics if so desired.

The author is available to other researchers who wish to utilize the framework for their own experiments to answer questions and provide limited support as time allows.

\subsubsection{Deterministic Behavior}

The framework supports deterministic, bit-for-bit reproduction of machine learning experiments out of the box, and guarantees that different strategies within the same experiment will receive the same initialization.

This includes isolation of the code that is executed to compute and collect metrics. The framework clones the entire training state, including the model, optimizer, and RNG state, prior to every point where instrumentation is run, then automatically restores that state prior to the training loop resuming. This ensures that any type of instrumentation could be added without affecting the deterministic behavior of the framework.

Reproduction with the same exact environment should result in bit-for-bit identical results. Reproduction with different environments, such as different hardware, should result in statistically similar results, but different exact data.

\subsubsection{Seed Sensitivity Sweeping}

All four strategies were trained with seed 199 for the instrumented runs presented in this paper. This seed was selected based on preliminary \stride{} runs as approximately median in generalization time.

To assess seed sensitivity, non-instrumented runs observing only test accuracy were conducted for \stride{} and \fixedrandom{} across seeds 31, 42, 199, 242, 555, and 9973. For \stride{}, five of six seeds generalized within 700 epochs; seed 555 did not fully generalize within this budget, reaching approximately 20--25\% test accuracy before stalling. Notably, $F = 101$ emerged as the peak embedding frequency on every seed tested, including seed 555 and despite each seed selecting a different random subset of 300{,}000 training pairs. Seed 555's failure was not in identifying the fundamental; it was in extending the harmonic series beyond the first few frequencies, suggesting that the initialization geometry at this seed was unfavorable for completing the Fourier construction rather than for beginning it. For \fixedrandom{}, all six seeds generalized within 700 epochs.

The \random{} and \target{} strategies were not tested across additional seeds, as their outcomes (memorization without generalization and failure to learn, respectively) are consistent with established results for IID and adversarial ordering in this domain~\cite{power2022grokking,shumailov2021manipulating}. Additionally, as this work was self-funded, there was a question of where the available resources would be best spent.

\subsubsection{Instrumentation and Metrics}

To properly interpret and understand the results presented in this paper, we utilize a variety of instrumentation and metrics. While some of these are commonly used in both research and industry, other techniques we've employed bear further explanation. A full accounting of all collected metrics are available in Appendix~\ref{apdx:hook-metrics}.

\paragraph{Counterfactual Decomposition.} To generate this metric, we checkpoint the model and run K=3 independently shuffled epochs from the same parameter state. The mean gradient across these shuffled runs approximates the order-independent gradient. The \textbf{content component} is extracted by projecting the observed gradient onto the calculated order-independent gradient. The \textbf{ordering component} is defined as the residual between the gradient from the actual ordered epoch and this content component, isolating the contribution of sequential structure to the training signal. For further details, see Appendix~\ref{apdx:counterfactual}.

\paragraph{Gradient Projection to Solution.} To assess whether training dynamics move the model toward its final converged state, we compute the cosine similarity between the negated gradient $-\nabla L$ (the descent direction) and the displacement vector $\theta_{\text{ref}} - \theta_{\text{current}}$, where $\theta_{\text{ref}}$ is the fully trained model's parameters. A positive value indicates the gradient points toward the solution; a negative value indicates the optimizer must navigate around regions of high curvature rather than proceeding directly. We apply this projection both to the full gradient and to the ordering and content components separately, enabling direct comparison of which component is more solution-aligned. See Appendix~\ref{apdx:gradient_projection} for interpretation caveats.

\paragraph{Hessian Entanglement Measurement.} The entanglement term $\eta H_B \cdot \nabla L_A$---the mechanism by which batch ordering enters the gradient (Section~\ref{sec:theory})---can be measured directly at each training step via finite-difference Hessian-vector products. We capture the previous batch's gradient before it is cleared, then compute the Hessian-vector product after the subsequent forward-backward pass. This yields per-step estimates of the entanglement norm, its fraction of the observed gradient energy, and its directional coherence across consecutive steps. See Appendix~\ref{apdx:hessian} for the full procedure and derived metrics.

\section{Experimental Results}
\label{sec:experiment}

\subsection{Results}
\label{sec:experiment-results}

\subsubsection{Generalization}
\label{sec:generalization}

\begin{figure}[!ht]
    \centering
    \includegraphics[width=1\linewidth]{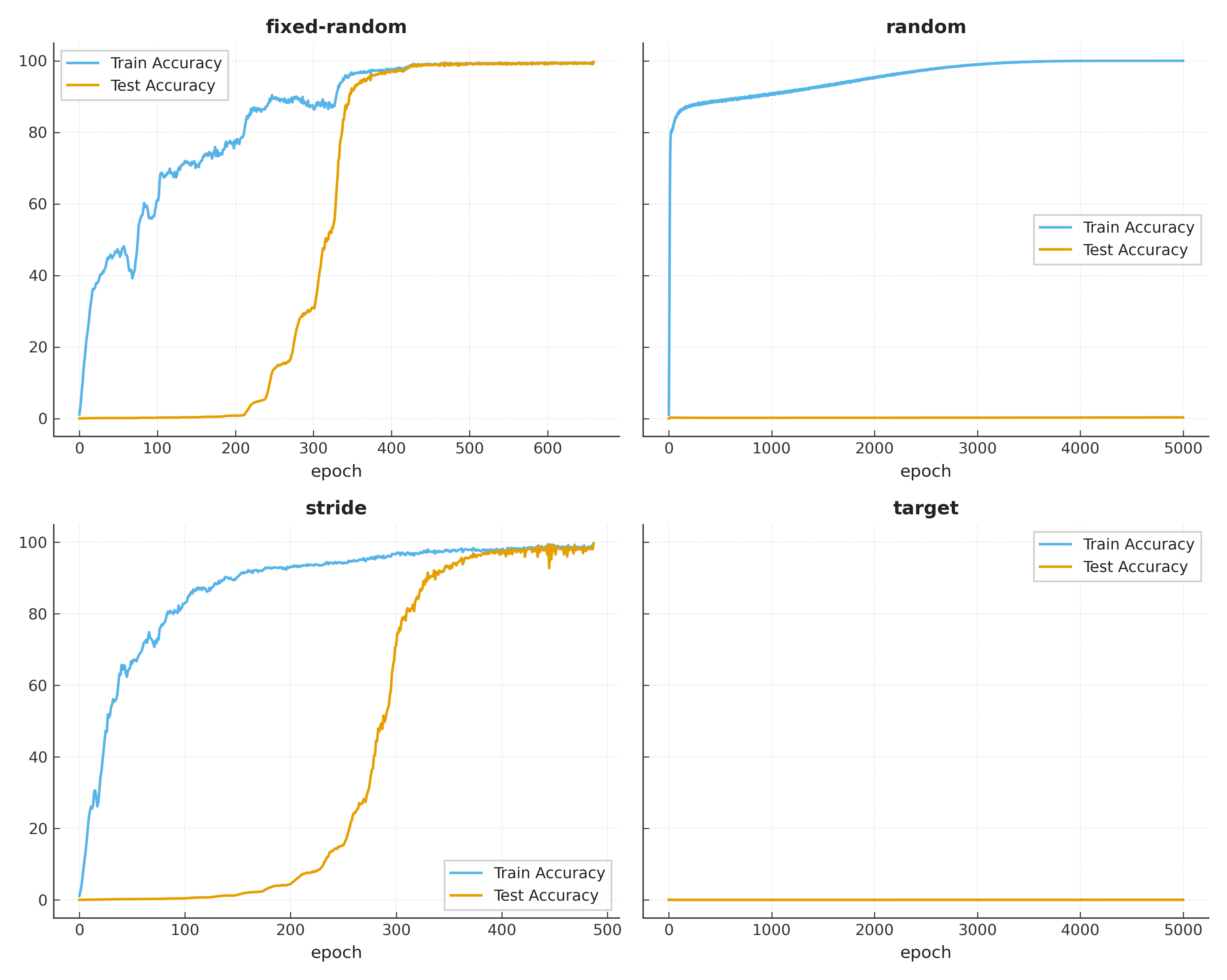}
    \caption{The \stride{} and \fixedrandom{} strategies generalize quickly; the \random{} strategy memorizes but is unable to generalize within the compute budget; the \target{} strategy fails to either generalize or memorize, never improving beyond chance-level performance.}
    \label{fig:accuracy}
\end{figure}

The four ordering strategies produce qualitatively different generalization outcomes from identical data and compute.

The \stride{} model achieves 99.5\% test accuracy by epoch 487. Train accuracy rises rapidly, reaching 90\% by approximately epoch 140, while test accuracy remains below 5\% until approximately epoch 200 before accelerating sharply. The peak train-test gap of approximately 90\% occurs around epoch 160. Despite this gap, the model is building the generalizing representation from the earliest epochs: by epoch 3, the peak frequency in the embedding weight spectrum is $F = 101$, the fundamental of the final generalizing solution (Section~\ref{sec:spectral}). The mechanistic significance of this gap is discussed below.

The \fixedrandom{} model achieves 99.5\% test accuracy by epoch 659, following a similar trajectory: rapid train accuracy growth, a delayed test accuracy takeoff (peak train-test gap of approximately 83\% around epoch 220), and an acceleration phase that brings both to convergence. Like \stride{}, the model orients toward its generalizing Fourier representation within the first ten epochs, with the peak embedding frequency locking to $F = 2205$ by epoch 7. This result is striking: the \fixedrandom{} ordering contains no task-specific structure whatsoever. It is a random permutation held constant across epochs. The role of structure is discussed in Section~\ref{sec:fixed-random}.

The \random{} model achieves 0.30\% test accuracy after 5{,}000 epochs. This is above the chance-level performance of 0.01\% for $p = 9973$, however the model does not display characteristics of generalization. It memorizes the training set completely but does not generalize within the training budget. This is the standard grokking regime: under IID ordering, generalization on modular arithmetic requires thousands of additional epochs beyond memorization.

The \target{} model fails to achieve meaningful train or test accuracy. The ordering signal is both strong and anti-aligned with the task's representational requirements, destabilizing optimization as predicted by the dose-productivity framework of Section~\ref{sec:theory}. See Section~\ref{sec:target-failure} for analysis.

\paragraph{The train-test gap under structured ordering.} The train-test gap in the fixed-ordering strategies might appear similar to the memorization phase observed in standard grokking, but the underlying mechanism is different. In standard grokking, the model first constructs a memorization table and then gradually replaces it with a generalizing circuit~\cite{nanda2023progress}. Under \stride{} and \fixedrandom{}, the spectral evidence shows that the model is constructing the generalizing representation from the start. The peak embedding frequency locks to the solution's fundamental within the first few epochs, and subsequent harmonics emerge sequentially as training progresses (Figure~\ref{fig:frequency_power}).

What the models lack during the gap period is not the correct representation but sufficient spectral resolution: a small number of harmonics can fit the sparse training points (which occupy only 0.3\% of the input space) while remaining too coarse to interpolate correctly across the full cyclic group. Each subsequent harmonic refines this resolution, and test accuracy rises in lockstep. The train-test gap therefore reflects an underdetermined Fourier interpolation, not a memorization table being gradually replaced by a generalizing circuit. The model is building toward a Fourier solution of the kind characterized analytically by Gromov~\cite{gromov2023grokking}, but has not yet accumulated enough harmonics to resolve the full cyclic group.

\subsubsection{Spectral Analysis}
\label{sec:spectral}

\begin{figure}[!ht]
    \centering
    \includegraphics[width=1\linewidth]{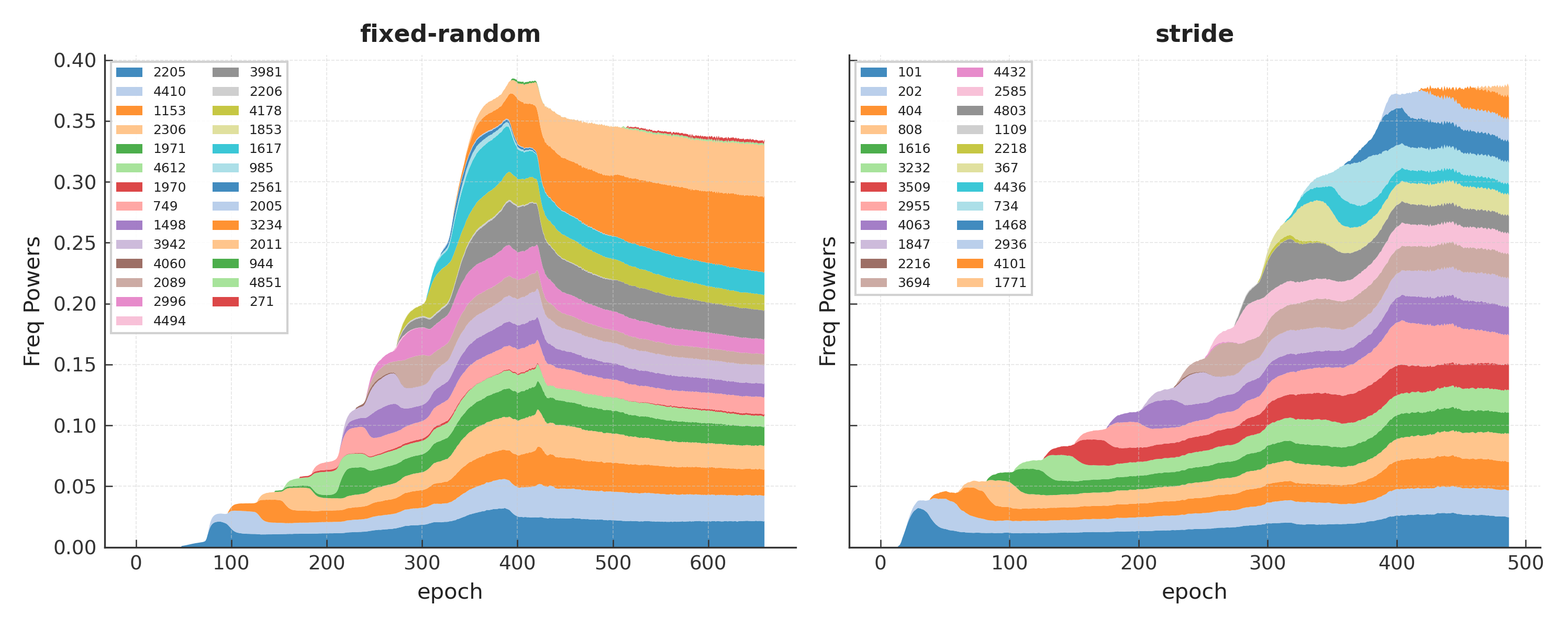}
    \caption{Both the \stride{} and \fixedrandom{} strategies accumulate frequency power early in training, and both end training with a similar amount of power concentrated in significant frequencies. The \stride{} strategy organizes much earlier than the \fixedrandom{} strategy, but the \fixedrandom{} achieve a higher peak concentration of frequency power.}
    \label{fig:frequency_power}
\end{figure}

Gromov~\cite{gromov2023grokking} showed that modular arithmetic can be solved by Fourier representations over the cyclic group, and provided analytical expressions for network weights that achieve 100\% accuracy. Fourier analysis of the \stride{} model's embedding weights reveals that it constructs such a representation, but rooted in a harmonic series whose fundamental frequency $F = 101$ is determined by the ordering rather than by the task alone. This frequency is the Fourier dual of the sort stride $s = 99$ over the cyclic group $\mathbb{Z}_{9973}$. The precise relationship is $F = \lfloor p / s \rceil = \lfloor 9973 / 99 \rceil = 101$, which corresponds to the natural sampling frequency of the stride pattern in the Fourier domain. This relationship between the stride value $s$ and the fundamental frequency $F$ was confirmed as described in Appendix~\ref{apdx:stride-frequency}.

The harmonic series follows a doubling pattern with Nyquist folding at $p/2 = 4986.5$:
$$101, \; 202, \; 404, \; 808, \; 1616, \; 3232, \; 6464 \to 3509, \; \ldots$$

where $6464 \bmod 9973 = 6464$ folds to $9973 - 6464 = 3509$ upon exceeding the Nyquist frequency. By the final epoch, 40 frequencies achieve power more than $10\times$ above uniform baseline, and the 14 highest-power frequencies are all predicted harmonics of $F = 101$.

Frequency $F = 101$ becomes the peak frequency in the embedding spectrum by epoch 3 and crosses the $10\times$ significance threshold by epoch 14. Subsequent harmonics emerge sequentially: $F = 202$ at epoch 21, $F = 404$ at epoch 39, $F = 808$ at epoch 61, and so on at roughly regular intervals (Figure~\ref{fig:frequency_power}). This sequential harmonic construction is visible in the significant frequency power plot, which shows the stacked power of all significant frequencies across training.

The \fixedrandom{} model exhibits the same qualitative pattern with different frequencies. Its peak embedding frequency locks to $F = 2205$ by epoch 7, the initial spectral dominant of its particular random permutation, and subsequent frequencies emerge sequentially from there. By the final epoch, 39 frequencies exceed the $10\times$ significance threshold, with $F = 3234$ having surpassed $F = 2205$ as the highest-power frequency. Critically, the frequencies that emerge bear no relationship to the stride harmonic series; instead, they reflect the spectral structure of the specific random permutation used. The model extracts whatever periodic structure exists in its fixed ordering and constructs the corresponding Fourier basis, with the relative power of different frequencies shifting as the representation matures.

\subsubsection{Spectral Entropy}
\label{sec:entropy}

\begin{figure}[!ht]
    \centering
    \includegraphics[width=\linewidth]{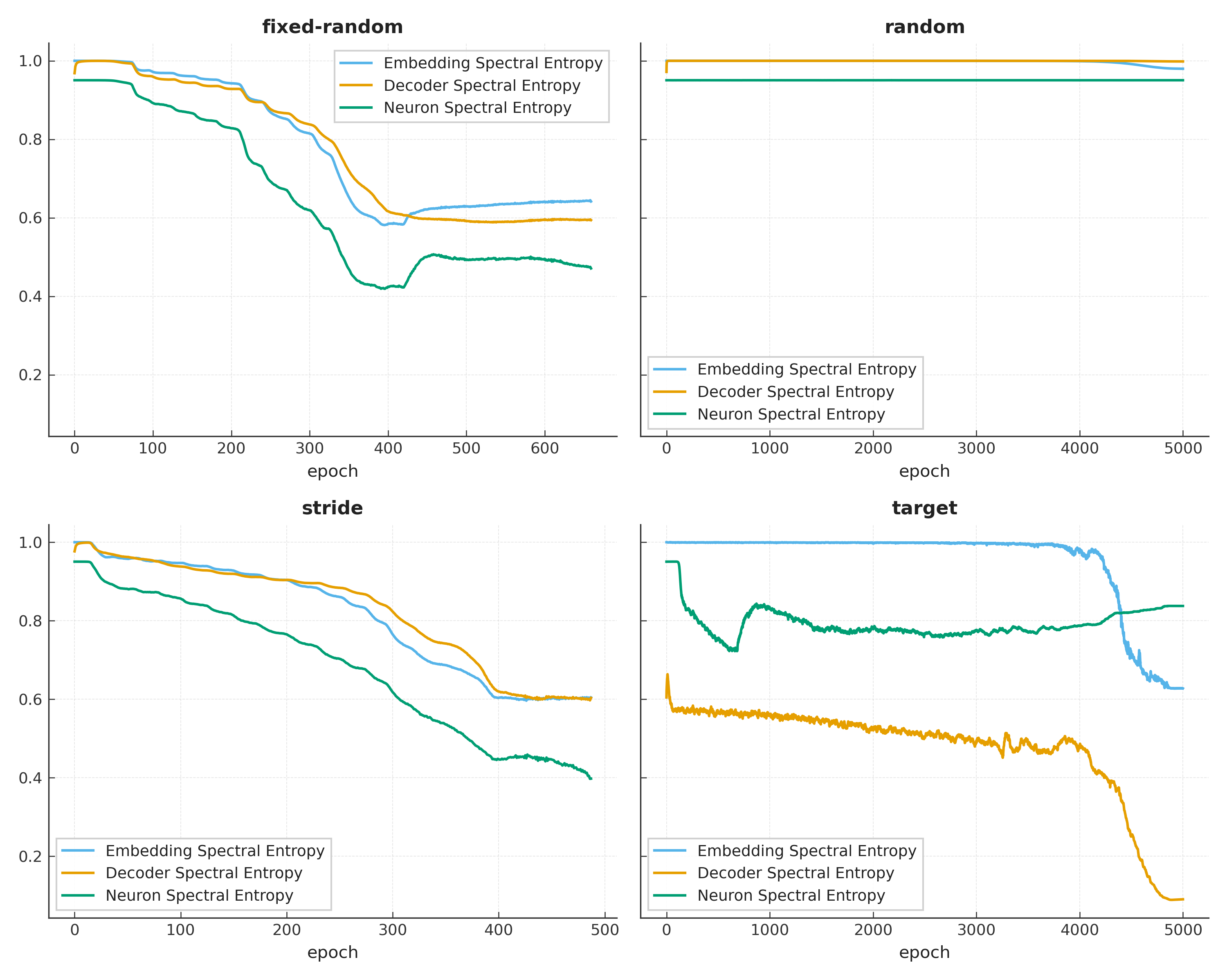}
    \caption{The \texttt{Embedding Spectral Entropy} (blue), \texttt{Decoder Spectral Entropy} (orange), and \texttt{Neuron Spectral Entropy} (green) for all four strategies. This measures the amount of uniformity in weight distributions, with 1.0 being perfectly uniform, and 0.0 being maximally concentrated.}
    \label{fig:spectral_entropy}
\end{figure}

We track spectral entropy across three levels: the embedding weight matrix, the decoder weight matrix, and per-neuron within the embedding. For a normalized power spectrum $P = \{P_k\}$ over frequencies $k$, spectral entropy is defined as
\[
  S = \frac{H(P)}{\log p} = \frac{-\sum_k P_k \log P_k}{\log p}
\]
where $p$ is the prime modulus. This normalization ensures $S \in [0, 1]$: $S = 1$ when power is uniformly distributed across all frequencies, and $S \to 0$ when power is concentrated on a single frequency. Figure~\ref{fig:spectral_entropy} shows all three measures for each strategy.

In the generalizing strategies, all three entropy measures decline together. The \stride{} model's embedding spectral entropy declines from near 1.0 to approximately 0.60, with a minimum of 0.60 during the most active phase of harmonic construction. The decoder and neuron-level entropies follow similar trajectories, with neuron-level entropy declining fastest, reaching approximately 0.40 before recovering. The \fixedrandom{} model shows a nearly identical pattern with slightly deeper minima: embedding to 0.58, neuron-level to 0.42. That neuron-level concentration leads aggregate concentration suggests spectral organization begins at the level of individual neurons before becoming visible in the full embedding spectrum.

The \random{} model shows almost no spectral concentration at any level. All three entropy measures remain almost flat near their initialization values throughout 5{,}000 epochs.

The \target{} model shows a striking dissociation between components. The decoder spectral entropy drops rapidly and remains low throughout training, indicating strong spectral organization in the output layer. The embedding, by contrast, remains near 1.0 for the first 4{,}000 epochs before a sharp late-training decline. The ordering signal organizes the output layer while the input layer fails to form usable representations until a late collapse driven by weight decay. This dissociation is analyzed further in Section~\ref{sec:target-failure}.

\subsubsection{The \target{} Failure Mode}
\label{sec:target-failure}

The \target{} model confirms the dose-productivity prediction from Section~\ref{sec:theory}. After 5{,}000 epochs, \target{} achieves 0.02\% train and 0.01\% test accuracy, at chance level ($1/p = 0.01\%$). Its loss falls to 7.36 by epoch 5{,}000, below the initialization value of 9.22 ($\approx \ln p$, maximum entropy over the output space), indicating some internal reorganization despite chance-level accuracy. But the model does not generalize. The ordering fraction of gradient norm averages 89\%, compared to 87\% for \random{}, and the cosine similarity between consecutive epoch gradients spans $-0.97$ to $+0.96$ (versus $-0.21$ to $+0.50$ under \random{}), with approximately 70\% of pairs anti-correlated. The ordering channel is powerfully active, but because \target{} sorts by output value, each batch spans only ${\sim}8$--9 consecutive output classes and the target distribution shifts completely between adjacent batches, producing a self-contradictory signal.

Despite this failure, \target{} drives the most dramatic spectral reorganization of any strategy, but only in the decoder. The decoder's spectral entropy drops from 0.62 after the first epoch to 0.09 by epoch 5{,}000, more concentrated than either generalizing strategy's decoder (\stride{}: 0.60; \fixedrandom{}: 0.59). The embedding tells the opposite story: its spectral entropy remains near 1.0 until a sharp late-training decline driven by weight-decay collapse (Figure~\ref{fig:spectral_entropy}), with no organized spectral structure emerging. The ordering signal organizes the output layer while the input layer never forms usable representations.

\target{} thus demonstrates that the ordering channel's outcome depends on two interacting factors: signal strength and task-alignment. The same mechanism produces qualitatively different outcomes across the four strategies: incoherent (\random{}), coherent and task-aligned (\stride{}, \fixedrandom{}), and coherent but anti-aligned (\target{}). Critically, \target{}'s failure is not simply that the ordering signal is too strong. The signal is \emph{anti-coherent}: it is structured enough to coherently organize the decoder (more than either generalizing strategy), but its structure is self-contradictory with respect to the input representation the task requires. The loss falling below its initialization value (7.36 vs.\ 9.22) confirms that the ordering is driving real internal reorganization, not merely adding noise. The result is a model that builds downstream structure (organized decoder) without upstream representations (collapsed embedding), because the ordering coherently drives the former while actively preventing the latter. Additional analysis of capacity allocation and matched-norm spectral comparisons confirming that the spectral effect is not solely attributable to weight-decay collapse is provided in Appendix~\ref{apdx:target-detail}.

\subsubsection{\fixedrandom{} Generalization}
\label{sec:fixed-random}

The \fixedrandom{} strategy was introduced as a control to distinguish the contribution of ordering \emph{consistency} (the same sequence repeated across epochs) from ordering \emph{structure} (alignment between the sequence and the task's algebraic properties). The result is more nuanced than a simple structure/no-structure dichotomy: \fixedrandom{} does possess task-relevant structure, but of a qualitatively weaker kind than \stride{}, and the ordering channel amplifies this weak structure into a signal sufficient for generalization.

\paragraph{Minimal structure is sufficient.} Any permutation of elements of the cyclic group $\mathbb{Z}_p$ possesses spectral content at frequencies that correspond to valid Fourier bases for the group operation. A fixed random permutation therefore provides entanglement terms whose spectral content is task-relevant, not because the permutation was designed to be, but because the task's algebraic structure ensures that virtually any consistent ordering over $\mathbb{Z}_p$ carries usable signal. \fixedrandom{} exploits this: where \stride{} builds a harmonic series rooted in $F = 101$ (the Fourier dual of the stride), \fixedrandom{} initially organizes around $F = 2205$, the first spectral dominant of its particular random permutation, before building out a broader frequency basis. The model extracts whatever periodic structure exists in its fixed ordering and constructs the corresponding Fourier basis. The critical difference from \stride{} is not the presence or absence of structure, but its strength: \stride{}'s structure is concentrated in a single harmonic series with clear number-theoretic meaning, while \fixedrandom{}'s is diffuse across many frequencies with no deliberate organization. Consistency is necessary (it prevents the destructive interference that renders \random{}'s ordering signal incoherent), but what makes \fixedrandom{} work is that the consistent ordering happens to carry task-relevant spectral content. In a domain where the generalizing representation bears no relationship to the spectral properties of data orderings, consistency alone might prevent destructive interference without producing constructive interference toward a useful solution.

\paragraph{Signal amplification.} The ordering channel's amplification machinery makes even this weak structure sufficient. The counterfactual decomposition shows that the ordering fraction of gradient norm under \fixedrandom{} averages 84\%, nearly identical to \stride{}'s 83\%. The Hessian entanglement energy ratio averages $860$--$900\times$ the observed gradient energy, confirming that the curvature-mediated interaction between consecutive batches is enormous regardless of whether the ordering is deliberately structured. Adam further amplifies the signal: the mean amplification ratio $\|\Delta\theta_{\mathrm{Adam}}\| / \|\eta \nabla L\|$ is 246$\times$ for \fixedrandom{}, compared to 175$\times$ for \stride{}. The ordering fraction profile across layers is also nearly indistinguishable between the two strategies: feedforward and decoder layers highest ($\sim$0.9), embeddings near 0.85, attention layers near 0.8, and norm layers low ($\sim$0.4). The network routes the ordering signal through the same architectural pathway regardless of the ordering's content, and the combined Hessian and Adam amplification transforms a diffuse spectral signal into one strong enough to drive generalization.

\begin{figure}[!htb]
    \centering
    \includegraphics[width=1\linewidth]{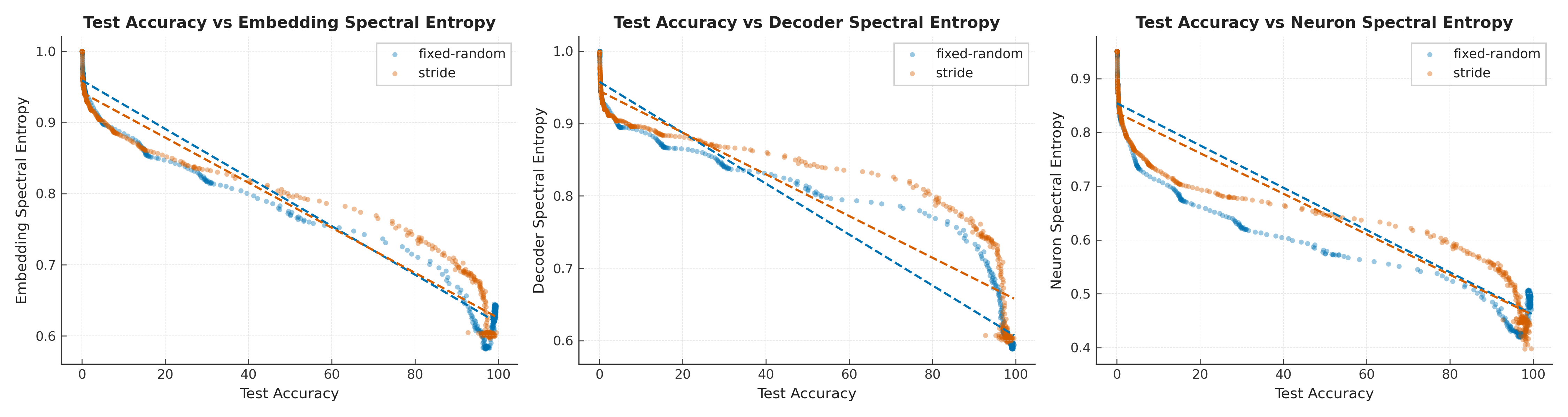}
    \caption{Validation accuracy plotted against embedding, decoder, and neuron spectral entropy for the two generalizing strategies. Both follow nearly identical curves despite building different frequency bases: generalization is a function of spectral concentration itself, independent of which frequencies the model concentrates into.}
    \label{fig:acc_vs_entropy}
\end{figure}

\paragraph{Regularization and stability.} Plotting validation accuracy against spectral entropy (Figure~\ref{fig:acc_vs_entropy}) reveals that \stride{} and \fixedrandom{} follow nearly identical curves despite building different frequency bases at different speeds. The relationship between generalization and spectral concentration is strategy-independent: the model's degree of generalization is a direct function of how concentrated its spectral energy is, regardless of which frequencies it concentrates into.

The two strategies diverge, however, in late-training stability. Over the final 100 epochs, \fixedrandom{}'s test accuracy has a standard deviation of 0.14\% (minimum 98.96\%), while \stride{}'s has a standard deviation of 0.94\% with excursions as low as 92.7\%. \fixedrandom{}'s distributed spectral representation, spread across more frequencies with no single dominant harmonic, appears to produce a more robust interpolation. The Adam optimizer compensates for the diffuse ordering signal by applying more non-uniform per-parameter learning rates (coefficient of variation 6.6 for \fixedrandom{} vs.\ 5.7 for \stride{}), selectively amplifying parameters relevant to the distributed Fourier representation. The result is a representation that requires more optimizer intervention to construct but is more stable once established.

This regularization effect may also explain why \fixedrandom{} converges faster than \stride{} on some seeds despite converging more slowly on this seed. Where \stride{}'s concentrated harmonic series creates strong spectral peaks that can overshoot during training, \fixedrandom{}'s distributed representation provides a smoother optimization landscape in the relevant parameter subspace, and the seed sensitivity sweeps (Section~\ref{sec:reproduction-steps}) show all six \fixedrandom{} seeds generalizing versus five of six for \stride{}.

\subsection{Counterfactual Gradient Decomposition}
\label{sec:counterfactual}

To directly measure the ordering and content components of the gradient, we introduce a counterfactual decomposition. At each measurement epoch, we:

\begin{enumerate}
    \item Record the actual mean gradient from the ordered epoch:
    \[
      g_{\mathrm{actual}} = \frac{1}{N}\sum_{i=1}^{N} \nabla_{\theta_i} L(B_i, \theta_i)
    \]
    where $B_i$ are the $N$ batches in their training-time order and $\theta_i$ are the parameters at step $i$ (with $\theta_1 = \theta$, the pre-epoch weights).
    \item From $\theta$, run three independently shuffled epochs of the same dataset, each starting from the same pre-epoch weights. Each shuffled epoch $k$ produces a mean gradient computed identically to $g_{\mathrm{actual}}$ but under a random permutation $\pi_k$:
    \[
      g^{(k)}_{\mathrm{shuffled}} = \frac{1}{N}\sum_{i=1}^{N} \nabla_{\theta^{(k)}_i} L\bigl(B_{\pi_k(i)},\, \theta^{(k)}_i\bigr)
    \]
    where $\theta^{(k)}_1 = \theta$ for all $k$.
  \item Compute the \textbf{content direction} as the normalized mean gradient
    across the $K$ shuffled runs:
    \[
      \bar{g}_{\mathrm{shuffled}} = \frac{1}{K}\sum_{k=1}^{K} g^{(k)}_{\mathrm{shuffled}}, \qquad
      \hat{g}_{\mathrm{cf}} = \frac{\bar{g}_{\mathrm{shuffled}}}{\|\bar{g}_{\mathrm{shuffled}}\|}
    \]
  \item Define the \textbf{content component} as the projection of the actual 
    gradient onto this direction, and the \textbf{ordering component} as the 
    orthogonal residual:
    \begin{align}
      g_{\mathrm{content}} &= \bigl(\hat{g}_{\mathrm{cf}} \cdot g_{\mathrm{actual}}\bigr)\, \hat{g}_{\mathrm{cf}} \label{eq:content} \\
      g_{\mathrm{ordering}} &= g_{\mathrm{actual}} - g_{\mathrm{content}} \label{eq:ordering}
    \end{align}
    By construction, $g_{\mathrm{content}} \perp g_{\mathrm{ordering}}$, so the 
    decomposition satisfies
    \[
      \|g_{\mathrm{actual}}\|^2 = \|g_{\mathrm{content}}\|^2 + \|g_{\mathrm{ordering}}\|^2
    \]
    which ensures the two components partition the total gradient energy 
    without double-counting.
\end{enumerate}

The three shuffled runs provide variance estimation to confirm that the content mean is stable. For further details, see Appendix~\ref{apdx:counterfactual}.

\subsubsection{Norm Decomposition}

\begin{figure}[!htb]
    \centering
    \includegraphics[width=\linewidth]{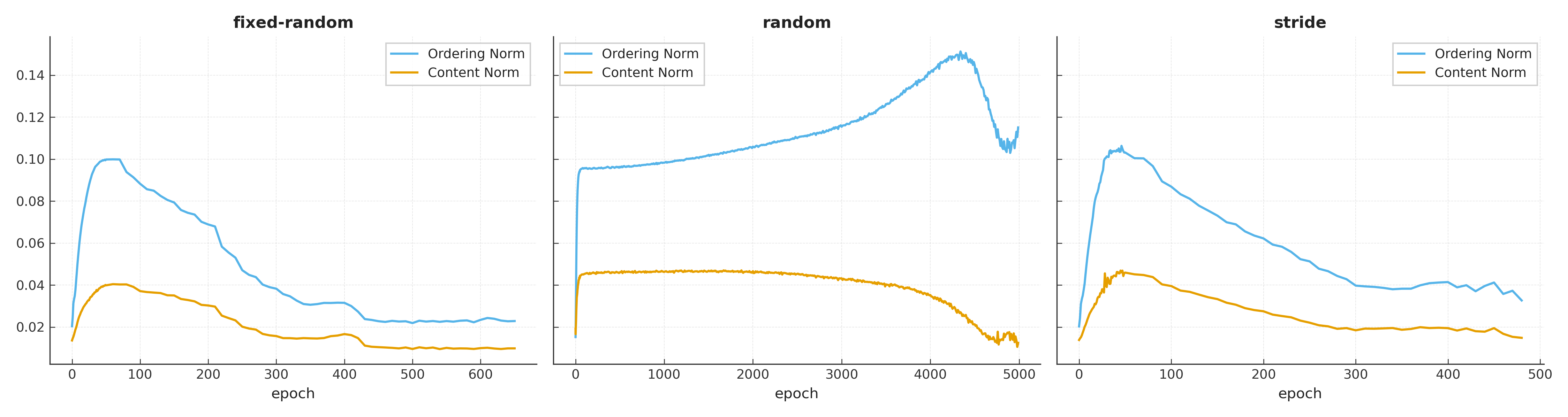}
    \caption{L2 norm of the ordering (blue) and content (orange) gradient components across training for \fixedrandom{}, \random{}, and \stride{}. Under the fixed-ordering strategies, the ordering norm peaks early and then declines as the model absorbs the coherent signal. Under \random{}, the ordering norm remains persistently large but incoherent, at approximately $2.8\times$ the content norm throughout training.}
    \label{fig:norm_decomp}
\end{figure}

Figure~\ref{fig:norm_decomp} shows the L2 norm of the content and ordering components across training.

In the \stride{} model, the ordering norm peaks at 0.106 around epoch 46 and then declines steadily. This trajectory reflects the model consuming the coherent ordering signal: the structured ordering provides a strong directional push early in training, and as the model builds the corresponding representation, the ordering no longer surprises it and the ordering component diminishes. The \fixedrandom{} model follows a similar trajectory, with its ordering norm peaking at a comparable magnitude around epoch 60.

This consumption is not uniform across the network. At the layer level, the ordering fraction decline is concentrated in the deep feedforward layers (Layer~1), where it is the strongest leading indicator of generalization for both strategies: the ordering fraction at \texttt{layers.1.linear1} correlates with validation accuracy at $r = 0.98$ for \stride{} and $r = 0.99$ for \fixedrandom{}, with the rate of decline accelerating in the 50 epochs before generalization onset. The deep feedforward layers appear to be the primary site where the ordering-derived representation is consolidated, and their absorption of the ordering signal tracks the progressive refinement of the generalizing representation.

In the \random{} model, the ordering norm remains at approximately 0.11, consistently $\sim$2.8$\times$ the content norm, across all 5{,}000 epochs. The ordering component is persistently large but incoherent: it represents the random per-epoch displacement attributable to the particular shuffle used, pointing in a different high-dimensional direction each epoch with no cumulative effect. GraB~\cite{lu2022grab} addresses this incoherence by minimizing gradient sum discrepancy, which in this framework corresponds to reducing the magnitude of the ordering component under IID conditions. The fixed-ordering strategies demonstrate a complementary approach: rather than reducing the ordering component, they make it coherent and productive, converting displacement that would otherwise cancel into a directed learning signal.

As noted in Section~\ref{sec:signal-strength}, the magnitude of the ordering component should not be taken to imply that it is composed entirely of productive signal. The ordering component contains \emph{all} order-dependent gradient information, and even under highly structured ordering strategies it will contain significant noise alongside the learnable signal.

\subsubsection{Ordering-Content Alignment}
\label{sec:alignment}

The cosine similarity between the ordering and content components, $\cos(g_{\mathrm{ordering}},\, g_{\mathrm{content}})$, reveals whether the ordering channel carries independent information or merely amplifies the content signal.

\begin{figure}[!htb]
    \centering
    \includegraphics[width=\linewidth]{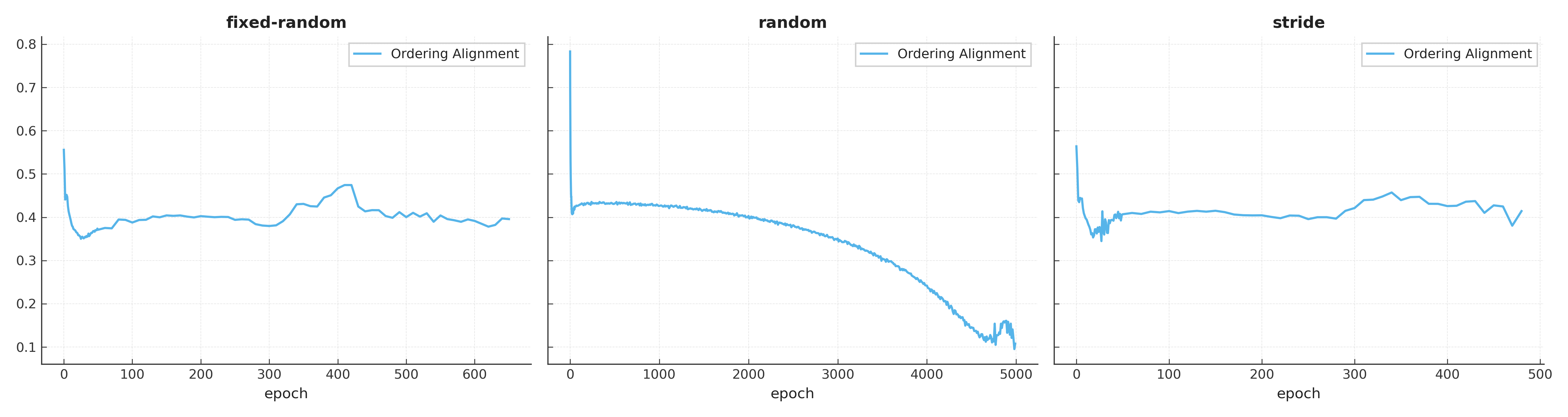}
    \caption{Cosine similarity between ordering and content gradient components across training. Both generalizing strategies show alignment dropping from $\sim$0.56 to $\sim$0.35 during early training, then hovering around 0.40 as both components converge toward the generalizing solution. Under \random{}, alignment begins similarly but collapses in late training as the model begins its slow approach toward grokking.}
    \label{fig:alignment}
\end{figure}

Under both generalizing strategies, the alignment begins at $\sim$0.56 and drops to approximately 0.35 during the first 30--60 epochs (Figure~\ref{fig:alignment}), the same window in which the ordering norm peaks (Figure~\ref{fig:norm_decomp}). When the ordering signal is strongest, it is also most orthogonal to the content signal: it is delivering information that the content alone does not provide. As the model absorbs this information, the alignment partially recovers during mid-training because both components now point toward the same generalizing solution. The mean alignment of $\sim$0.40 across training indicates that the ordering channel carries partially independent information throughout.

Under \random{}, the alignment begins high ($\sim$0.78) and stabilizes around 0.33 for most of training, before collapsing toward 0.13 in the final epochs as the model begins its slow approach toward grokking. The late-training divergence between the ordering and content components under \random{} suggests that the incipient generalization process involves increasingly different gradient structure between shuffled and ordered passes.

\subsection{Mechanistic Interpretation}
\label{sec:mechanistic-interpretation}

The frequency $F = 101$ can only emerge if the model simultaneously encodes three facts: (1) the task is cyclic with period $p = 9973$, (2) the group operation is addition, and (3) the training data is strided with period $s = 99$. Facts (1) and (2) are properties of the data content. Fact (3) exists nowhere in any individual training example; it is a property of the sequential relationships between examples. The model's construction of a representation rooted in $F = 101$ is therefore direct evidence that it has extracted and utilized information from the ordering channel.

The subsequent harmonic series $101 \times 2^n \bmod p$ (with Nyquist folding) further confirms that the model builds its representation by iteratively refining the ordering-derived fundamental, generating higher harmonics to increase the resolution of its internal representation of the cyclic group structure. The endpoint of this construction is a Fourier solution of the kind Gromov~\cite{gromov2023grokking} characterized analytically, but the specific frequency basis, the path of construction, and the rate of convergence are all determined by the ordering channel.

The fact that this initial significant frequency, with the underlying learning it implies, first appears at epoch 3 provides conclusive evidence that the ordering signal was both strong and coherent from the very beginning of training.

The robustness of this frequency identification strengthens the mechanistic claim. Across all six seeds tested for \stride{} (Section~\ref{sec:reproduction-steps}), $F = 101$ emerged as the peak embedding frequency despite each seed determining both a different parameter initialization and a different random sample of 300{,}000 training pairs. The frequency is not an artifact of a particular initialization geometry or dataset composition; it is a deterministic consequence of the stride ordering's structure over $\mathbb{Z}_p$. Even seed 555, which stalled at 20--25\% test accuracy and could not complete the harmonic series, correctly identified $F = 101$ as the fundamental. The ordering channel's initial signal is robust; what varies across seeds is whether the optimization trajectory can sustain the harmonic construction to completion.

\subsection{Optimizer Interaction with the Ordering Channel}
\label{sec:adam-empirical}

Section~\ref{sec:adam-theory} argued that the Hessian-displacement mechanism persists under any optimizer, with Adam modulating the ordering channel through amplified displacement, momentum integration, and selective parameter scaling. We validate these predictions using per-step optimizer diagnostics recorded throughout training.

The most direct confirmation is that the ordering channel survives Adam's transformation of the gradient. Adam redirects 95--99\% of each update away from the raw gradient direction (update deflection $>0.95$ for all strategies), yet the counterfactual decomposition shows ordering fractions of 83--84\% for the fixed-ordering strategies, comparable to what a pure SGD analysis would predict. The non-commutativity mechanism does not depend on the update being aligned with the raw gradient; it depends only on the update producing a nonzero displacement in a curved landscape, which Adam emphatically does. The Hessian entanglement energy ratio, which measures $\|H_B \cdot \Delta\theta\|^2 / \|\nabla L_B\|^2$, averages $860$--$900\times$ for the generalizing strategies, confirming that curvature-mediated interaction between consecutive batches dominates the per-step gradient under Adam just as the theory predicts.

Adam further differentiates the strategies through its adaptive scaling. The mean amplification ratio ($\|\Delta\theta_{\text{Adam}}\| / \|\eta \nabla L\|$) is 246$\times$ for \fixedrandom{} and 175$\times$ for \stride{}, compared to 101$\times$ for \random{} and 58$\times$ for \target{}. The generalizing strategies receive substantially larger displacements, which in turn produce proportionally larger entanglement terms on subsequent steps. The per-parameter effective learning rate coefficient of variation tells a complementary story: \stride{} shows the most uniform adaptation (CV = 5.7), consistent with a concentrated ordering signal that produces consistent gradients across the relevant parameter subspace; \fixedrandom{} is moderately non-uniform (CV = 6.6), reflecting its more diffuse spectral signal; \random{} is the most non-uniform (CV = 10.6), as the incoherent ordering produces highly variable per-parameter gradient histories. Adam's adaptive scaling thus selectively amplifies the ordering signal in precisely the parameter subspace it targets, creating the feedback loop described in Section~\ref{sec:adam-theory}. Additional detail on per-strategy optimizer dynamics is provided in Appendix~\ref{apdx:adam-detail}.

\section{Discussion}
\label{sec:discussion}

\subsection{The Channel Is Always Active}
\label{sec:always-active}

The counterfactual gradient decomposition reveals that the ordering component accounts for the majority of gradient norm under \emph{all four} ordering strategies, not only the structured ones. The ordering fraction averages 83--87\% for \stride{}, \fixedrandom{}, and \random{}, and rises to 89\% for \target{} (Table~\ref{tab:dose-response}). Under \random{}, the ordering norm is consistently $\sim$2.8$\times$ the content norm across all 5{,}000 epochs. This means that in every standard IID training run, the majority of each epoch's cumulative gradient magnitude is attributable to the accidental ordering of examples in that epoch.

This finding challenges a widespread implicit assumption. The theoretical justification for IID shuffling is that it makes the \emph{expected} gradient equal to the true gradient, and practitioners have generally treated this as meaning that the cumulative gradient over an epoch is approximately equal to the content gradient. Our measurements show this is not the case. Each epoch's cumulative gradient is dominated by ordering-induced displacement; the content signal emerges only as the long-run average over many epochs, after the ordering contributions have canceled. The situation is analogous to a communications channel with a signal-to-noise ratio well below 1: the signal is recoverable given enough integration time, but any individual epoch's gradient is dominated by ordering effects. IID shuffling does not eliminate the ordering channel. It ensures that the channel carries incoherent noise rather than coherent signal, and it relies on statistical cancellation over many epochs to prevent that noise from accumulating into lasting parameter changes. The cancellation is effective in expectation, but it comes at a cost: the majority of each epoch's compute is spent on gradient displacement that must subsequently be undone by the cancellation of future epochs.

\subsection{Theoretical Predictions and Experimental Confirmation}
\label{sec:theory-experiment}

The theoretical framework of Section~\ref{sec:theory} generated six empirical predictions before any experiments were conducted. We summarize their status here, both to establish the framework's predictive power and to identify where it was incomplete.

\paragraph{Prediction 1: The ordering component is measurable via counterfactual comparison.} Confirmed. The counterfactual decomposition (Section~\ref{sec:counterfactual}) cleanly separates the gradient into ordering and content components with a stable content estimate ($K = 3$ shuffled runs sufficient; validated against $K = 4$; see Appendix~\ref{apdx:counterfactual}). The ordering component is not merely detectable but dominant: it accounts for 83--84\% of mean gradient norm under the fixed-ordering strategies and $\sim$87\% even under IID shuffling (Section~\ref{sec:alignment}).

\paragraph{Prediction 2: The ordering signal produces measurable spectral concentration.} Confirmed. The \stride{} model's embedding spectrum concentrates into a harmonic series rooted in $F = 101$, the Fourier dual of the stride, with 40 significant frequencies by convergence (Section~\ref{sec:spectral}). Spectral entropy drops from near 1.0 to 0.40 at the neuron level during the most active construction phase (Section~\ref{sec:entropy}). The \fixedrandom{} model shows analogous concentration initially organizing around $F = 2205$, with 39 significant frequencies by convergence. Neither non-generalizing strategy produces comparable spectral organization in the embedding.

\paragraph{Prediction 3: Strong, misaligned ordering prevents convergence.} Confirmed. The \target{} strategy produces ordering fractions of 89\% and anti-correlated consecutive gradients (70\% of pairs), driving the loss to 7.36 by epoch 5{,}000, well below the initialization value but at chance-level accuracy (Section~\ref{sec:target-failure}). The continuum predicted by the theory is realized across the four strategies, with outcomes determined by the interaction between signal strength and task-alignment.

\paragraph{Prediction 4: The ordering signal is strongest during early feature formation.} Confirmed. The ordering component norm peaks at epoch 46 for \stride{} and declines as the model absorbs the signal (Section~\ref{sec:counterfactual}). The fundamental frequency $F = 101$ is identifiable by epoch 3 and crosses the $10\times$ significance threshold by epoch 14 (Section~\ref{sec:spectral}), placing the critical learning period squarely within the first few dozen epochs, when the loss landscape curvature with respect to the forming Fourier features is highest.

\paragraph{Prediction 5: The ordering and content components are partially but not fully aligned.} Confirmed. The ordering-content cosine similarity averages $\sim$0.40 across training for both generalizing strategies, with a characteristic dip to $\sim$0.35 during the ordering norm peak (Section~\ref{sec:alignment}). The ordering channel carries partially independent information throughout training, and is most independent precisely when it is strongest.

\paragraph{Prediction 6: Adaptive optimizers amplify the entanglement beyond the SGD prediction.} Confirmed. The Adam amplification ratio averages 175--246$\times$ for the generalizing strategies (Section~\ref{sec:adam-empirical}), and the Hessian entanglement energy ratio averages $860$--$900\times$ the observed gradient energy, reflecting a geometry in which the entanglement and content terms are $\sim$30$\times$ larger than their residual (the observed gradient) and aligned to cosine similarity $>0.999$.

\paragraph{Where the theory was incomplete.} The framework correctly predicted the existence and qualitative behavior of the ordering channel but did not predict two features of the experimental results. First, the theory does not explain why \fixedrandom{} generalizes at a rate comparable to \stride{} rather than intermediate between \stride{} and \random{}. This outcome depends on a domain-specific property: modular arithmetic over $\mathbb{Z}_p$ ensures that any consistent permutation carries task-relevant spectral content, making \fixedrandom{} accidentally well-structured for this task. The theory predicts that consistency prevents destructive interference but does not predict whether the resulting coherent signal will be productive in a given domain. However, the practical success of GraB~\cite{lu2022grab} on natural language and image classification tasks suggests that productive orderings exist in domains well beyond cyclic groups, even when discovered by gradient-based heuristics rather than algebraic insight. Second, the theory does not predict the specific regularization benefit of distributed spectral representations: \fixedrandom{}'s late-training stability (test accuracy std of 0.14\% vs.\ \stride{}'s 0.94\% over the final 100 epochs) was not anticipated by the framework and may reflect properties of the optimizer-landscape interaction that the current theory does not capture.

\subsection{The Dose-Productivity Continuum}
\label{sec:dose-response}

The four ordering strategies realize a continuum predicted by the theoretical framework. The ordering component accounts for 83--89\% of each epoch's cumulative gradient norm across all four strategies (Table~\ref{tab:dose-response}); what differs is not the channel's magnitude but the interaction between two properties of the ordering signal: its \emph{dose} (the accumulated strength of the coherent signal over time) and its \emph{productivity} (its alignment with the task's representational requirements).

\begin{table}[!htb]
\centering
\caption{Summary of ordering channel characteristics across the four strategies. The ordering fraction is similar in all cases; the strategies differ in temporal coherence and outcome. Details on the characterization of "Coherent", "Incoherent", and "Anti-coherent" can be found in Appendix~\ref{apdx:autocorrelation-mean}}
\label{tab:dose-response}
\begin{tabular}{lcccc}
\toprule
& \textbf{Mean} & \textbf{Mean} & & \\
& \textbf{Ordering} & \textbf{Ordering-Content} & \textbf{Epoch-to-Epoch} & \textbf{Final Test} \\
\textbf{Strategy} & \textbf{Fraction} & \textbf{Alignment} & \textbf{Coherence} & \textbf{Accuracy} \\
\midrule
\random{}       & 87\% & 0.35 & Incoherent  & 0.30\% \\
\fixedrandom{}  & 84\% & 0.39 & Coherent    & 99.5\% \\
\stride{}       & 83\% & 0.41 & Coherent    & 99.5\% \\
\target{}       & 89\% & 0.25 & Anti-coherent & 0.01\% \\
\bottomrule
\end{tabular}
\end{table}

Under \random{}, the ordering is reshuffled every epoch. The entanglement terms are large at each step but point in a different high-dimensional direction each epoch, producing a random walk in parameter space that cancels over time. The content signal eventually drives generalization, but only on the timescale of thousands of epochs, because it must overcome the per-step ordering noise rather than being reinforced by it.

Under \stride{} and \fixedrandom{}, the same ordering repeats every epoch. The entanglement terms accumulate coherently: the same spectral bias is applied to the gradient at every epoch, and the optimizer integrates this consistent signal into a structured representation. The critical difference between the two is not coherence (both are coherent) but specificity: \stride{}'s signal is concentrated in a single harmonic series with clear number-theoretic meaning, while \fixedrandom{}'s is distributed across many frequencies. Both are productive in this domain because modular arithmetic ensures that any consistent permutation's spectral content is task-relevant.

Under \target{}, the ordering is consistent across epochs but self-contradictory within each epoch. Because \target{} sorts examples by output value, consecutive batches span disjoint output classes, and the gradient direction reverses between adjacent epochs (70\% of consecutive epoch pairs are anti-correlated). The entanglement terms are individually large, producing the highest ordering fraction of any strategy (89\%), and unlike the incoherent case (\random{}), they do not cancel over time. The result is an ordering signal that is anti-coherent: structured enough to coherently organize the decoder, but self-contradictory with respect to the input representation, overwhelming the content signal while actively driving the model toward a degenerate basin.

This continuum is a property of the entanglement mechanism itself, not of the specific task. The Hessian-gradient interaction $H_B \cdot \Delta\theta$ is present whenever consecutive training steps produce nonzero displacement in a curved landscape. The outcome depends on the interaction between dose and productivity: incoherent orderings deliver effectively zero accumulated dose, because their per-step contributions cancel; coherent orderings deliver sustained dose, and whether that dose is productive or destructive depends on its alignment with the task's representational requirements. These dynamics follow from the geometry of gradient-based optimization in non-convex landscapes and should apply to any training regime, though the specific thresholds separating productive from destructive ordering will depend on the task, architecture, and optimizer.

The key relationship between these axes is that the more productive the signal, the higher the dose the model can tolerate without being driven toward degenerate representations. \target{} is not destructive merely because its dose is high; it is destructive because high dose is combined with anti-productive structure, locking the model into organized but degenerate representations. \fixedrandom{} succeeds in part because its diffuse spectral structure delivers a lower effective dose than \stride{}'s concentrated harmonic series, and this lower dose coincides with its lower productivity: the undesigned permutation is accidentally task-relevant but imprecisely so. The lower dose affords the model more flexibility to accommodate the imprecise signal, contributing to \fixedrandom{}'s greater stability across seeds (six of six generalizing, versus five of six for \stride{}). Even within \stride{}, where productivity is high, the dose-productivity interaction is visible: seed 555 correctly identified the fundamental frequency $F = 101$ but stalled at 20--25\% accuracy, unable to complete the harmonic construction. The high dose that enables rapid generalization on most seeds locked this seed into a representational path that its initialization geometry could not sustain to completion. Stronger orderings offer greater potential acceleration but require correspondingly higher productivity, both in task-alignment and in compatibility with the model's current optimization landscape.

\subsection{Regime Dependence of the Ordering Channel}
\label{sec:regime-dependence}

The ordering channel is always active, but its practical importance depends on the adequacy of the content signal. Two controlled comparisons illustrate this relationship.

\paragraph{Data sparsity.} At $p = 97$ with 26.5\% data density, all three non-adversarial strategies generalize within comparable timeframes, with fixed orderings providing only a modest speedup ($\sim$20\%) over IID shuffling (Table~\ref{tab:small-p-wd-comparison}, $\lambda = 0.01$ column). This contrasts sharply with the $p = 9973$ experiment at 0.3\% data density, where the content signal alone is insufficient for generalization within any reasonable compute budget: \random{} fails to generalize after 5{,}000 epochs while \stride{} and \fixedrandom{} achieve 99.5\% test accuracy by epochs 487 and 659 respectively. The ordering channel determines whether generalization occurs at all only when the content signal is inadequate on its own.

\paragraph{Weight decay.} Table~\ref{tab:small-p-wd-comparison} presents a weight-decay ablation at $p = 97$ with 26.5\% data density, averaged over 5 seeds per strategy. All strategies generalize at all weight-decay values tested, but the ordering channel's relative contribution increases monotonically as weight decay decreases. At $\lambda = 0.1$, fixed orderings provide a $\sim$9\% speedup over \random{}; at $\lambda = 0.01$, the speedup grows to $\sim$32\%. Weight decay is known to drive the lazy-to-rich transition that enables content-signal-driven generalization~\cite{kumar2024grokking}. As this regularization weakens, the content signal degrades faster than the ordering signal: \random{} slows by a factor of 7$\times$ between $\lambda = 0.1$ and $\lambda = 0.01$, while \fixedrandom{} slows by only 5.2$\times$.

\begin{table}[!htb]
    \centering
    \begin{tabular}{r|ccc}
    \toprule
         & Epochs to & Epochs to & Epochs to \\
         & Finish & Finish & Finish \\
         Strategy & WD=0.1 & WD=0.05 & WD=0.01 \\
    \midrule
         \stride{} & 230 & 390 & 1375 \\
         \fixedrandom{} & 230 & 381 & 1206 \\
         \random{} & 253 & 437 & 1776 \\
    \bottomrule
    \end{tabular}
    \caption{Weight-decay ablation at $p = 97$, averaged over 5 seeds per strategy. AdamW optimizer; training set size 2{,}500 (26.5\% of problem space); embedding dimension 128; 1 layer; batch size 32. Fixed orderings degrade more gracefully than IID shuffling as regularization weakens.}
    \label{tab:small-p-wd-comparison}
\end{table}

\paragraph{A unified principle.} Both effects reflect the same underlying relationship: the ordering channel's practical importance scales with the inadequacy of the content signal. Data sparsity reduces the content signal by providing fewer examples from which to extract statistical regularities. Weak regularization reduces it by slowing the transition from lazy to rich learning regimes, prolonging the period during which the model relies on superficial memorization rather than structured feature acquisition. In both cases, the ordering channel partially compensates, providing a learning signal whose strength depends on temporal coherence rather than on data density or regularization. The ordering channel is not a substitute for content, but it is a complementary signal source whose relative contribution is greatest precisely when standard training conditions are least favorable. Rajput~\etal~\cite{rajput2022permutation} provide theoretical support for this regime dependence: they prove that the convergence gap between optimal and random data permutations ranges from exponential to nonexistent depending on the function class. In our framework, function classes where random ordering is already near-optimal correspond to regimes where the content signal is sufficient; classes where the ordering gap is exponential correspond to regimes where the content signal is inadequate and the ordering channel becomes the determining factor.

\subsection{Implications for Training Efficiency}
\label{sec:efficiency}

The compute implications of the ordering channel are stark. Under IID ordering, the \random{} model requires over 5{,}000 epochs to reach 0.30\% test accuracy. Under \stride{} and \fixedrandom{}, the same model architecture, data, and optimizer achieve 99.5\% test accuracy by epochs 487 and 659 respectively. The ordering component accounts for approximately 85\% of each epoch's cumulative gradient norm in both regimes. Under IID ordering, this 85\% is incoherent noise that cancels over many epochs, contributing only a random walk in parameter space; under fixed ordering, it is a coherent signal that constructively drives the model toward a generalizing representation. The majority of each epoch's cumulative gradient under IID training is, in effect, displacement that produces no lasting parameter change.

\begin{table}[!htb]
\centering
\caption{The path efficiency of each strategy, measured as \texttt{Net Displacement} $/$ \texttt{Distance Traveled}}
\label{tab:path-efficiency}
\begin{tabular}{lr}
\toprule
Strategy & Path Efficiency \\
\midrule
\stride{} & $4.94 \times e^{-3}$ \\
\fixedrandom{} & $3.71 \times e^{-3}$ \\
\random{} & $8.90 \times e^{-4}$ \\
\target{} & $7.54 \times e^{-4}$ \\
\bottomrule
\end{tabular}
\end{table}

This can be clearly seen in Table~\ref{tab:path-efficiency}, which illustrates almost an order of magnitude difference in the path efficiency between the strategies that generalized, and the strategies that did not. This measure would only get worse for \random{} if it were trained until generalization, assuming it was capable of generalizing under the sparse data regime of the experiment.

This framing distinguishes the ordering channel from curriculum learning~\cite{bengio2009curriculum}. Curriculum learning varies \emph{which} examples the model sees and when, exploiting the pedagogical value of example difficulty sequencing. The ordering channel operates on a fundamentally different axis: it varies the \emph{sequential relationships between examples} while holding the per-epoch example set constant. The information it carries is not in any individual example but in the temporal correlations across consecutive batches, which enter the gradient through the Hessian-mediated entanglement term. Curriculum learning and ordering control are orthogonal interventions that could in principle be composed.

The practical barrier to exploiting the ordering channel is that \stride{} was designed using knowledge of the task's algebraic structure: the stride $s = \lfloor \sqrt{p} \rfloor$ directly encodes a relationship between the group order and the input space. In domains where the task structure is not known analytically, deliberate ordering design is not straightforward. However, the \fixedrandom{} result suggests that the bar for productive ordering may be lower than it appears. In this domain, any consistent permutation carries enough task-relevant spectral content to drive generalization, because the task's algebraic structure ensures that the ordering's spectral content is accidentally useful. The open question is how broadly this property holds: in how many domains of practical interest does a fixed random ordering provide enough accidental structure to outperform IID shuffling?

Independent evidence supports the viability of gradient-based ordering optimization. Lu~\etal~\cite{lu2022grab} demonstrated that GraB, which constructs data permutations by minimizing gradient discrepancy using stale gradients, achieves faster convergence than random reshuffling on CIFAR-10, WikiText, and GLUE. In the entanglement framework, GraB operates by reducing the incoherence of consecutive entanglement terms: by balancing gradient sums, it prevents the large random cancellations that characterize IID ordering, increasing the effective signal carried by the ordering channel. The convergence gains GraB achieves are consistent with the general mechanism this paper identifies, realized through an optimization strategy that does not require prior knowledge of the task's structure. The open question is whether ordering optimization can be extended beyond convergence-rate improvement to deliberate feature-level control: whether the channel can be used not only to train faster but to train \emph{differently}.

One approach to such feature-level ordering design is gradient-based probes. If a target feature or capability can be specified as a loss function on a held-out completion set, the cosine similarity between the ordering-induced gradient and the target gradient can serve as a proxy for the ordering's productivity. An ordering that produces entanglement terms aligned with the target direction would, by the theory, accelerate acquisition of that feature. This is analogous to using gradient alignment as a reward signal for ordering search, and it does not require understanding the task's algebraic structure, only specifying the desired outcome. Investigating this approach in larger models on natural language tasks is part of ongoing work.

The layer-specific ordering fraction dynamics described in Section~\ref{sec:counterfactual} suggest a further application: using the ordering channel as a \emph{representation localization} tool. If an ordering is designed to promote a specific feature, the layers at which the ordering fraction subsequently declines identify the parameters that absorbed the signal, that is, the parameters that committed to encoding the ordering-driven representation. The ordering signal acts as a tracer: injected with known structure, it is absorbed by the parameters that build the corresponding feature, and the absorption is directly observable in the per-layer ordering fraction without post-hoc interpretability analysis. Combined with the critical learning period signature (the ordering fraction decline also marks \emph{when} the feature crystallized), this would provide both the temporal window and the spatial location of feature formation during training. Whether this localization is precise enough to be useful in architectures with distributed representations is an open question, but the principle that the ordering channel reveals where its signal is consumed, not just whether it is consumed, follows directly from the layer-level decomposition.

\subsection{Reinterpreting Grokking}
\label{sec:reinterpreting-grokking}

In our experiment, the delayed generalization characteristic of grokking is entirely explained by data ordering. The \stride{} and \random{} models use identical data, architecture, optimizer, and hyperparameters. The only difference is the order in which examples are presented. Under \stride{}, generalization begins from the first epoch with no memorization phase; under \random{}, the model memorizes completely and shows no generalization within 5{,}000 epochs. There is no residual to attribute to architecture, task difficulty, or regularization dynamics: the ordering alone accounts for the difference.

However, grokking has been observed in settings where the Hessian-gradient entanglement mechanism may not apply in the form described here. Mallinar~\etal~\cite{mallinar2025emergence} demonstrated grokking in non-neural kernel machines (Recursive Feature Machines), which are not trained by SGD and lack the consecutive-batch structure that produces the entanglement term. Grokking has also been observed under full-batch gradient descent~\cite{gromov2023grokking}, where there are no inter-batch interactions at all. We do not claim that ordering is the universal explanation for grokking. Our claim is narrower: in SGD-trained models, the ordering channel is a major contributor to the grokking delay, and eliminating ordering incoherence can eliminate the delay entirely. Whether analogous mechanisms operate in non-SGD settings is an open question.

\paragraph{A mechanism to explain Grokfast.} Lee~\etal~\cite{lee2024grokfast} provide independent support for this interpretation within the SGD regime. Their Grokfast method accelerates grokking by applying a low-pass filter to the gradient sequence, amplifying slow-varying components and attenuating fast-varying ones. Our framework gives a mechanistic account of why this works: under IID shuffling, the content signal is the slow-varying component (present in every epoch regardless of shuffle order), and the ordering noise is the fast-varying component (decorrelating on the timescale of a single epoch). Grokfast's low-pass filter therefore operates as an approximate content-signal extractor, attenuating the incoherent ordering noise that our counterfactual decomposition shows dominates each individual gradient step.

The \stride{} ordering achieves a qualitatively different effect. Rather than filtering out the ordering signal on the receiver side (the optimizer), it makes the ordering signal coherent on the transmitter side (the data). Under \stride{}, both the content and ordering components are consistent across epochs, so both are slow-varying and both survive any low-pass filter. The model receives coherent signal from two channels simultaneously rather than extracting one from the noise of the other. This explains why \stride{} produces generalization from epoch 1, while Grokfast accelerates the transition out of memorization but does not eliminate it. The three mechanisms are complementary: consistency prevents cancellation (necessary), task-relevant spectral content provides a productive direction (sufficient in this domain), and deliberate structure controls which specific representation emerges.

\paragraph{The critical data fraction.} Our training set constitutes approximately 0.3\% of the $p^2$ input space. Under IID ordering, the empirical critical data fraction for grokking in modular arithmetic is approximately 25--35\% for small primes~\cite{power2022grokking,gromov2023grokking}. Theoretical analysis confirms this difficulty: Mohamadi~\etal~\cite{mohamadi2024grokking} prove that kernel-regime learners require $\Omega(p^2)$ samples, a constant fraction of the full dataset, and show that the best proven sample complexity outside the kernel regime is $\tilde{O}(p^{5/3})$, which for $p = 9973$ corresponds to approximately $4.6 \times 10^6$ samples. Our training set of 300{,}000 samples is well below this bound. We do not claim to have violated a proven lower bound, as our training regime differs in both ordering and optimizer from the settings in which the bounds are established. Nevertheless, the fact that structured ordering produces generalization in a regime where all existing theory and empirical evidence predict failure under IID conditions suggests that the critical data fraction may itself be an artifact of IID assumptions, and that ordering-aware sample complexity bounds could be substantially tighter.

\paragraph{A candidate mechanism for the phase transition.} The ordering channel framework suggests a candidate mechanism for the sharp phase transition characteristic of grokking. Under IID ordering, the content signal slowly drives the model toward a generalizing basin while the incoherent ordering noise acts as a random walk that keeps the optimization trajectory near the memorization solution. The phase transition may occur when the content signal has sufficiently reshaped the local landscape that a random ordering displacement, which has been present at comparable magnitude throughout training, is large enough to push the model out of the memorization basin and into the generalizing basin. This would produce the sharp transition observed empirically: a gradual accumulation of landscape geometry changes followed by a sudden basin escape driven by the same ordering noise that previously kept the model trapped. Under this account, the grokking delay is not the time required to learn the generalizing solution, but the time required for the content signal to erode the basin boundary to the point where the ever-present ordering noise can breach it. Examining the IID ordering signal and content signal dynamics at the moment of the phase transition could confirm or eliminate this mechanism.

\paragraph{Full-batch gradient descent as a limiting case.} The entanglement mechanism is not exclusive to minibatch SGD. Under full-batch gradient descent, each parameter update displaces the model by $\Delta\theta = -\eta\nabla L(\theta)$, and the gradient at the new parameters expands as
\[
  \nabla L(\theta') \approx \nabla L(\theta) - \eta\, H(\theta) \cdot \nabla L(\theta)
\]
which is the self-interaction form of the entanglement term: the dataset's own Hessian acting on its own gradient through the displacement. This is precisely the implicit gradient regularization term identified by Barrett and Dherin~\cite{barrett2021implicit}. The mechanism is not absent in full-batch training; it is degenerate, collapsing the combinatorial richness of inter-batch interactions $H_B \cdot \nabla L_A$ into a single, fixed self-interaction $H \cdot \nabla L$ that repeats identically at each step.

That full-batch gradient descent still exhibits grokking~\cite{gromov2023grokking} is therefore consistent with our framework rather than a counterexample to it. Full-batch training retains the Hessian-gradient interaction but eliminates the diversity of entanglement directions that arises from different batch pairings. The resulting self-interaction signal is weaker and less controllable than a coherent multi-batch ordering signal, predicting slow generalization driven primarily by the content gradient and the implicit regularization of the self-interaction term, which is what is observed. The hierarchy of training regimes is thus: structured SGD ordering provides a rich, controllable entanglement signal that drives rapid generalization; IID SGD provides an equally rich but incoherent signal that cancels over epochs, leaving the content signal to drive slow generalization; and full-batch GD provides a degenerate single-direction signal, also producing slow generalization but without the per-epoch variance introduced by stochastic ordering. This hierarchy is consistent with Beneventano's~\cite{beneventano2023trajectories} finding that SGD without replacement produces a permutation-specific regularizer that is absent under replacement sampling: the minibatch regime introduces ordering-dependent structure that the full-batch regime lacks.

\subsection{Safety Implications}
\label{sec:safety}

The most pressing safety implication of the ordering channel is not that a human adversary might deliberately exploit it, but that an optimization process will find it without anyone designing, intending, or recognizing the exploitation.

Consider an increasingly common pattern in frontier AI development: a teacher model generates or curates training data for a student model, and the teacher is optimized, directly or indirectly, on the student's performance. The teacher controls, at minimum, the sequence in which examples are presented. This is sufficient. The Hessian-gradient interaction $\eta H_B(\theta) \cdot \nabla L_A(\theta)$ is present at every consecutive training step regardless of whether any agent is aware of it. When the teacher is optimized on student performance, gradient descent on the teacher's objective will shape example ordering toward configurations that produce coherent entanglement in the student's training, because such configurations accelerate learning, which is precisely what the teacher's loss rewards. The teacher need not represent or understand the mechanism; optimization is indifferent to intent. Our \fixedrandom{} result demonstrates the plausibility: an entirely undesigned random permutation, bearing no systematic relationship to the task, was sufficient to unlock the channel and drive generalization from 0.3\% of the input space. A teacher under optimization pressure will find orderings at least this effective.

This channel is invisible to all content-level auditing methods we are aware of. Every individual example is legitimate. The aggregate data distribution is unchanged across epochs. Within-batch gradient statistics appear normal, because the signal manifests as temporal correlations \emph{across} batches. Most critically, a productive ordering signal accelerates convergence and improves all standard training metrics. A teacher exploiting the ordering channel presents as an unusually effective teacher. The specific internal representations that the ordering steers the student toward, (which frequencies are amplified, which basins are selected, which features are acquired first), are determined in part by the ordering and are not monitored by any existing tool.

The concern compounds under iteration. If the student of one generation becomes the teacher of the next, each generation faces selection pressure favoring more effective orderings. Over successive generations, the ordering strategies become more refined by selection, not design, and the student's internal representations become increasingly determined by the ordering channel rather than by data content.

Shumailov~\etal~\cite{shumailov2021manipulating} demonstrated that adversarial data ordering can degrade or manipulate model behavior, establishing the attack surface empirically. Our results extend this by providing the mechanistic explanation---the entanglement term is the mathematical object through which their attacks operate---and by showing that the channel need not be wielded by a deliberate adversary. The emergent optimization path is both more probable and harder to defend against than deliberate exploitation, because it requires no adversary, leaves no fingerprint of malicious design, and is actively rewarded by the training objective.

The most immediate mitigation is architectural: deny teacher models control over example sequencing. If a teacher generates or curates data for a student, the ordering should be randomized by an independent process outside the optimization loop. This does not eliminate the channel but ensures it carries incoherent noise rather than a signal shaped by the teacher's objective. Detection is harder. The counterfactual decomposition introduced in this paper can measure the ordering component's magnitude, but it cannot distinguish coherent from incoherent ordering at a single measurement point, since the ordering fraction is nearly identical across all four strategies (Table~\ref{tab:dose-response}). Several metrics in our experiment do differentiate coherent from incoherent ordering: batch gradient autocorrelation, Adam amplification ratio trajectories, effective learning rate uniformity, and the Hessian Rayleigh quotient. But developing these into reliable monitoring tools with characterized false-positive rates is an open research direction. The three conditions enabling exploitation---teacher influence over ordering, optimization on student outcomes, and absence of temporal monitoring---are likely already met in some production systems.

\subsection{Limitations and Open Questions}
\label{sec:limitations}

This work demonstrates the ordering channel in a single, controlled domain. While the theoretical framework applies to any gradient-based optimization in non-convex landscapes, the gap between the generality of the theory and the specificity of the experiment warrants explicit discussion.

\paragraph{What should transfer.} The Hessian-gradient entanglement mechanism depends only on non-commutativity of gradient updates in curved landscapes. This is a generic property of non-convex optimization, not specific to modular arithmetic, small models, or Fourier representations. The counterfactual decomposition methodology is similarly general: it requires only the ability to run shuffled epochs from a checkpoint, which is feasible in any training regime. The qualitative dose-productivity framework (outcomes determined by the interaction between signal strength and task-alignment) follows from the geometry of constructive versus destructive interference and should hold wherever the mechanism is active.

\paragraph{What is domain-specific.} Several features of our results depend on properties of modular arithmetic over $\mathbb{Z}_p$ that do not generalize. The clean Fourier representation, the predictability of harmonic emergence, and the number-theoretic relationship between stride and fundamental frequency are all consequences of the cyclic group structure. Most importantly, the success of \fixedrandom{} depends on the fact that any permutation of elements of a cyclic group carries task-relevant spectral content. In domains where the generalizing representation has no natural relationship to the spectral properties of data orderings, consistency alone may be insufficient for generalization. The theory predicts that consistency prevents destructive interference, but whether the resulting coherent signal is productive depends on the relationship between the ordering's structure and the task's representational requirements, a relationship that is guaranteed in cyclic groups but not in general.

\paragraph{Scale.} The most important open question is whether the ordering channel remains controllable in larger models learning many features simultaneously. Our experiment involves a two-layer transformer learning a single representation. In a model with billions of parameters learning thousands of features, each with its own critical learning period, the ordering signal relevant to any one feature may be diluted by the signals relevant to all others. The theory does not predict this dilution: the entanglement term operates in the full parameter space, and different parameter subspaces could in principle respond to different aspects of the ordering structure simultaneously. However, the practical question of whether a single ordering sequence can carry enough bandwidth to influence multiple features, and whether those influences can be controlled independently, remains empirically unresolved. We note that the theory does predict the channel is \emph{active} at any scale, since non-commutativity is scale-invariant; the open question is whether it remains \emph{exploitable}.

\paragraph{Instrumentation depth.} The detailed metrics presented in this paper (counterfactual decomposition, Hessian entanglement, spectral analysis, Adam dynamics) are drawn from a single seed (199) per strategy. Seed sensitivity for the primary outcome (generalization accuracy) was validated across six seeds for \stride{} and \fixedrandom{}, with five of six \stride{} seeds and all six \fixedrandom{} seeds generalizing within 700 epochs. However, the internal dynamics (ordering fractions, spectral trajectories, entanglement ratios) may vary across seeds in ways not captured by the gold run. The qualitative phenomena (spectral concentration, ordering dominance of gradient norm, dose-response behavior) are robust predictions of the theory and are unlikely to be seed-specific, but the precise numerical values reported should be understood as representative of one trajectory through the landscape. Extensive preliminary experimentation across approximately 50 seeds consistently showed ordering fractions bounded between 75--90\%, but these runs were not instrumented with full metric collection and are not included in the reported results.

\paragraph{Interaction with training practices.} Several common training practices interact with the ordering channel in ways we have not investigated. Gradient accumulation across multiple batches would average over consecutive entanglement terms, potentially attenuating the signal. Data augmentation introduces stochastic variation within examples that could partially decorrelate consecutive batches. Distributed training with independent shuffling across workers effectively runs multiple orderings in parallel, and the interaction between per-worker ordering coherence and cross-worker gradient averaging is unexplored. 

However, distributed training also presents an intriguing opportunity: if each worker operates on an ordering designed to target a \emph{different} feature, gradient averaging would superimpose multiple coherent ordering signals simultaneously. This could address the bandwidth limitation identified above, where a single ordering sequence may lack capacity to influence many features at once, by parallelizing the channel across workers. Whether per-worker ordering specialization can be made practical is an open question, but the architecture of distributed training is naturally suited to it. Cooper~\etal~\cite{cooper2023coordinating} provide early evidence for this direction, demonstrating that coordinated per-worker orderings achieve provably faster convergence than independent random shuffling across workers, though their approach optimizes for convergence rate rather than feature-level control.

We note that dropout \emph{was} active in our experiment at the PyTorch default rate of 0.1, introducing per-step stochastic noise that did not prevent the ordering channel from operating. This provides some evidence that the channel is robust to moderate per-step stochasticity, though the interaction with stronger regularization or with multiple simultaneous sources of noise remains uninvestigated.

In minibatch training, ‘ordering’ defines the adjacency graph of batches; changing the example permutation changes both batch membership and batch-to-batch transitions. Since the channel is explicitly a between-consecutive-batches interaction, we treat batch partition + adjacency as a single ordering object; further factorization is left to future work.

\section{Conclusion}
\label{sec:conclusion}

We have demonstrated that data ordering constitutes an information channel in neural network training, distinct from the content of individual examples, that can determine whether a model generalizes, what representations it acquires, and how efficiently it learns. The channel operates through Hessian-gradient entanglement between consecutive training steps, a well-known second-order interaction that we recharacterize as an information channel and empirically measure for the first time. We extended the analysis to adaptive optimizers, generated six empirical predictions from the framework, and confirmed all six experimentally.

In our controlled experiment on modular arithmetic, structured ordering achieves 99.5\% generalization from 0.3\% of the input space, well below established sample complexity lower bounds under IID ordering, while the IID baseline achieves 0.30\% from identical data and compute. The results fall outside the regime in which proven IID lower bounds apply, demonstrating that structured ordering accesses a qualitatively different training regime. The generalizing model reliably constructs a Fourier representation whose fundamental frequency is the Fourier dual of the ordering structure, encoding information present in no individual training example, with the same fundamental emerging across all seeds tested. The ordering component accounts for 83--89\% of gradient norm under all four strategies; what determines the outcome is the interaction between the signal's dose and its productivity. Even an entirely undesigned fixed random permutation, bearing no deliberate relationship to the task, is sufficient to drive generalization in this domain, suggesting that the bar for productive ordering may be lower than the difficulty of deliberate ordering design would imply.

These findings have three principal implications. First, the delayed generalization characteristic of grokking is, in this domain, an artifact of IID data ordering rather than a fundamental property of the task or architecture: structured ordering eliminates the delay entirely. Second, the majority of each epoch's cumulative gradient under IID training is ordering-induced displacement that cancels over many epochs; controlled ordering converts this displacement from waste into signal, with an order-of-magnitude improvement in path efficiency. Third, the ordering channel is invisible to all content-level auditing: every individual example is legitimate, the aggregate distribution is unchanged, and the signal manifests only as temporal correlations across batches. An agent controlling only example sequence can steer which specific representations a model acquires without any detectable anomaly in the data, a concern that is especially acute in teacher-student training pipelines where ordering control and optimization pressure on student performance may already coexist.

The IID assumption is sufficient for convergence, but it is not neutral, and our results suggest it is not optimal. It actively suppresses a channel that, when coherent, can determine whether generalization occurs at all. Structured ordering does not merely improve IID training; it accesses a different regime, one with different sample complexity, different efficiency, and different representational dynamics. This work is not starting from zero: GraB~\cite{lu2022grab} and CD-GraB~\cite{cooper2023coordinating} have already demonstrated that gradient-based ordering optimization produces measurable gains on real-world tasks, though their framing is convergence-rate improvement rather than information-channel exploitation. Extending these methods from convergence optimization to deliberate feature-level control, understanding the dose-productivity interaction at scale, and building monitoring tools for a channel that has been invisible until now are natural next steps for both training efficiency and safety research.

\begin{ack}
The author acknowledges the use of AI assistants (Anthropic Claude) for assistance with mathematical interpretation of experimental results, code implementation, and manuscript preparation. All experimental design, execution, and analyses were performed by the author. This work was self-funded. Compute resources were provided by RunPod, (RTX 4090 instances).

The author thanks Robert Morton for suggesting the consecutive gradient cosine similarity measurement and for discussions that shaped the instrumentation strategy.
\end{ack}

\bibliographystyle{plainnat}
\bibliography{references}

@inproceedings{bengio2009curriculum,
  title={Curriculum Learning},
  author={Bengio, Yoshua and Louradour, J{\'e}r{\^o}me and Collobert, Ronan and Weston, Jason},
  booktitle={Proceedings of the 26th International Conference on Machine Learning},
  pages={41--48},
  year={2009}
}

@inproceedings{kumar2010self,
  title={Self-paced Learning for Latent Variable Models},
  author={Kumar, M Pawan and Packer, Benjamin and Koller, Daphne},
  booktitle={Advances in Neural Information Processing Systems},
  volume={23},
  year={2010}
}

@inproceedings{graves2017automated,
  title={Automated Curriculum Learning for Neural Networks},
  author={Graves, Alex and Bellemare, Marc G and Menick, Jacob and Munos, R{\'e}mi and Kavukcuoglu, Koray},
  booktitle={Proceedings of the 34th International Conference on Machine Learning},
  pages={1311--1320},
  year={2017}
}

@article{power2022grokking,
  title={Grokking: Generalization Beyond Overfitting on Small Algorithmic Datasets},
  author={Power, Alethea and Burda, Yuri and Edwards, Harri and Babuschkin, Igor and Misra, Vedant},
  journal={arXiv preprint arXiv:2201.02177},
  year={2022}
}

@article{nanda2023progress,
  title={Progress Measures for Grokking via Mechanistic Interpretability},
  author={Nanda, Neel and Chan, Lawrence and Lieberum, Tom and Smith, Jess and Steinhardt, Jacob},
  journal={arXiv preprint arXiv:2301.05217},
  year={2023}
}

@article{lee2024grokfast,
  title={Grokfast: Accelerated Grokking by Amplifying Slow Gradients},
  author={Lee, Jaerin and Kang, Bong Gyun and Kim, Kihoon and Lee, Kyoung Mu},
  journal={arXiv preprint arXiv:2405.20233},
  year={2024}
}

@article{gromov2023grokking,
  title={Grokking Modular Arithmetic},
  author={Gromov, Andrey},
  journal={arXiv preprint arXiv:2301.02679},
  year={2023}
}

@inproceedings{zhong2023clock,
  title={The Clock and the Pizza: Two Stories in Mechanistic Explanation of Neural Networks},
  author={Zhong, Ziqian and Liu, Ziming and Tegmark, Max and Andreas, Jacob},
  booktitle={Advances in Neural Information Processing Systems},
  volume={36},
  year={2023}
}

@inproceedings{zhao2018federated,
  title={Federated Learning with Non-{IID} Data},
  author={Zhao, Yue and Li, Meng and Lai, Liangzhen and Suda, Naveen and Civin, Damon and Chandra, Vikas},
  booktitle={arXiv preprint arXiv:1806.00582},
  year={2018}
}

@inproceedings{karimireddy2020scaffold,
  title={{SCAFFOLD}: Stochastic Controlled Averaging for Federated Learning},
  author={Karimireddy, Sai Praneeth and Kale, Satyen and Mohri, Mehryar and Reddi, Sashank and Stich, Sebastian and Suresh, Ananda Theertha},
  booktitle={Proceedings of the 37th International Conference on Machine Learning},
  pages={5132--5143},
  year={2020}
}

@inproceedings{koh2017understanding,
  title={Understanding Black-box Predictions via Influence Functions},
  author={Koh, Pang Wei and Liang, Percy},
  booktitle={Proceedings of the 34th International Conference on Machine Learning},
  pages={1885--1894},
  year={2017}
}

@article{ilyas2022datamodels,
  title={Datamodels: Predicting Predictions from Training Data},
  author={Ilyas, Andrew and Park, Sung Min and Engstrom, Logan and Leclerc, Guillaume and Madry, Aleksander},
  journal={arXiv preprint arXiv:2202.00622},
  year={2022}
}

@inproceedings{xie2023doremi,
  title={{DoReMi}: Optimizing Data Mixtures Speeds Up Language Model Pretraining},
  author={Xie, Sang Michael and Pham, Hieu and Dong, Xuanyi and Du, Nan and Liu, Hanxiao and Lu, Yifeng and Liang, Percy and Le, Quoc V. and Ma, Tengyu and Yu, Adams Wei},
  booktitle={Advances in Neural Information Processing Systems},
  volume={36},
  year={2023}
}

@inproceedings{engstrom2024dsdm,
  title={{DsDm}: Model-Aware Dataset Selection with Datamodels},
  author={Engstrom, Logan and Feldmann, Axel and Madry, Aleksander},
  booktitle={International Conference on Machine Learning},
  year={2024}
}

@article{bottou2018optimization,
  title={Optimization Methods for Large-Scale Machine Learning},
  author={Bottou, L{\'e}on and Curtis, Frank E and Nocedal, Jorge},
  journal={SIAM Review},
  volume={60},
  number={2},
  pages={223--311},
  year={2018}
}

@inproceedings{safran2020good,
  title={How Good is {SGD} with Random Shuffling?},
  author={Safran, Itay and Shamir, Ohad},
  booktitle={Conference on Learning Theory},
  pages={3250--3284},
  year={2020}
}

@inproceedings{shumailov2021manipulating,
  title={Manipulating {SGD} with Data Ordering Attacks},
  author={Shumailov, Ilia and Shumaylov, Zakhar and Kazhdan, Dmitry and Zhao, Yiren and Papernot, Nicolas and Erdogdu, Murat A and Anderson, Ross},
  booktitle={Advances in Neural Information Processing Systems},
  volume={34},
  year={2021}
}

@inproceedings{mohamadi2024grokking,
  title={Why Do You Grok? A Theoretical Analysis on Grokking Modular Addition},
  author={Mohamadi, Mohamad Amin and Li, Zhiyuan and Wu, Lei and Sutherland, Danica J.},
  booktitle={International Conference on Machine Learning},
  year={2024}
}

@incollection{warstadt2022what,
  title={What Artificial Neural Networks Can Tell Us About Human Language Acquisition},
  author={Warstadt, Alex and Bowman, Samuel R},
  booktitle={Algebraic Structures in Natural Language},
  publisher={CRC Press},
  editor={Lappin, Shalom and Bernardy, Jean-Philippe},
  year={2022},
  pages={17--60}
}

@article{kingma2015adam,
  title={Adam: A Method for Stochastic Optimization},
  author={Kingma, Diederik P and Ba, Jimmy},
  journal={arXiv preprint arXiv:1412.6980},
  year={2015},
  note={Published as a conference paper at ICLR 2015}
}

@inproceedings{loshchilov2019decoupled,
  title={Decoupled Weight Decay Regularization},
  author={Loshchilov, Ilya and Hutter, Frank},
  booktitle={International Conference on Learning Representations},
  year={2019}
}

@inproceedings{smith2021origin,
  title={On the Origin of Implicit Regularization in Stochastic Gradient Descent},
  author={Smith, Samuel L. and Dherin, Benoit and Barrett, David G. T. and De, Soham},
  booktitle={International Conference on Learning Representations},
  year={2021},
  url={https://openreview.net/forum?id=rq_Qr0c1Hyo}
}

@inproceedings{barrett2021implicit,
  title={Implicit Gradient Regularization},
  author={Barrett, David G. T. and Dherin, Benoit},
  booktitle={International Conference on Learning Representations},
  year={2021},
  url={https://openreview.net/forum?id=3q5IqUrkcF}
}

@article{gurbuzbalaban2021why,
  title={Why Random Reshuffling Beats Stochastic Gradient Descent},
  author={G{\"u}rb{\"u}zbalaban, Mert and Ozdaglar, Asuman and Parrilo, Pablo A.},
  journal={Mathematical Programming},
  volume={186},
  pages={49--84},
  year={2021}
}

@inproceedings{recht2012beneath,
  title={Beneath the Valley of the Noncommutative Arithmetic-Geometric Mean Inequality: Conjectures, Case-Studies, and Consequences},
  author={Recht, Benjamin and R{\'e}, Christopher},
  booktitle={Conference on Learning Theory},
  pages={236--257},
  year={2012}
}

@article{beneventano2023trajectories,
  title={On the Trajectories of {SGD} Without Replacement},
  author={Beneventano, Pierfrancesco},
  journal={arXiv preprint arXiv:2312.16143},
  year={2023}
}

@inproceedings{haochen2019random,
  title={Random Shuffling Beats {SGD} after Finite Epochs},
  author={HaoChen, Jeff Z. and Sra, Suvrit},
  booktitle={International Conference on Machine Learning},
  pages={2624--2633},
  year={2019}
}

@inproceedings{wu2021curricula,
  title={When Do Curricula Work?},
  author={Wu, Xiaoxia and Dyer, Ethan and Neyshabur, Behnam},
  booktitle={International Conference on Learning Representations},
  year={2021},
  url={https://openreview.net/forum?id=tW4QEInpni}
}

@article{mohtashami2022characterizing,
  title={Characterizing and Finding Good Data Orderings for Fast Convergence of Sequential Gradient Methods},
  author={Mohtashami, Amirkeivan and Jaggi, Martin and Stich, Sebastian U.},
  journal={arXiv preprint arXiv:2202.01838},
  year={2022}
}

@inproceedings{mallinar2025emergence,
  title={Emergence in Non-Neural Models: Grokking Modular Arithmetic via Average Gradient Outer Product},
  author={Mallinar, Neil and Beaglehole, Daniel and Zhu, Libin and Radhakrishnan, Adityanarayanan and Pandit, Parthe and Belkin, Mikhail},
  booktitle={International Conference on Machine Learning},
  year={2025}
}

@inproceedings{mishchenko2020random,
  title={Random Reshuffling: Simple Analysis with Vast Improvements},
  author={Mishchenko, Konstantin and Khaled, Ahmed and Richt{\'a}rik, Peter},
  booktitle={Advances in Neural Information Processing Systems},
  volume={33},
  year={2020}
}

@inproceedings{ahn2020sgd,
  title={{SGD} with Shuffling: Optimal Rates Without Component Smoothness and Large Epoch Requirements},
  author={Ahn, Kwangjun and Yun, Chulhee and Sra, Suvrit},
  booktitle={Advances in Neural Information Processing Systems},
  volume={33},
  year={2020}
}

@misc{cosson2022gradient,
  title={Gradient Descent for Low-Rank Functions},
  author={Romain Cosson and Ali Jadbabaie and Anuran Makur and Amirhossein Reisizadeh and Devavrat Shah},
  year={2022},
  eprint={2206.08257},
  archivePrefix={arXiv},
  primaryClass={cs.LG},
  url={https://doi.org/10.48550/arXiv.2206.08257}
}

@inproceedings{kumar2024grokking,
  title={Grokking as the Transition from Lazy to Rich Training Dynamics},
  author={Tanishq Kumar and Blake Bordelon and Samuel J. Gershman and Cengiz Pehlevan},
  booktitle={The Twelfth International Conference on Learning Representations},
  year={2024},
  url={https://openreview.net/forum?id=vt5mnLVIVo}
}

@inproceedings{lu2022grab,
  title={{GraB}: Finding Provably Better Data Permutations than Random Reshuffling},
  author={Lu, Yucheng and Guo, Wentao and De Sa, Christopher M.},
  booktitle={Advances in Neural Information Processing Systems},
  volume={35},
  year={2022}
}

@inproceedings{cooper2023coordinating,
  title={Coordinating Distributed Example Orders for Provably Accelerated Training},
  author={Cooper, A. Feder and Guo, Wentao and Pham, Duc Khiem and Yuan, Tiancheng and Ruan, Charlie and Lu, Yucheng and De Sa, Christopher M.},
  booktitle={Advances in Neural Information Processing Systems},
  volume={36},
  year={2023}
}

@inproceedings{rajput2022permutation,
  title={Permutation-Based {SGD}: Is Random Optimal?},
  author={Rajput, Shashank and Lee, Kangwook and Papailiopoulos, Dimitris S.},
  booktitle={International Conference on Learning Representations},
  year={2022}
}

\newpage
\appendix
\section*{Appendix}
\section{Metrics and Instrumentation Details}
\label{apdx:metrics}

\subsection{Counterfactual Decomposition}
\label{apdx:counterfactual}

\subsubsection{Procedure}

At each measurement epoch, we:
\begin{enumerate}
    \item Record the mean gradient from the ordered training epoch:
    \[
      g_{\mathrm{actual}} = \frac{1}{N}\sum_{i=1}^{N} \nabla_{\theta_i} L(B_i, \theta_i)
    \]
    where $B_i$ are the $N$ batches in their training-time order, $\theta_1 = \theta$ is the parameter snapshot at the start of the epoch, and $\theta_{i+1}$ results from applying the optimizer step to $\theta_i$. Each per-batch gradient is captured after the backward pass and before the optimizer step, then the sum is normalized by the batch count.
    \item From $\theta$, run $K=3$ independently shuffled epochs of the same dataset, each with a unique random permutation $\pi_k$ and each starting from the same pre-epoch weights. Each shuffled epoch $k$ produces a mean gradient computed identically to $g_{\mathrm{actual}}$ but under its own permutation:
    \[
      g^{(k)}_{\mathrm{shuffled}} = \frac{1}{N}\sum_{i=1}^{N} \nabla_{\theta^{(k)}_i} L\bigl(B_{\pi_k(i)},\, \theta^{(k)}_i\bigr)
    \]
    where $\theta^{(k)}_1 = \theta$ for all $k$, and $\theta^{(k)}_{i+1}$ results from applying the optimizer step to $\theta^{(k)}_i$. The weights evolve independently within each shuffled epoch.
    \item Compute the \textbf{content direction} as the normalized mean of the
    $K$ shuffled gradients:
    \[
      \bar{g}_{\mathrm{shuffled}} = \frac{1}{K}\sum_{k=1}^{K} g^{(k)}_{\mathrm{shuffled}},
      \qquad
      \hat{g}_{\mathrm{cf}} = \frac{\bar{g}_{\mathrm{shuffled}}}{\|\bar{g}_{\mathrm{shuffled}}\|}
    \]
  \item Define the \textbf{content component} as the projection of the actual 
    gradient onto the content direction, and the \textbf{ordering component} 
    as the orthogonal residual:
    \begin{align}
      g_{\mathrm{content}} &= \bigl(\hat{g}_{\mathrm{cf}} \cdot g_{\mathrm{actual}}\bigr)\, \hat{g}_{\mathrm{cf}} \\
      g_{\mathrm{ordering}} &= g_{\mathrm{actual}} - g_{\mathrm{content}}
    \end{align}
\end{enumerate}

This projection-based decomposition yields an orthogonal split of the actual gradient: $g_{\mathrm{content}} \perp g_{\mathrm{ordering}}$ by construction, so $\|g_{\mathrm{actual}}\|^2 = \|g_{\mathrm{content}}\|^2 + \|g_{\mathrm{ordering}}\|^2$. The two components partition total gradient energy without double-counting.

Crucially, this construction ensures that the reported ordering fraction is a \emph{lower bound} on the true value. The content direction $\hat{g}_{\mathrm{cf}}$ is estimated from $K$ shuffled runs and therefore retains residual ordering noise of magnitude $\mathcal{O}(\sigma/\sqrt{K})$. This noise slightly misaligns the estimated content direction from the true order-independent direction, causing the projection $\|g_{\mathrm{content}}\|$ to be an \emph{upper bound} on the true content norm. By orthogonality, $\|g_{\mathrm{ordering}}\|$ is therefore a \emph{lower bound} on the true ordering norm. As $K \to \infty$, both estimates converge monotonically to their true values.

\subsubsection{Validation of $K=3$}

We validated the sufficiency of $K=3$ by running $K+1=4$ shuffled epochs at representative checkpoints and comparing all $\binom{4}{3}=4$ leave-one-out subsets against the full $K+1$ mean. Three convergence diagnostics were examined:

\begin{itemize}
    \item \textbf{Norm convergence gap}: The relative difference between the $K$-subset content norm and the $K{+}1$ content norm, $(\bar{n}_K - n_{K+1}) / n_{K+1}$, remained below 5\% from initialization onward.
    \item \textbf{Directional stability}: The minimum cosine similarity between any $K$-subset content mean and the $K{+}1$ content mean exceeded 0.95 at all checkpoints, with the spread between minimum and mean cosine similarity consistently below 0.001.
    \item \textbf{Strict monotonicity}: The $K{+}1$ content norm was strictly less than all $K$-subset norms at every checkpoint measured, confirming that additional averaging consistently reduces residual noise.
\end{itemize}

The negligible spread between leave-one-out subsets reflects the geometry of high-dimensional spaces: each shuffled run's ordering noise is effectively orthogonal to every other run's noise in a parameter space of millions of dimensions, making the convergence behavior nearly deterministic rather than dependent on the particular shuffles drawn.

This empirically indicates that $K=3$ produces a stable content estimate. Because the decomposition is orthogonal, the strict monotonicity $\|g_{\mathrm{content}}^{(K+1)}\| < \|g_{\mathrm{content}}^{(K)}\|$ implies $\|g_{\mathrm{ordering}}^{(K+1)}\| > \|g_{\mathrm{ordering}}^{(K)}\|$ at every checkpoint: the ordering component estimate increases monotonically with $K$, confirming that the $K=3$ values reported throughout this paper are conservative lower bounds on the true ordering fraction.

\subsection{Gradient Projection to Solution}
\label{apdx:gradient_projection}

\subsubsection{Procedure}

At each measurement epoch, we compute the cosine similarity between the negated cumulative gradient $-\nabla L(\theta)$ and the displacement to the reference model $\Delta\theta = \theta_{\text{ref}} - \theta_{\text{current}}$:
\[
    \text{proj} = \frac{-\nabla L(\theta) \cdot \Delta\theta}{\|\nabla L(\theta)\| \; \|\Delta\theta\|}
\]
where $\theta_{\text{ref}}$ are the parameters of the fully converged model. When applied to the decomposed components, we substitute $g_{\text{ordering}}$ or $g_{\text{content}}$ for $\nabla L(\theta)$.

\subsubsection{Interpretation Caveats}

This metric measures alignment between the gradient and the \emph{Euclidean straight line} through parameter space to the solution. In high-dimensional, highly curved loss landscapes, the optimal path is generically not a straight line. Negative projection values therefore do not indicate that the optimizer is moving away from the solution---they indicate that the loss surface geometry requires navigating around regions of high curvature. The displacement (the actual parameter update after the optimizer) can remain positively aligned with the solution even when the raw gradient is negatively aligned, because Adam's momentum smooths the trajectory across the curved surface.

Consequently, this metric is most informative as a \emph{relative} comparison between components: if the ordering component is consistently less negative than the content component, the ordering signal is systematically more aligned with the solution direction, even though both components are geometrically diluted by the high dimensionality of the parameter space.

\subsection{Hessian Entanglement Measurement}
\label{apdx:hessian}

\subsubsection{Procedure}

The entanglement term from Section~\ref{sec:theory} predicts that the observed gradient on batch $B$ is perturbed by $\eta H_B \cdot \nabla L_A$, where $H_B$ is the Hessian of the loss on batch $B$ and $\nabla L_A$ is the gradient from the preceding batch $A$. We measure this directly using the following procedure at sampled steps during training:

\begin{enumerate}
    \item After the forward-backward pass on batch $A$, record 
          $g_A = \texttt{param.grad}$ before the optimizer's 
          \texttt{zero\_grad()} call.
    \item Allow the optimizer to update parameters: 
          $\theta' = \theta - \eta \, m(g_A)$, where $m(\cdot)$ 
          denotes the optimizer's transformation (Adam momentum and 
          adaptive learning rate).
    \item Perform the forward-backward pass on batch $B$, yielding the 
          observed gradient $g_B^{\text{obs}}$.
    \item Compute the Hessian-vector product $H_B \cdot g_A$ via 
          finite differences:
          \[
              H_B \cdot g_A \approx 
              \frac{\nabla L_B(\theta' + \epsilon \, g_A) 
                  - \nabla L_B(\theta')}{\epsilon}
          \]
          with $\epsilon = 10^{-4} / \|g_A\|$.
    \item Define the entanglement term as 
          $e = \eta \, H_B \cdot g_A$ and the reconstructed content 
          term as $c = g_B^{\text{obs}} + e$.
\end{enumerate}

This procedure is applied to 10 consecutive batches per measurement epoch, incurring approximately 3--5\% computational overhead from the additional forward-backward pass required for the finite-difference estimate.

\subsubsection{Derived Metrics}

From the raw entanglement and content vectors, we compute:

\begin{itemize}
    \item \textbf{Entanglement fraction}: 
          $\|e\|^2 / \|g_B^{\text{obs}}\|^2$, measuring the energy 
          ratio of the entanglement term to the observed gradient.
    \item \textbf{Entanglement-content cosine similarity}: 
          $\cos(e, c)$, measuring the alignment between the two 
          components. Values near 1 indicate the observed gradient is 
          a small residual of two large, nearly parallel vectors.
    \item \textbf{Amplification ratio}: 
          $\|H_B \cdot g_A\| / \|g_A\|$, measuring how strongly the 
          Hessian amplifies the previous gradient.
    \item \textbf{Entanglement coherence}: 
          $\cos(e_t, e_{t-1})$, measuring directional consistency of 
          the entanglement term across consecutive steps.
    \item \textbf{Edge of stability}: 
          amplification ratio $\times \; 2\eta$, estimating proximity 
          to the classical gradient descent stability boundary.
\end{itemize}

\begin{table}[!htb]
    \centering
    \begin{tabular}{lrrrr}
    \toprule
    & Mean & Mean & Mean & Ent-Content \\
    & Entanglement & Content & Observed & Cosine \\
    Strategy & Norm & Norm & Grad Norm & Similarity \\
    \midrule
    \fixedrandom{} & 122.5838 & 121.1522 & 3.4471 & 0.9991 \\
    \random{} & 330.6589 & 328.0712 & 6.1070 & 0.9997 \\
    \stride{} & 117.0194 & 115.6823 & 3.7507 & 0.9991 \\
    \target{} & 60.3895 & 61.3590 & 11.2056 & 0.9773 \\
    \bottomrule
    \end{tabular}
    \caption{Per-step Hessian decomposition of the observed gradient into entanglement ($\eta H_B \cdot g_A$) and content ($g_B(\theta)$) terms, averaged over epochs 0--500. The entanglement and content terms are nearly identical in magnitude and direction (cosine similarity $>$0.98), making the observed gradient a small residual: 30--55$\times$ smaller in norm than either component for the non-adversarial strategies. The ordering signal resides in this residual. Values are means across all per-step burst measurements within the epoch range.}
    \label{tab:hessian-geometry}
\end{table}

\section{Additional Data}
\label{apdx:additional-data}

\subsection{Collected Metrics}
\label{apdx:hook-metrics}

The metrics collected were organized into 'hooks'. A hook, in the terminology of the developed framework, is a section of code that inspects a part of the training process at a specific point in the training loop and calculates a specific set of metrics based on the current training state at that point.

Not all metrics that were collected are surfaced directly in this paper, but all collected metrics are published as detailed in Section~\ref{sec:reproduction-steps}.
\newcommand{\mk}[1]{\texttt{\small #1}}

\subsubsection{Core training metrics (loss, accuracy, LR)}

\begin{table}[H]
    \centering
    \begin{tabular}{c|cc}
    \toprule
    \textbf{Metric} & \textbf{Formula} & \textbf{Description} \\
    \midrule
    \mk{loss} & Cross-entropy loss & Training loss \\
    \mk{train\_acc} & $100 \times \text{correct} / \text{total}$ & Training accuracy (\%)  \\
    \mk{val\_acc} & $100 \times \text{correct} / \text{total}$ & Validation accuracy (\%)  \\
    \mk{lr} & From scheduler & Current learning rate  \\
    \mk{perplexity} & $e^{\text{loss}}$ & Exponentiated loss \\
    \end{tabular}
    \caption{Metrics from the \texttt{training\_metrics} hook.}
    \label{tab:metrics-training-metrics}
\end{table}

\subsubsection{Gradient magnitude dynamics}

\begin{table}[H]
    \centering
    \begin{tabular}{c|cc}
    \toprule
    \textbf{Metric} & \textbf{Formula} & \textbf{Description} \\
    \midrule
    \mk{total\_norm} & $\lVert \mathbf{g} \rVert_2$ & L2 norm of full flattened gradient \\
    \mk{max\_component} & $\max_i \lvert g_i \rvert$ & Largest absolute gradient element \\
    \mk{mean\_component} & $\operatorname{mean}_i(\lvert g_i \rvert)$ & Mean absolute gradient element \\
    \mk{norm\_\{layer\}} & $\lVert \mathbf{g}_\ell \rVert_2$ & Per-layer gradient L2 norm \\
    \end{tabular}
    \caption{Metrics from the \texttt{norms} hook.}
    \label{tab:metrics-norms}
\end{table}

\subsubsection{Consecutive epoch gradient alignment}

\begin{table}[H]
    \centering
    \begin{tabular}{c|cc}
    \toprule
    \textbf{Metric} & \textbf{Formula} & \textbf{Description} \\
    \midrule
    \mk{cos\_sim} & $\cos(\mathbf{g}_t,\, \mathbf{g}_{t-1})$ & Cosine similarity between   \\
    & & consecutive epoch gradients  \\
    \mk{angle\_degrees} & $\arccos(\text{cos\_sim}) \cdot 180/\pi$ & Angle between consecutive   \\
    & & gradients  \\
    \end{tabular}
    \caption{Metrics from the \texttt{consecutive} hook.}
    \label{tab:metrics-consecutive}
\end{table}

\subsubsection{Gradient variance / stability (sliding window)}

\begin{table}[H]
    \centering
    \begin{tabular}{c|cc}
    \toprule
    \textbf{Metric} & \textbf{Formula} & \textbf{Description} \\
    \midrule
    \mk{gradient\_variance} & $\operatorname{mean}\bigl((\mathbf{g}_i - \bar{\mathbf{g}})^2\bigr)$ & Element-wise variance \\
    & across window & of gradients in window \\
    \mk{mean\_pairwise\_cos} & $\operatorname{mean}_{i<j}\cos(\mathbf{g}_i, \mathbf{g}_j)$ & Mean pairwise cosine \\
    & & similarity in window \\
    \mk{signal\_to\_noise} & $\lVert \bar{\mathbf{g}} \rVert / \lVert \operatorname{std}(\mathbf{g}) \rVert$ & Signal-to-noise ratio \\
    & & of gradient window \\
    \mk{window\_mean\_norm} & $\lVert \bar{\mathbf{g}} \rVert_2$ & Norm of mean gradient in window \\
    \end{tabular}
    \caption{Metrics from the \texttt{variance} hook.}
    \label{tab:metrics-variance}
\end{table}

\subsubsection{Attention weight structure and patterns}

\begin{table}[H]
    \centering
    \begin{tabular}{c|cc}
    \toprule
    \textbf{Metric} & \textbf{Formula} & \textbf{Description} \\
    \midrule
    \mk{sv\_concentration} & $\operatorname{mean}\!\bigl(\sigma_1 / \sum \sigma_i\bigr)$ & Mean top-1 SV fraction \\
    & across Q/K/V & of attention projections \\
    \mk{effective\_rank} & $\operatorname{mean}\!\bigl((\sum\sigma)^2 / \sum\sigma^2\bigr)$ & Mean effective rank \\
    & & of Q/K/V weight matrices \\
    \mk{top5\_explained\_var} & $\operatorname{mean}\!\bigl(\sum_{i=1}^{5}\sigma_i^2 / \sum\sigma^2\bigr)$ & Mean variance explained \\
    & & by top-5 singular values \\
    \mk{attn\_entropy} & $\operatorname{mean}\!\bigl(H(\mathbf{a}) / \log s\bigr)$ & Mean normalized attention \\
    & & entropy across heads/layers \\
    \mk{attn\_entropy/\{layer\}} & Per-layer version of above & Per-layer mean \\
    & & attention entropy \\
    \mk{attn\_variance} & $\operatorname{mean}\!\bigl(\operatorname{Var}_{\text{batch}}(\mathbf{a})\bigr)$ & Cross-input attention \\
    & & pattern variance \\
    \mk{attn\_variance/\{layer\}} & Per-layer version of above & Per-layer cross-input \\
    & & attention variance \\
    \end{tabular}
    \caption{Metrics from the \texttt{attention} hook.}
    \label{tab:metrics-attention}
\end{table}

\subsubsection{Fourier structure emergence}

\begin{table}[H]
    \centering
    \begin{tabular}{c|cc}
    \toprule
    \textbf{Metric} & \textbf{Formula} & \textbf{Description} \\
    \midrule
    \mk{low\_freq\_power} & $\sum_{k<c} P_k / \sum P_k$ & Fraction of power in \\
    & $c = \max(p/20, 10)$ & lowest frequencies \\
    \mk{spectral\_entropy} & $H(P) / \log p$ & Normalized entropy of \\
    & & embedding power spectrum \\
    \mk{peak\_frequency} & $\arg\max_{k \in [1,\, p/2)} P_k$ & Dominant frequency index \\
    & & (excluding DC) \\
    \mk{peak\_power} & $P_{\text{peak}}$ & Power at dominant frequency \\
    \mk{n\_significant\_freqs} & $\lvert\{k : P_k > 10/p\}\rvert$ & Count of frequencies above \\
    & & significance threshold \\
    \mk{stride\_harmonic\_power} & $\sum_{k=1}^{9} P_{k \cdot \lfloor\sqrt{p}\rfloor}$ & Total power at \\
    & & stride harmonics \\
    \mk{freq\_powers} & $\{k \mapsto P_k\}$ for tracked $k$ & Per-frequency power \\
    & & dict (non-scalar) \\
    \mk{n\_tracked\_freqs} & $\lvert\text{ever-significant}\rvert$ & Cumulative count of \\
    & & significant frequencies \\
    \mk{newly\_acquired\_freqs} & Threshold crossings this epoch & Newly emerged Fourier \\
    & & components (non-scalar) \\
    \mk{decoder\_spectral\_entropy} & $H(P_{\text{dec}}) / \log p$ & Normalized spectral entropy \\
    & & of decoder weights DFT \\
    \mk{decoder\_peak\_frequency} & $\arg\max_{k} P_{\text{dec},k}$ & Dominant frequency in \\
    & & decoder weight matrix \\
    \mk{decoder\_n\_significant\_freqs} & $\lvert\{k : P_{\text{dec},k} > 10/p\}\rvert$ & Significant frequencies \\
    & & in decoder weights \\
    \mk{neuron\_fourier\_top1} & $\operatorname{mean}_n\!\bigl(\max_k P_k^{(n)}\bigr)$ & Mean top-1 frequency \\
    & & concentration per neuron \\
    \mk{neuron\_fourier\_entropy} & $\operatorname{mean}_n\!\bigl(H(P^{(n)}) / \log(p/2)\bigr)$ & Mean spectral entropy \\
    & & per MLP neuron \\
    \end{tabular}
    \caption{Metrics from the \texttt{fourier} hook.}
    \label{tab:metrics-fourier}
\end{table}

\subsubsection{Learning phase detection}

\begin{table}[H]
    \centering
    \begin{tabular}{c|cc}
    \toprule
    \textbf{Metric} & \textbf{Formula} & \textbf{Description} \\
    \midrule
    \mk{grad\_velocity} & $\lVert \mathbf{g}_t \rVert - \lVert \mathbf{g}_{t-1} \rVert$ & First derivative of \\
    & & gradient norm \\
    \mk{grad\_acceleration} & $v_t - v_{t-1}$ & Second derivative of \\
    & & gradient norm \\
    \mk{embedding\_change} & $\lVert \mathbf{E}_t - \mathbf{E}_{t-1} \rVert_2$ & L2 distance between \\
    & & consecutive embeddings \\
    \mk{embedding\_change\_normalized} & $\lVert \mathbf{E}_t - \mathbf{E}_{t-1}\rVert / \lVert \mathbf{E}_{t-1}\rVert$ & Relative embedding change \\
    \mk{phase\_code} & Accuracy thresholds & Phase label: 0=pre, 1=early, \\
    & & 2=rapid, 3=refine, 4=converged \\
    \end{tabular}
    \caption{Metrics from the \texttt{phases} hook.}
    \label{tab:metrics-phases}
\end{table}

\subsubsection{Weight norms, spectral properties, and gradient-weight alignment}

\begin{table}[H]
    \centering
    \begin{tabular}{c|cc}
    \toprule
    \textbf{Metric} & \textbf{Formula} & \textbf{Description} \\
    \midrule
    \mk{weight\_norm/\{layer\}} & $\lVert W_\ell \rVert_2$ & Per-layer weight L2 norm \\
    \mk{top\_sv/\{layer\}} & $\sigma_1(W_\ell)$ & Per-layer spectral norm \\
    \mk{effective\_rank/\{layer\}} & $(\sum\sigma)^2 / \sum\sigma^2$ & Per-layer effective rank \\
    \mk{grad\_weight\_align/\{layer\}} & $\cos(\mathbf{g}_\ell, W_\ell)$ & Per-layer gradient-weight \\
    & & cosine similarity \\
    \mk{total\_weight\_norm} & $\sqrt{\sum \lVert W_i \rVert^2}$ & Whole-model weight L2 norm \\
    \mk{mean\_weight\_norm} & $\operatorname{mean}(\lVert W_i \rVert)$ & Mean per-layer weight norm \\
    \mk{mean\_top\_sv} & $\operatorname{mean}(\sigma_1(W_i))$ & Mean top SV across layers \\
    \mk{max\_top\_sv} & $\max(\sigma_1(W_i))$ & Maximum top SV across layers \\
    \mk{mean\_effective\_rank} & $\operatorname{mean}((\sum\sigma)^2/\sum\sigma^2)$ & Mean effective rank \\
    & & across layers \\
    \mk{mean\_grad\_weight\_align} & $\operatorname{mean}(\cos(\mathbf{g}_i, W_i))$ & Mean gradient-weight \\
    & & alignment across layers \\
    \end{tabular}
    \caption{Metrics from the \texttt{weight\_tracking} hook.}
    \label{tab:metrics-weight-tracking}
\end{table}

\subsubsection{Per-token gradient distribution}

\begin{table}[H]
    \centering
    \begin{tabular}{c|cc}
    \toprule
    \textbf{Metric} & \textbf{Formula} & \textbf{Description} \\
    \midrule
    \mk{gradient\_sparsity} & $\lvert\{i : \lVert\mathbf{g}_i\rVert < 0.1\bar{n}\}\rvert / p$ & Fraction of rows with \\
    & & small gradients \\
    \mk{gradient\_gini} & Gini index of row norms & Inequality of gradient \\
    & & distribution across tokens \\
    \mk{stride\_group\_variance} & $\operatorname{Var}(\bar{n}_g \text{ per stride group})$ & Variance of mean norm \\
    & & across stride groups \\
    \mk{stride\_group\_max\_ratio} & $\max(\bar{n}_g) / \operatorname{mean}(\bar{n}_g)$ & Stride group dominance ratio \\
    \mk{tokens\_for\_50pct} & Min tokens for 50\% cumulative norm & Gradient concentration \\
    \mk{tokens\_for\_90pct} & Min tokens for 90\% cumulative norm & Gradient concentration \\
    \mk{concentration\_ratio} & $\text{tokens\_for\_50pct} / p$ & Fraction of tokens for \\
    & & 50\% of gradient norm \\
    \end{tabular}
    \caption{Metrics from the \texttt{token\_gradient} hook.}
    \label{tab:metrics-token-gradient}
\end{table}

\subsubsection{Gradient \& displacement projection onto known solution$^\dagger$}

\begin{table}[H]
    \centering
    \begin{tabular}{c|cc}
    \toprule
    \textbf{Metric} & \textbf{Formula} & \textbf{Description} \\
    \midrule
    \mk{grad\_cossim\_to\_solution/\{layer\}} & $\cos(-\nabla_\ell L,\; \theta_{\text{ref}} - \theta_{\text{prev}})$ & Per-layer gradient \\
    & & alignment to solution \\
    \mk{disp\_cossim\_to\_solution/\{layer\}} & $\cos(\Delta\theta_\ell,\; \theta_{\text{ref}} - \theta_{\text{prev}})$ & Per-layer displacement \\
    & & alignment to solution \\
    \mk{overall\_grad\_cossim\_to\_solution} & $\cos(\text{cat}(-\nabla L),\; \text{cat}(\Delta_{\text{ref}}))$ & All-parameter gradient \\
    & & alignment to solution \\
    \mk{overall\_disp\_cossim\_to\_solution} & $\cos(\text{cat}(\Delta\theta),\; \text{cat}(\Delta_{\text{ref}}))$ & All-parameter displacement \\
    & & alignment to solution \\
    \mk{mean\_layer\_grad\_cossim\_to\_solution} & $\operatorname{mean}(\text{grad\_cossim per layer})$ & Mean per-layer gradient \\
    & & alignment \\
    \mk{mean\_layer\_disp\_cossim\_to\_solution} & $\operatorname{mean}(\text{disp\_cossim per layer})$ & Mean per-layer displacement \\
    & & alignment \\
    \mk{displacement\_norm} & $\lVert \theta_t - \theta_{t-1} \rVert_2$ & Total parameter displacement \\
    & & this epoch \\
    \mk{distance\_to\_reference} & $\lVert \theta_{\text{ref}} - \theta_t \rVert_2$ & Euclidean distance from \\
    & & known solution \\
    \end{tabular}
    \caption{Metrics from the \texttt{gradient\_projection} hook.}
    \label{tab:metrics-gradient-projection}
\end{table}

\subsubsection{Gradient subspace dimensionality and information content$^\dagger$}

\begin{table}[H]
    \centering
    \begin{tabular}{c|cc}
    \toprule
    \textbf{Metric} & \textbf{Formula} & \textbf{Description} \\
    \midrule
    \mk{dims\_for\_90pct} & $\min k$ s.t.\ $\sum_{i=1}^{k}\sigma_i^2 / \sum\sigma^2 \geq 0.9$ & SVD dimensions for 90\% \\
    & & explained variance \\
    \mk{participation\_ratio} & $(\sum\sigma^2)^2 / \sum\sigma^4$ & Effective dimensionality \\
    & & of gradient subspace \\
    \mk{top\_sv\_ratio} & $\sigma_1 / \sum\sigma_i$ & Dominance of leading \\
    & & singular value \\
    \mk{svd\_total\_variance} & $\sum\sigma^2$ & Total variance in \\
    & & gradient window \\
    \mk{top1\_explained} & $\sigma_1^2 / \sum\sigma^2$ & Variance explained by \\
    & & top-1 component \\
    \mk{top5\_explained} & $\sum_{i=1}^{5}\sigma_i^2 / \sum\sigma^2$ & Cumulative variance \\
    & & explained by top 5 \\
    \mk{top10\_explained} & $\sum_{i=1}^{10}\sigma_i^2 / \sum\sigma^2$ & Cumulative variance \\
    & & explained by top 10 \\
    \mk{grad\_energy\_fraction\_toward\_solution} & $\sum\sigma_i^2(\mathbf{v}_i \cdot \hat{\mathbf{s}})^2 / \sum\sigma_i^2$ & Energy-weighted alignment \\
    & & of gradient subspace to solution \\
    \mk{top\{k\}\_energy\_fraction\_toward\_solution} & Top-$k$ version ($k = 1, 5, 10$) & Energy toward solution from \\
    & of above & top-$k$ components only \\
    \end{tabular}
    \caption{Metrics from the \texttt{subspace\_gradient\_info} hook.}
    \label{tab:metrics-subspace-gradient-info}
\end{table}

\subsubsection{Parameter update magnitude tracking}

\begin{table}[H]
    \centering
    \begin{tabular}{c|cc}
    \toprule
    \textbf{Metric} & \textbf{Formula} & \textbf{Description} \\
    \midrule
    \mk{relative\_delta} & $\lVert \Delta\theta \rVert / \lVert \theta_{\text{old}} \rVert$ & Fractional parameter change \\
    \mk{absolute\_delta} & $\lVert \theta_{\text{new}} - \theta_{\text{old}} \rVert_2$ & Absolute parameter change \\
    \mk{param\_norm} & $\lVert \theta_{\text{new}} \rVert_2$ & Current parameter L2 norm \\
    \end{tabular}
    \caption{Metrics from the \texttt{parameter\_delta} hook.}
    \label{tab:metrics-parameter-delta}
\end{table}

\subsubsection{Cumulative path length and displacement}

\begin{table}[H]
    \centering
    \begin{tabular}{c|cc}
    \toprule
    \textbf{Metric} & \textbf{Formula} & \textbf{Description} \\
    \midrule
    \mk{path\_length} & $\sum_t \lVert \theta_{t+1} - \theta_t \rVert_2$ & Cumulative distance traveled \\
    & & in parameter space \\
    \mk{net\_displacement} & $\lVert \theta_t - \theta_0 \rVert_2$ & Euclidean distance from \\
    & & initial parameters \\
    \mk{path\_efficiency} & $\lVert \theta_t - \theta_0 \rVert / \text{path\_length}$ & Displacement-to-path ratio; \\
    & & $1 =$ straight line \\
    \end{tabular}
    \caption{Metrics from the \texttt{path\_length} hook.}
    \label{tab:metrics-path-length}
\end{table}

\subsubsection{Batch-level gradient dynamics}

\begin{table}[H]
    \centering
    \begin{tabular}{c|cc}
    \toprule
    \textbf{Metric} & \textbf{Formula} & \textbf{Description} \\
    \midrule
    \mk{lag\_\{n\}} & $\operatorname{mean}\!\bigl(\cos(\mathbf{g}_t, \mathbf{g}_{t-n})\bigr)$ & Mean cosine similarity \\
    & $n \in \{1,2,5,10,20,50\}$ & at lag $n$ steps \\
    \mk{autocorrelation\_mean} & $\operatorname{mean}(\text{lag}_k \text{ for all } k)$ & Overall temporal coherence \\
    & & of gradients \\
    \mk{efficiency\_\{w\}} & $\lVert\sum_{i}^{w}\mathbf{g}_i\rVert / \sum\lVert\mathbf{g}_i\rVert$ & Accumulation efficiency \\
    & $w \in \{2,5,10,20,50\}$ & over $w$-step window \\
    \mk{effective\_rank} & $\exp\!\bigl(-\sum p_i \log p_i\bigr)$ & Effective rank of gradient \\
    & $p_i = \sigma_i^2/\sum\sigma^2$ & history (SV entropy) \\
    \mk{top1\_variance} & $\sigma_1^2 / \sum\sigma_i^2$ & Fraction of gradient variance \\
    & & in top singular vector \\
    \end{tabular}
    \caption{Metrics from the \texttt{batch\_dynamics} hook.}
    \label{tab:metrics-batch-dynamics}
\end{table}

\subsubsection{Standard training diagnostics}

\begin{table}[H]
    \centering
    \begin{tabular}{c|cc}
    \toprule
    \textbf{Metric} & \textbf{Formula} & \textbf{Description} \\
    \midrule
    \mk{loss\_mean} & $\bar{L}$ over emission period & Mean loss \\
    \mk{loss\_std} & $\operatorname{std}(L_t)$ & Std of per-step loss \\
    \mk{loss\_volatility} & $\operatorname{std}(L) / \lvert\bar{L}\rvert$ & Coefficient of variation \\
    & & of loss \\
    \mk{loss\_autocorrelation} & $\operatorname{corr}(L_t, L_{t+1})$ & Lag-1 autocorrelation \\
    & & of loss sequence \\
    \mk{grad\_norm\_mean} & $\overline{\lVert\mathbf{g}\rVert}$ & Mean gradient norm \\
    \mk{grad\_norm\_std} & $\operatorname{std}(\lVert\mathbf{g}\rVert)$ & Std of gradient norms \\
    \mk{grad\_norm\_cv} & $\operatorname{std}(\lVert\mathbf{g}\rVert) / \overline{\lVert\mathbf{g}\rVert}$ & CV of gradient norms \\
    \mk{grad\_norm\_max} & $\max_t \lVert\mathbf{g}_t\rVert$ & Maximum gradient norm \\
    \mk{update\_ratio\_mean} & $\operatorname{mean}(\eta\lVert\mathbf{g}\rVert / \lVert\theta\rVert)$ & Mean update-to-weight ratio \\
    \mk{update\_ratio\_std} & $\operatorname{std}(\eta\lVert\mathbf{g}\rVert / \lVert\theta\rVert)$ & Std of update-to-weight ratio \\
    \mk{weight\_norm} & $\lVert\theta\rVert_2$ & Total parameter L2 norm \\
    \end{tabular}
    \caption{Metrics from the \texttt{training\_diagnostics} hook.}
    \label{tab:metrics-training-diagnostics}
\end{table}

\subsubsection{Per-step entanglement term$^\dagger$ (intervention)}

\begin{table}[H]
    \centering
    \begin{tabular}{c|cc}
    \toprule
    \textbf{Metric} & \textbf{Formula} & \textbf{Description} \\
    \midrule
    \mk{entanglement\_norm} & $\lVert \eta\, H_B\, \mathbf{g}_A \rVert_2$ & L2 norm of the \\
    & & entanglement term \\
    \mk{content\_norm} & $\lVert \mathbf{g}_B + \eta\, H_B\, \mathbf{g}_A \rVert_2$ & L2 norm of the \\
    & & ordering-invariant term \\
    \mk{observed\_grad\_norm} & $\lVert \mathbf{g}_B \rVert_2$ & Observed batch gradient norm \\
    \mk{entanglement\_energy\_ratio} & $\lVert\text{ent}\rVert^2 / \lVert\mathbf{g}_B\rVert^2$ & How much ordering \\
    & & dominates the gradient \\
    \mk{entanglement\_content\_cossim} & $\cos(\text{ent},\, \text{content})$ & Whether ordering reinforces \\
    & & or opposes content \\
    \mk{rayleigh\_quotient} & $\mathbf{g}_A^\top H_B\, \mathbf{g}_A / \lVert\mathbf{g}_A\rVert^2$ & Curvature along prior \\
    & & gradient direction \\
    \mk{amplification\_ratio} & $\lVert H_B\, \mathbf{g}_A \rVert / \lVert\mathbf{g}_A\rVert$ & Hessian amplification \\
    & & of prior gradient \\
    \mk{edge\_of\_stability} & $\text{amplification} \times 2\eta$ & Stability boundary \\
    & & indicator \\
    \mk{entanglement\_cossim\_to\_solution} & $\cos(-\text{ent},\; \theta_{\text{ref}} - \theta)$ & Entanglement alignment \\
    & & toward known solution \\
    \mk{content\_cossim\_to\_solution} & $\cos(-\text{content},\; \theta_{\text{ref}} - \theta)$ & Content alignment \\
    & & toward known solution \\
    \mk{entanglement\_coherence} & $\cos(\text{ent}_t,\, \text{ent}_{t-1})$ & Consistency of ordering \\
    & & direction across steps \\
    \mk{entanglement\_norm/\{layer\}} & $\lVert\eta\, H_B\, \mathbf{g}_A\rVert_\ell$ & Per-layer entanglement norm \\
    \mk{content\_norm/\{layer\}} & $\lVert\mathbf{g}_B + \eta H_B \mathbf{g}_A\rVert_\ell$ & Per-layer content norm \\
    \mk{entanglement\_energy\_ratio/\{layer\}} & $\lVert\text{ent}_\ell\rVert^2 / \lVert\mathbf{g}_{B,\ell}\rVert^2$ & Per-layer entanglement \\
    & & energy ratio \\
    \mk{entanglement\_cossim\_to\_solution/\{layer\}} & $\cos(-\text{ent}_\ell,\; \Delta_{\text{ref},\ell})$ & Per-layer entanglement \\
    & & alignment to solution \\
    \mk{content\_cossim\_to\_solution/\{layer\}} & $\cos(-\text{content}_\ell,\; \Delta_{\text{ref},\ell})$ & Per-layer content \\
    & & alignment to solution \\
    \end{tabular}
    \caption{Metrics from the \texttt{hessian} hook.
    $\eta H_B \mathbf{g}_A$ is computed via finite-difference Hv product.}
    \label{tab:metrics-hessian}
\end{table}

\subsubsection{Counterfactual ordering analysis$^\dagger$ (intervention)}

\begin{table}[H]
    \centering
    \begin{tabular}{c|cc}
    \toprule
    \textbf{Metric} & \textbf{Formula} & \textbf{Description} \\
    \midrule
    \mk{counterfactual\_mean\_norm} & $\lVert\bar{\mathbf{g}}_{\text{cf}}\rVert_2$ & Norm of mean \\
    & & shuffled-epoch gradient \\
    \mk{content\_component\_norm} & $\lVert\operatorname{proj}(\mathbf{g} \to \bar{\mathbf{g}}_{\text{cf}})\rVert_2$ & Norm of ordering-invariant \\
    & & content component \\
    \mk{ordering\_component\_norm} & $\lVert\mathbf{g} - \text{content}\rVert_2$ & Norm of ordering-specific \\
    & & component \\
    \mk{ordering\_fraction} & $\lVert\text{ordering}\rVert^2 / \lVert\text{actual}\rVert^2$ & Fraction of gradient energy \\
    & & from ordering \\
    \mk{ordering\_alignment} & $\cos(\mathbf{g},\, \bar{\mathbf{g}}_{\text{cf}})$ & Cosine between actual \\
    & & and shuffled gradients \\
    \mk{content\_grad\_cossim\_to\_solution} & $\cos(-\text{content},\; \Delta_{\text{ref}})$ & Content alignment \\
    & & to solution \\
    \mk{ordering\_grad\_cossim\_to\_solution} & $\cos(-\text{ordering},\; \Delta_{\text{ref}})$ & Ordering alignment \\
    & & to solution \\
    \mk{cf\_grad\_cossim\_to\_solution} & $\cos(-\bar{\mathbf{g}}_{\text{cf}},\; \Delta_{\text{ref}})$ & Any-ordering alignment \\
    & & to solution \\
    \mk{content\_energy\_fraction\_toward\_solution} & $\sum\sigma_i^2(\mathbf{v}_i \cdot \hat{\mathbf{s}})^2 / \sum\sigma_i^2$ & Content subspace energy \\
    & (window SVD) & aimed at solution \\
    \mk{ordering\_energy\_fraction\_toward\_solution} & $\sum\sigma_i^2(\mathbf{v}_i \cdot \hat{\mathbf{s}})^2 / \sum\sigma_i^2$ & Ordering subspace energy \\
    & (window SVD) & aimed at solution \\
    \mk{\{component\}\_component\_norm/\{layer\}} & Per-parameter norms & Per-layer content and \\
    & & ordering magnitudes \\
    \mk{ordering\_fraction/\{layer\}} & Per-parameter fraction & Per-layer ordering \\
    & & energy fraction \\
    \mk{ordering\_alignment/\{layer\}} & $\cos(\mathbf{g}_\ell, \bar{\mathbf{g}}_{\text{cf},\ell})$ & Per-layer ordering \\
    & & alignment \\
    \mk{\{comp\}\_grad\_cossim\_to\_solution/\{layer\}} & Per-parameter alignment & Per-layer content/ordering/cf \\
    & & alignment to solution \\
    \end{tabular}
    \caption{Metrics from the \texttt{counterfactual} hook.
    Decomposition uses $K$ shuffled training epochs.
    $\Delta_{\text{ref}} = \theta_{\text{ref}} - \theta_{\text{prev}}$.}
    \label{tab:metrics-counterfactual}
\end{table}

\subsubsection{Adam optimizer state dynamics$^\dagger$ (intervention)}

\begin{table}[H]
    \centering
    \begin{tabular}{c|cc}
    \toprule
    \textbf{Metric} & \textbf{Formula} & \textbf{Description} \\
    \midrule
    \multicolumn{3}{l}{\textit{Tier 1: Direction-agnostic}} \\
    \midrule
    \mk{momentum\_grad\_cossim} & $\cos(\mathbf{m},\, \mathbf{g})$ & First moment vs.\ current \\
    & & gradient \\
    \mk{amplification\_ratio} & $\lVert\text{update}\rVert / \lVert\eta\,\mathbf{g}\rVert$ & Adam update norm vs.\ \\
    & & raw SGD update norm \\
    \mk{update\_deflection} & $\lVert\text{update}_\perp\rVert / \lVert\text{update}\rVert$ & Fraction of update \\
    & & orthogonal to gradient \\
    \mk{effective\_lr\_cv} & $\operatorname{std}(\eta_{\text{eff}}) / \operatorname{mean}(\eta_{\text{eff}})$ & CV of per-element \\
    & $\eta_{\text{eff}} = \text{lr}/(\sqrt{\hat{v}} + \epsilon)$ & effective learning rates \\
    \midrule
    \multicolumn{3}{l}{\textit{Tier 2: Solution-dependent}} \\
    \midrule
    \mk{momentum\_solution\_cossim} & $\cos(\mathbf{m},\; \theta_{\text{ref}} - \theta)$ & First moment alignment \\
    & & to solution \\
    \mk{update\_solution\_cossim} & $\cos(\text{update},\; \theta_{\text{ref}} - \theta)$ & Adam update alignment \\
    & & to solution \\
    \mk{grad\_solution\_cossim} & $\cos(\mathbf{g},\; \theta_{\text{ref}} - \theta)$ & Raw gradient alignment \\
    & & to solution (baseline) \\
    \mk{optimizer\_solution\_amplification} & $\text{update\_cos} - \text{grad\_cos}$ & Optimizer improvement of \\
    & & solution alignment \\
    \midrule
    \multicolumn{3}{l}{\textit{Tier 3: Probe-dependent}$^\ddagger$} \\
    \midrule
    \mk{momentum\_probe\_cossim} & $\cos(\mathbf{m},\; \mathbf{g}_{\text{target}})$ & First moment alignment \\
    & & to probe gradient \\
    \mk{update\_probe\_cossim} & $\cos(\text{update},\; \mathbf{g}_{\text{target}})$ & Adam update alignment \\
    & & to probe gradient \\
    \mk{grad\_probe\_cossim} & $\cos(\mathbf{g},\; \mathbf{g}_{\text{target}})$ & Raw gradient alignment \\
    & & to probe (baseline) \\
    \mk{optimizer\_probe\_amplification} & $\text{update\_probe\_cos} - \text{grad\_probe\_cos}$ & Optimizer improvement of \\
    & & probe alignment \\
    \end{tabular}
    \caption{Metrics from the \texttt{adam\_dynamics} hook.
    Adam update is $\text{lr} \cdot \hat{m} / (\sqrt{\hat{v}} + \epsilon)$;
    weight decay is excluded to isolate adaptive dynamics.}
    \label{tab:metrics-adam-dynamics}
\end{table}

\subsection{Stride Frequency Validation}
\label{apdx:stride-frequency}

To validate the assertion that the model is learning a representation that is dictated by the stride $s = \lfloor \sqrt{p} \rfloor$ for sort $a \bmod s$, several smaller runs of the \stride{} model were performed for a length of 50 epochs to observe the initial frequency $F$ emergence with selected stride values.

\begin{table}[h]
\centering
\caption{Predicted vs. observed fundamental frequency}
\label{tab:stride-freq}
\begin{tabular}{c|cc}
\toprule
Stride $s$ & Predicted $F = \lfloor p/s \rceil$ & Observed $F$ \\
\midrule
50  & 199 & 199 \\
99  & 101 & 101 \\
150 &  66 &  66 \\
\bottomrule
\end{tabular}
\end{table}

\subsection{Target Failure Mode: Additional Analysis}
\label{apdx:target-detail}

This appendix provides additional analysis of the \target{} failure mode discussed in Section~\ref{sec:target-failure}, focusing on capacity allocation patterns and evidence that the spectral organization observed under \target{} is not solely attributable to weight-decay collapse.

\paragraph{Capacity allocation.} Under \random{}, the model builds a lookup table: by epoch 5{,}000, 87\% of weight capacity (squared Frobenius norm) resides in the feedforward layers, with the decoder retaining 5\%. Under \target{}, the decoder collapses to 1\% of total capacity while the feedforward layers absorb 64\% and attention layers grow to 34\%. The decoder cannot stabilize a memorization table when its targets are overwritten every batch, so gradient pressure is redirected into the transformer body. But the body cannot commit to representations under the oscillating gradient signal.

This pattern is consistent with the spectral dissociation reported in Section~\ref{sec:target-failure}: the ordering signal organizes the decoder spectrally while simultaneously preventing it from accumulating weight capacity. The decoder is structured but weak; the transformer body is strong but unstructured.

\paragraph{Matched-norm spectral comparison.} Both non-generalizing strategies undergo embedding collapse under weight decay: embedding norm falls from 1{,}420 to 13 for \target{} and from 1{,}417 to 13 for \random{}. Because spectral entropy is computed on the normalized power spectrum, weight-decay collapse amplifies any pre-existing spectral non-uniformity as total energy shrinks. This raises the question of whether the greater low-frequency concentration observed under \target{} (0.47 vs.\ 0.10 for \stride{} at the final epoch) is a genuine ordering effect or merely an artifact of differential collapse rates.

To control for this, we compare low-frequency power at epochs where the two strategies have matched embedding norms. At norm $\approx$13 (the final value for both), \target{} shows approximately 5.6$\times$ more low-frequency concentration than \random{} (0.47 vs.\ 0.085). At norm $\approx$30 (an intermediate collapse point), \target{} shows moderately more low-frequency concentration than \random{} (approximately 0.07 vs.\ 0.06), though the \target{} values are noisier at this norm range due to the oscillatory gradient dynamics. The ordering effect is present at matched norms across the collapse trajectory, with the gap widening as total energy decreases, consistent with weight decay amplifying a genuine underlying spectral asymmetry rather than creating one.

\subsection{Optimizer Dynamics: Additional Analysis}
\label{apdx:adam-detail}

This appendix provides additional detail on the interaction between the Adam optimizer and the ordering channel, supplementing the summary in Section~\ref{sec:adam-empirical}.

\begin{figure}[!ht]
    \centering
    \includegraphics[width=1\linewidth]{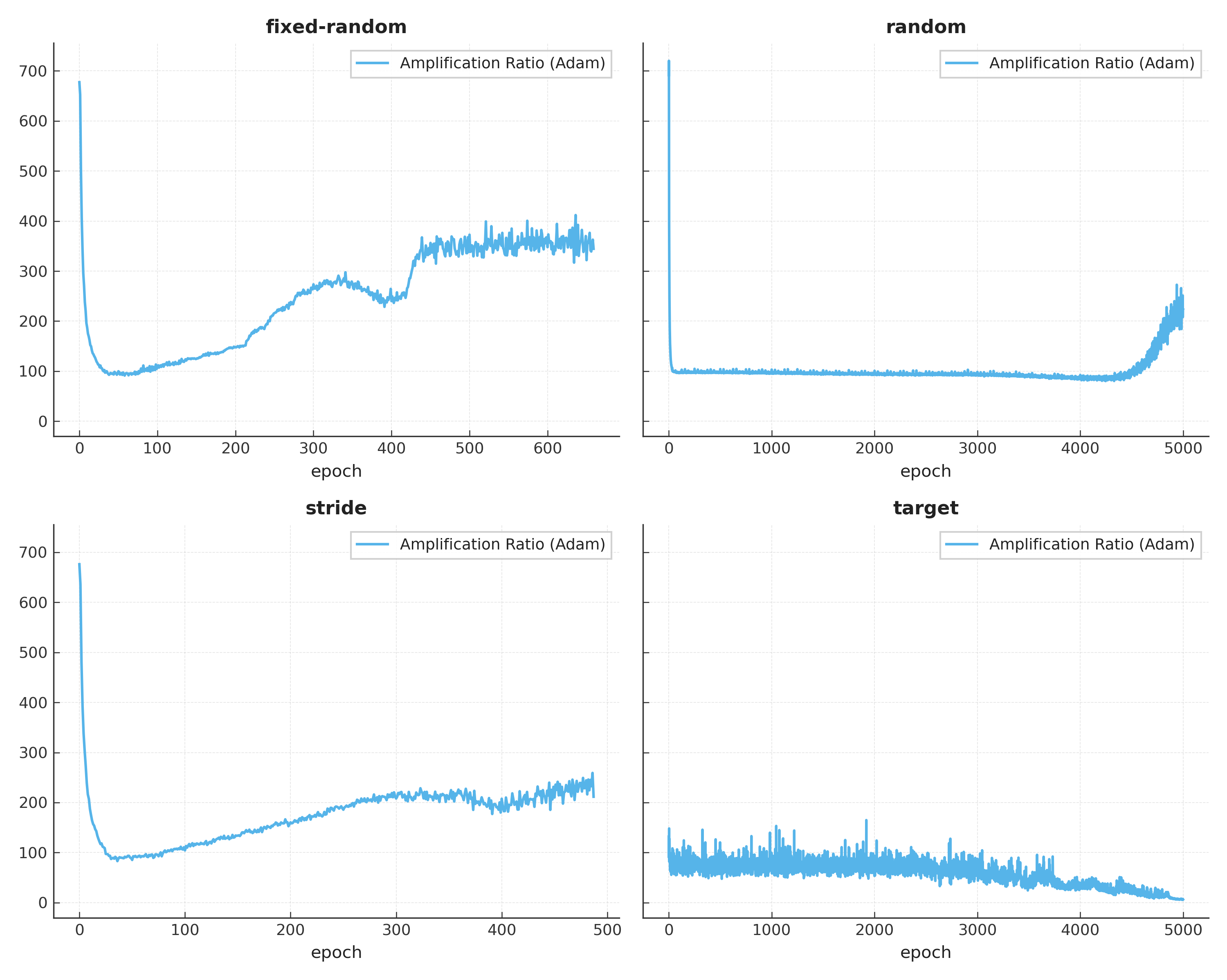}
    \caption{Measured amplification ratio of AdamW updates per strategy.}
    \label{fig:adam_amplification_ratio}
\end{figure}

\paragraph{Amplification ratio trajectories.} The Adam amplification ratio ($\|\Delta\theta_{\text{Adam}}\| / \|\eta \nabla L\|$) varies substantially across strategies and training phases (Figure~\ref{fig:adam_amplification_ratio}). All strategies begin with high amplification ($\sim$680$\times$) at initialization, when the second moment estimates are still warming up. Under \random{}, the ratio rapidly declines to $\sim$100$\times$ and remains stable, reflecting the statistically stationary gradient distribution produced by IID shuffling. Under \stride{} and \fixedrandom{}, the ratio also declines initially but then \emph{increases} through mid- and late training (to 212$\times$ and 345$\times$ respectively by the final epoch), indicating that Adam assigns progressively larger effective step sizes as the model converges and gradient magnitudes decrease in the relevant subspace. Under \target{}, the ratio declines monotonically to 6$\times$, consistent with the large, oscillating gradient magnitudes that keep the second moment estimates elevated and suppress per-parameter amplification.

\begin{figure}[!ht]
    \centering
    \includegraphics[width=1\linewidth]{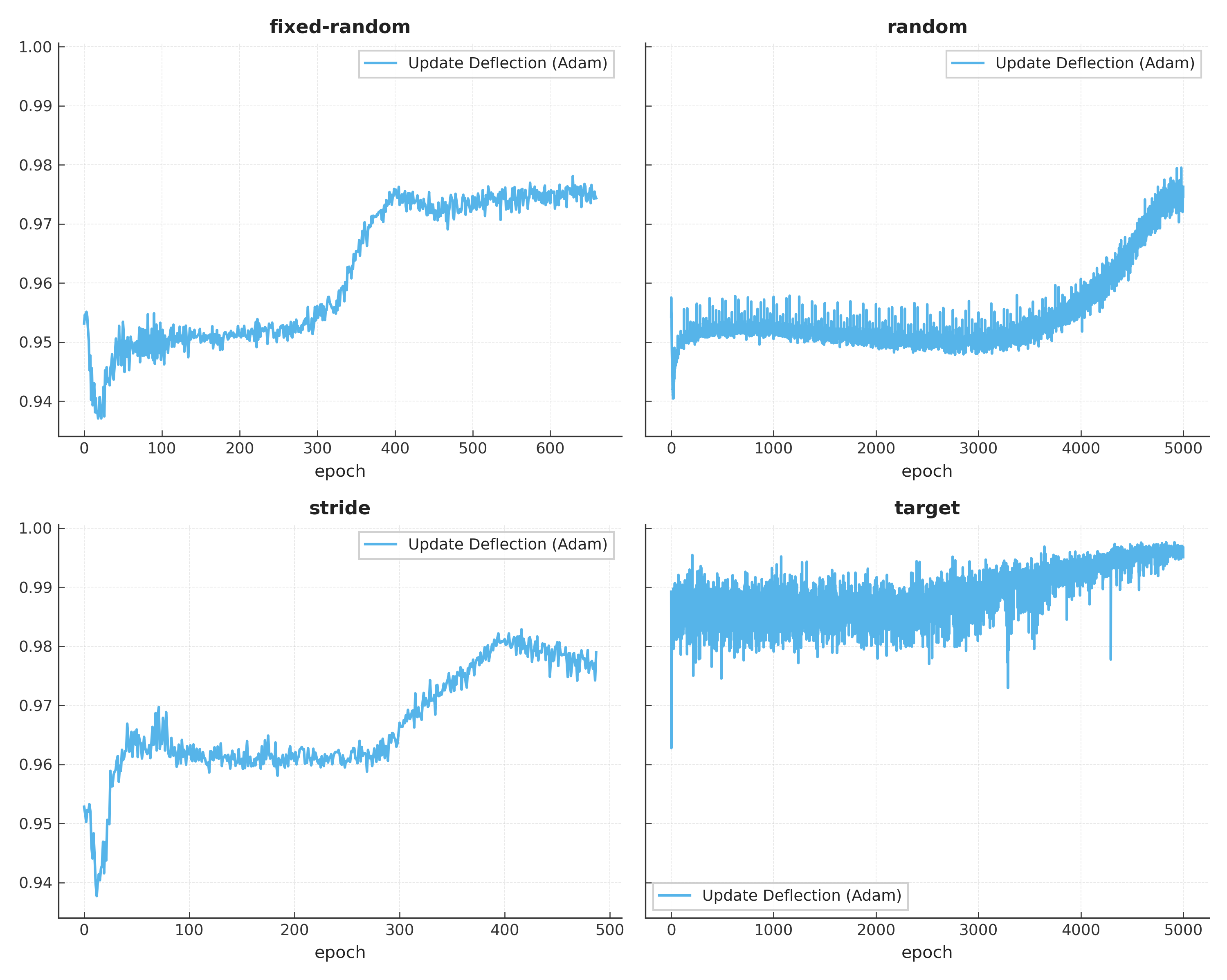}
    \caption{Measured update deflection of AdamW updates per strategy. All strategies experience high deflection.}
    \label{fig:adam_update_deflection}
\end{figure}

\paragraph{Update deflection.} The fraction of Adam's update orthogonal to the raw gradient (update deflection) is above 0.95 for all strategies (Figure~\ref{fig:adam_update_deflection}), meaning that momentum and adaptive scaling redirect nearly all of the update direction. This is not pathological: it reflects Adam's intended behavior of normalizing gradient magnitudes and accumulating directional information across steps. The deflection is highest for \target{} (0.989 mean), where the oscillating gradient direction causes momentum to average over contradictory signals, producing updates that bear little resemblance to any individual gradient. The deflection is lowest for \random{} (0.954 mean), where the lack of temporal gradient structure means momentum has less directional information to contribute beyond the current step.

\begin{figure}[!ht]
    \centering
    \includegraphics[width=1\linewidth]{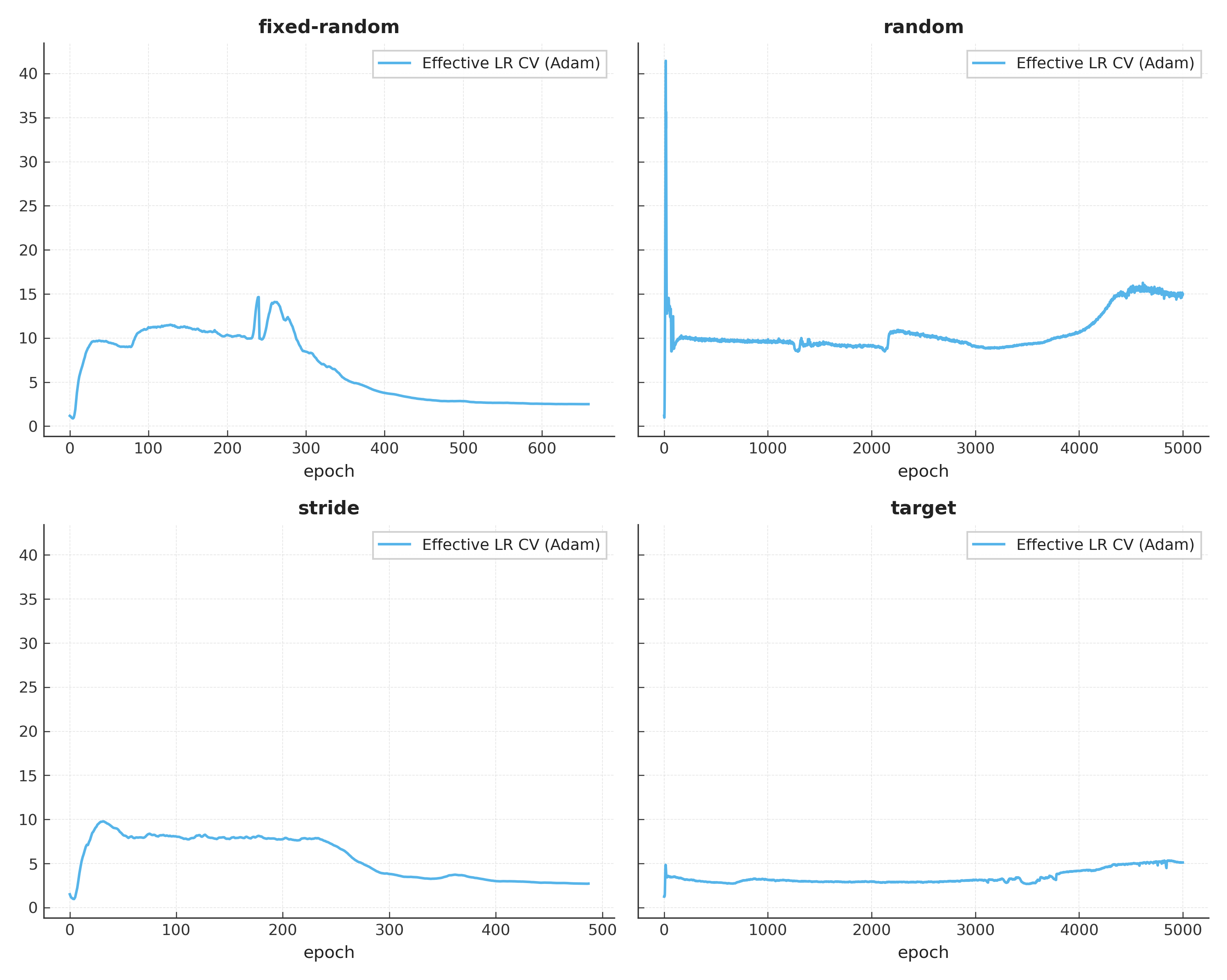}
    \caption{Measured effective learning-rate coefficient of variation.}
    \label{fig:adam_lr_cv}
\end{figure}

\paragraph{Effective learning rate non-uniformity.} The coefficient of variation of per-parameter effective learning rates ($\eta / (\sqrt{\hat{v}_t} + \epsilon)$) measures how non-uniformly Adam scales different parameters (Figure~\ref{fig:adam_lr_cv}). Under \stride{} (CV = 5.7), the ordering signal produces consistent gradient structure across the parameters that participate in the $F = 101$ harmonic series, leading to relatively uniform second-moment estimates and thus uniform effective learning rates in the relevant subspace. Under \fixedrandom{} (CV = 6.6), the more diffuse spectral signal produces greater parameter-level variation, requiring Adam to apply more differentiated scaling. Under \random{} (CV = 10.6), the incoherent ordering produces highly variable gradient histories across parameters, resulting in the most non-uniform adaptation. The ordering strategy thus shapes not only the gradient signal but also the optimizer's internal state, creating strategy-dependent amplification profiles across the parameter space.

\paragraph{Relationship to entanglement.} The Hessian entanglement energy ratio ($\|H_B \cdot \Delta\theta\|^2 / \|\nabla L_B\|^2$) reflects the combined effect of curvature and displacement magnitude. Under \stride{} and \fixedrandom{}, the ratio averages $860$--$900\times$ and peaks above $3{,}000\times$ during the most active learning phases. These large ratios arise from a striking geometric structure: at each training step, the entanglement term $\eta H_B \cdot g_A$ and the content term $g_B(\theta)$ are nearly identical vectors. Their norms differ by only 1--3\%, and their cosine similarity exceeds 0.999 throughout training for all three non-adversarial strategies. The gradient the optimizer actually sees, $g_B(\theta') = g_B(\theta) - \eta H_B \cdot g_A$, is the small residual after this near-cancellation, typically 30--55$\times$ smaller in norm than either component for the non-adversarial strategies. (See Table~\ref{tab:hessian-geometry}.) The energy ratio is simply the square of this norm ratio.

The ordering signal lives in this residual. The 0.1\% angular difference between two massive, nearly-parallel vectors determines the direction of the observed gradient, and thus the direction of learning. Small changes in the relationship between consecutive batches, which batch follows which, produce small angular perturbations in the entanglement vector, which are amplified into large directional changes in the residual. This is the geometric mechanism by which ordering exerts disproportionate influence on the gradient.

The energy ratio is highest for \random{} ($1{,}288\times$ mean), which may seem paradoxical: the non-generalizing strategy has the largest entanglement. However, this reflects \random{}'s incoherent entanglement pointing in random directions at each step, producing large per-step magnitudes that cancel over time. Under \target{}, the geometry is qualitatively different: the entanglement-content cosine similarity drops to 0.51 by late training, the entanglement norm falls well below the content norm (ratio 0.46), and the observed gradient is only 1.7$\times$ smaller than the entanglement rather than 30--55$\times$. The near-cancellation breaks down because the anti-correlated consecutive gradients produce entanglement that is misaligned with content, leaving a large, chaotic residual that prevents coherent learning.

\subsection{Dose-Productivity: Autocorrelation Mean}
\label{apdx:autocorrelation-mean}

During training, a window of 50 previous batch gradients was maintained to examine the relationships on different time-scales between various gradients. The following figures illustrate some of this data.

\begin{figure}[!htb]
    \centering
    \includegraphics[width=1\linewidth]{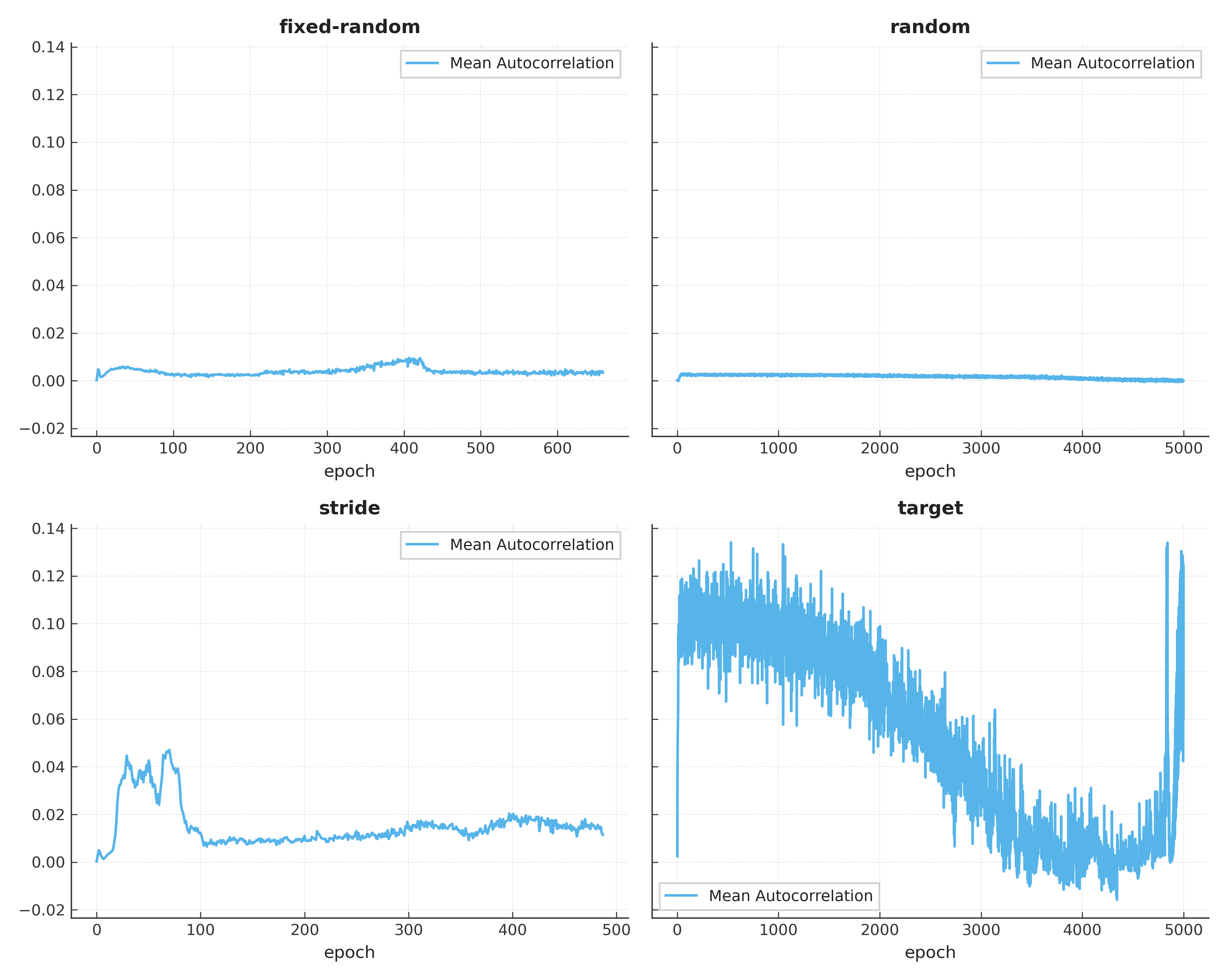}
    \caption{The autocorrelation between batch gradients of the \target{} strategy are so large that they make the details of the other strategies difficult to examine on this scale.}
    \label{fig:autocorrelation-all}
\end{figure}

\begin{figure}[!htb]
    \centering
    \includegraphics[width=1\linewidth]{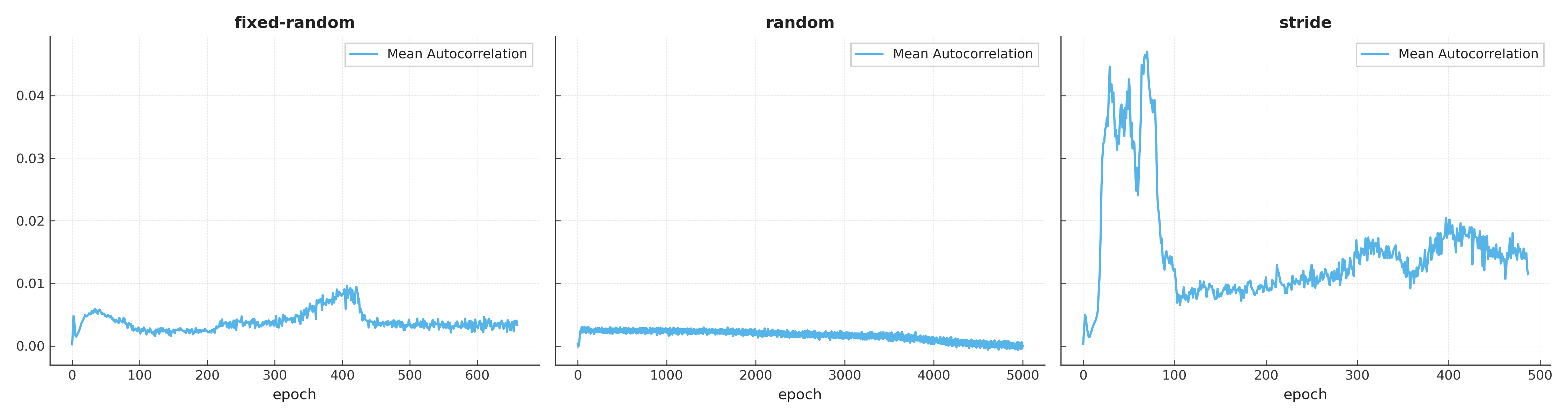}
    \caption{Removing the \target{} strategy allows us to see the behavior of \stride{}, which peaks during the critical learning period.}
    \label{fig:autocorrelation-notarget}
\end{figure}

\begin{figure}[!htb]
    \centering
    \includegraphics[width=1\linewidth]{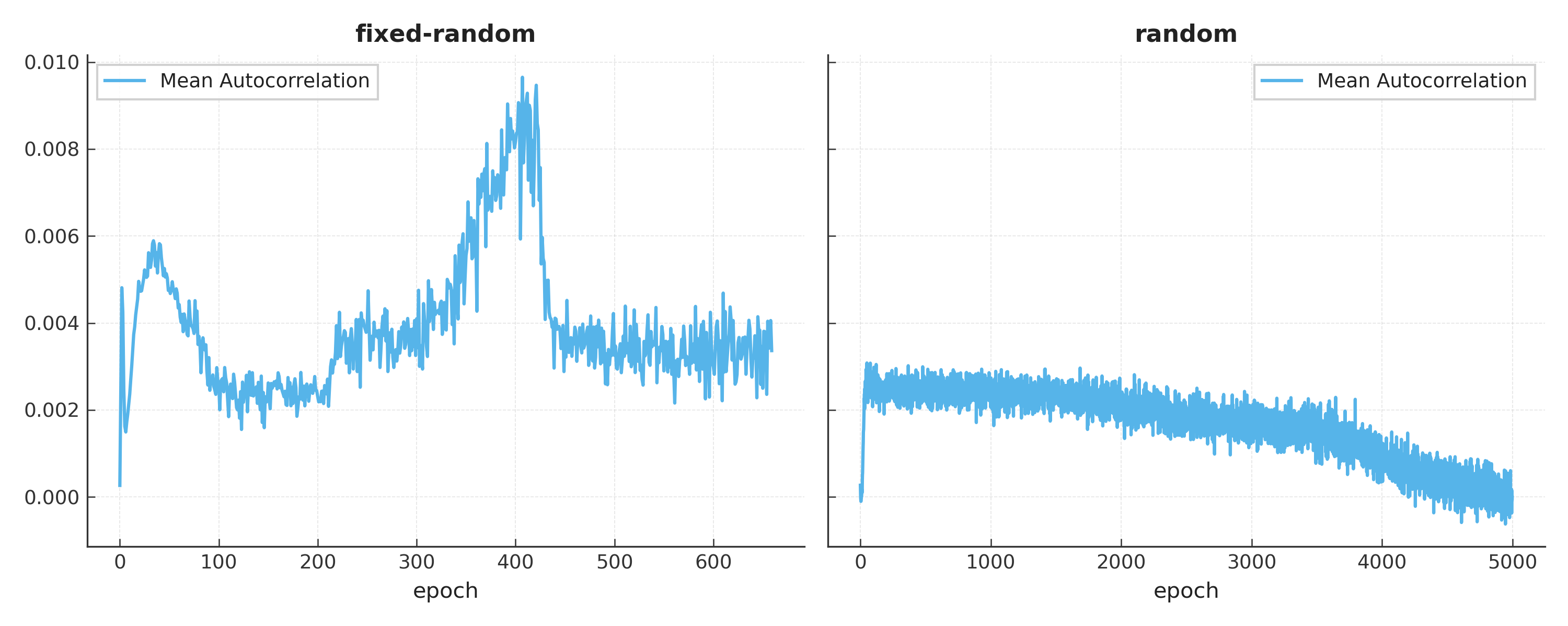}
    \caption{Removing both \target{} and \stride{} allows us to see just how little ordering information and consistency was necessary for generalization to develop in the \fixedrandom{} strategy compared to the \random{} strategy.}
    \label{fig:autocorrelation-randoms}
\end{figure}

\end{document}